%% file: main.tex
\documentclass[Afour,sageh,times]{sagej} 

\usepackage{moreverb,url}
\usepackage{soul}

\usepackage[colorlinks,bookmarksopen,bookmarksnumbered,citecolor=red,urlcolor=red]{hyperref}

\newcommand\BibTeX{{\rmfamily B\kern-.05em \textsc{i\kern-.025em b}\kern-.08em
T\kern-.1667em\lower.7ex\hbox{E}\kern-.125emX}}

\def\abovestrut#1{\rule[0in]{0in}{#1}\ignorespaces}
\def\belowstrut#1{\rule[-#1]{0in}{#1}\ignorespaces}

\input{math_definition.tex}
\hypersetup{draft}

\usepackage{xspace}
\newcommand{\rej}{{\sc{rejection}}\xspace}
\newcommand{\lse}{{\sc gp-lse}\xspace}
\newcommand{\diverse}{{\sc{diverse}}\xspace}
\newcommand{\adapt}{{\sc{adaptive}}\xspace}
\newcommand{\gk}{{\sc{diverse-gk}}\xspace}
\newcommand{\lk}{{\sc{diverse-lk}}\xspace}

\newcount\Comments  
\usepackage{array,multirow,graphicx}
\usepackage{color}
\usepackage{wrapfig}
\definecolor{darkgreen}{rgb}{0,0.5,0}
\definecolor{purple}{rgb}{1,0,1}

\newcommand{\kibitz}[2]{\ifnum\Comments=1\textcolor{#1}{#2}\fi}
\newcommand{\zw}[1]{\kibitz{purple}      {[ZW: #1]}}
\newcommand{\caelan}[1]  {\kibitz{blue}{[CG: #1]}}

\newcommand{\highlight}[1]{#1} 

\newcommand{\gp}{{\sc gp}}
\newcommand{\GP}{{\rm GP}}  
\newcommand{\PDDL}{{\sc PDDL}}
\newcommand{\stripstream}{\PDDL{}Stream}
\newcommand{\tamp}{{\sc tamp}}
\newcommand{\dpp}{{\sc dpp}}

\newcommand{\pddl}[1]{\texttt{#1}} 
\newcommand{\pddlkw}[1]{\textbf{\pddl{#1}}}
\newcommand{\proc}[1]{\textsc{#1}}

\newcommand{\SE}[1]{\mathrm{SE}(#1)}
\newcommand{\indicator}[1]{[#1]} 

\usepackage{listings}
\def\volumeyear{2020}

\begin{document}

\runninghead{Wang \textit{et~al.}}

\title{Learning compositional models of robot skills for task and motion planning}

\author{Zi Wang$^*$\affilnum{1,}\affilnum{2}, Caelan Reed Garrett$^*$\affilnum{1}, Leslie Pack Kaelbling\affilnum{1}, and Tom\'as Lozano-P\'erez\affilnum{1}}

\affiliation{$^*$Equal contribution. \affilnum{1}MIT CSAIL, MA. \affilnum{2}Now at Google. }

\corrauth{Caelan Reed Garrett,  
MIT CSAIL,
32 Vassar St, 
Cambridge, MA 02139.}

\email{caelan@csail.mit.edu}

\begin{abstract}
\input{abstract}
\end{abstract}


\maketitle
\input{intro3} 
\input{formulation}
\input{estimation}
\input{planning}
\input{related} 
\input{exp}
\input{real_world}
\begin{acks}
We gratefully acknowledge support from NSF grants 1523767 and 1723381; from AFOSR grant FA9550-17-1-0165; from ONR grant N00014-18-1-2847; from the Honda Research Institute; and from SUTD Temasek Laboratories.
Caelan Garrett is supported by an NSF GRFP fellowship with primary award number 1122374.
Any opinions, findings, and conclusions or recommendations expressed in this material are those of the authors and do not necessarily reflect the views of our sponsors.

We thank Kevin Chen, Nishad Gothoskar, Ivan Jutamulia, Alex LaGrassa, Jiayuan Mao, Skye Thompson, and Jingxi Xu for their help with developing the infrastructure for the simulated and real-world experiments.
\end{acks}

\input{appendix}

\bibliographystyle{SageH}

\end{document}

%% file: math_definition.tex
\usepackage{textcase}
\usepackage{subcaption}
\usepackage{graphicx} 
\usepackage{booktabs} 
\usepackage{amssymb}
\usepackage{times}
\usepackage{epsfig}
\usepackage{graphicx}
\usepackage{color}
\usepackage{url}
\usepackage{bbm}
\usepackage{multicol}

\providecommand{\hide}[1]{}

\usepackage{amsmath,amsfonts}
\usepackage{amsopn,amssymb}

\usepackage{amsthm}

\theoremstyle{plain}
\newtheorem{thm}{Theorem}

\newtheorem{cor}[thm]{Corollary}

\theoremstyle{definition}

\theoremstyle{remark}

\usepackage{bm} 
\usepackage{algorithmicx}
\usepackage{algorithm}
\PassOptionsToPackage{noend}{algpseudocode}
\usepackage{algpseudocode}
\algnewcommand{\LineComment}[1]{\Statex \(\triangleright\) #1}

\newcommand{\vct}[1]{\boldsymbol{#1}} 
\newcommand{\mat}[1]{\boldsymbol{#1}} 

\newcommand{\field}[1]{\mathbb{#1}}
\newcommand{\R}{\field{R}} 
\newcommand{\T}{^{\textrm T}} 

\newcommand{\ProbOpr}[1]{\mathbb{#1}}

\newcommand{\expect}[2]{%
\ifthenelse{\equal{#2}{}}{\ProbOpr{E}_{#1}}
{\ifthenelse{\equal{#1}{}}{\ProbOpr{E}\left[#2\right]}{\ProbOpr{E}_{#1}\left[#2\right]}}} 
\newcommand{\var}[2]{%
\ifthenelse{\equal{#2}{}}{\ProbOpr{VAR}_{#1}}
{\ifthenelse{\equal{#1}{}}{\ProbOpr{VAR}\left[#2\right]}{\ProbOpr{VAR}_{#1}\left[#2\right]}}} 

\DeclareMathOperator*{\argmax}{arg\,max}

\newcommand{\vx}{{\vct{x}}}

\newcommand{\vy}{\vct{y}}

\newcommand{\vk}{\vct{k}}

\newcommand{\vxi}{\vct{\xi}}

\newcommand{\mI}{\mat{I}}
\newcommand{\mK}{\mat{K}}

   \newcommand{\cd}{\mathcal{D}}

\algnewcommand{\algorithmicgoto}{\textbf{go to}}%
\algnewcommand{\Goto}[1]{\algorithmicgoto~\ref{#1}}%

%% file: abstract.tex
The objective of this work is to augment the basic abilities of a
robot by learning to use sensorimotor primitives to 
solve complex long-horizon manipulation problems. 
This requires flexible generative planning that
can combine primitive abilities in novel combinations and thus generalize across a wide variety of problems.
In order to plan with 
primitive actions, we must have models of 
the actions:  under what circumstances will executing
this primitive successfully achieve some particular effect in the world?

We use, and develop novel improvements on, state-of-the-art methods
for active learning and sampling. 
We use Gaussian process methods for
learning the constraints on skill effectiveness from small numbers
of expensive-to-collect training examples. 
Additionally, we develop efficient adaptive sampling methods for generating a comprehensive and diverse sequence of continuous candidate control parameter values (such as pouring waypoints for a cup) 
during planning. 
\highlight{These values become end-effector goals for traditional motion planners that then solve for a full robot motion that performs the skill.}
\highlight{By using learning and planning methods in conjunction, we take advantage of the strengths of each and plan for a wide variety of complex dynamic manipulation tasks.}
We demonstrate our approach in an integrated system, combining traditional robotics primitives with our newly learned models using an efficient robot task and motion planner. 
We evaluate our approach both in simulation and in the real world through measuring the quality of the selected primitive actions.
Finally, we apply our integrated system to a variety of long-horizon simulated and real-world manipulation problems.

%% file: intro3.tex
\section{Introduction} 
\label{sec:intro}
For robots to be useful in a home environment, they will have to be
endowed with a foundational set of capabilities, such as locomotion
and basic object manipulation.  They will then have to build on those
capabilities by acquiring more specialized skills such as pouring
milk or scooping cereal.  It is critical that these skills
be acquired {\em efficiently}, with relatively few training examples, and
that they be used {\em compositionally}, combining with existing
skills to generalize to a wide variety of situations and purposes for
which that skill can be usefully deployed.

The vast majority of research in robot learning has focused on
acquiring closed-loop sensorimotor skills, ranging from
pouring~\citep{yamaguchi2014learning} to manipulating a Rubik's
cube~\citep{openai2019solving}.  Very little work has focused on how to actually
combine and execute these skills to address problems in the world (but see the work of~\citet{wang2019learning} for a good example).  
In this paper, we provide a framework for integrating new skills with
existing ones by learning {\em skill models} and using them to plan
sequences of skill executions to achieve long-horizon goals in complex
environments. 

\begin{figure}[ht]
\centering
    \includegraphics[width=1.\columnwidth]{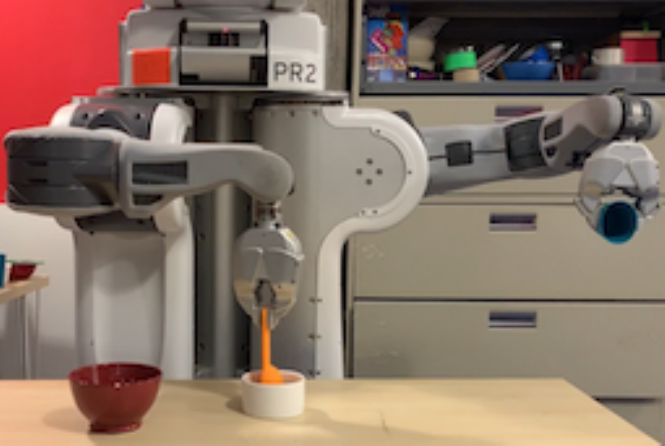}
    \caption{Making coffee in {\em KitchenPR2}, which requires pouring, scooping, dumping, and stirring. } 
\label{fig:settings}
\end{figure}

The overall class of problems we wish to address, known as {\em task and motion planning} ({\sc tamp}), considers a robot carrying out tasks
in an environment such as a kitchen, storage depot,
and construction site.  These tasks require the robot to manipulate
multiple objects, potentially moving things out of the way, or putting
them into or taking them out of containers, as well as to perform
additional operations, such as pouring or cleaning, in service of a
high-level objective.  {\sc tamp} planners combine robot motion
planning with the selection of continuous parameters, for example governing grasps
and placements of objects, and the high-level selection of which
operations to perform on which objects in what order.

In this paper, we will use making coffee as a motivating example.  This task involves
picking and placing a variety of objects, such as cups and spoons.  It also involves pouring cream from one container to another and scooping sugar from a bowl into a cup.  These actions need to be performed in a wide variety of object arrangements on a tabletop, with the relevant objects in arbitrary initial poses and possibly in the presence of extraneous objects.  The robot should be able to apply its skills of picking, placing, and pushing objects as well as its learned skills of pouring and scooping to enable successful completion of the coffee-making task in these arbitrary tabletop environments.

Figure~\ref{fig:settings} demonstrates this task using a real-world PR2 robot.
Figure~\ref{fig:pr2simu} ({\em right}) shows a 3D simulation version of this task in PyBullet~\citep{coumans2019}.  We use simulations to carry out extensive evaluation of our learning algorithms. 
However, crucially, learning on the real robot {\em does not} rely on the simulation.
\highlight{We do not want to be limited to skills for which a high-fidelity simulation is required, so we attempt to learn on a real robot in as few trials as possible.}

{\sc tamp} problems are {\em hybrid}, requiring discrete and continuous choices. {\sc tamp} planning approaches generally combine aspects of discrete planning methods
from symbolic artificial intelligence (AI) with constrained sampling or direct optimization to select
continuous parameters.  Our approach will be to learn models of new skills that allow
them to be incorporated into a {\sc tamp} framework and immediately
combined with existing skills in service of achieving high-level goals.

Given a parameterized sensorimotor policy (a {\em skill}) $\pi_{O(\omega)}$ that was intended to achieve some condition in the world (such as liquid being in a particular bowl or spoon),
our goal will be to characterize it in a form that enables a {\sc tamp} planner to deploy it in combination with existing skills.  To do this, we need to formally describe the
intended effects of the new skill as well as conditions on the state
in which the skill is initiated that would guarantee that the intended
effect occurs.  For example, the intended effect of a pouring skill is
that liquid be in some destination container, and the precondition of
that effect is that the robot is holding some other container
that has liquid in it.

There are three important constraints on the process of learning the
preconditions and effects of a skill:
\begin{enumerate}
   \item \highlight{Learning should characterize} a {\em comprehensive} set of control parameter values instead of a single value. The predicted values are subjected to downstream constraints, for example, arising from robot kinematics and collision avoidance. There might not be any robot motion that satisfies these for an individual control value.
  \item The result of learning should have {\em quantified uncertainty}:
    that is, we should be able to characterize possible starting
    states for skill execution in terms of how sure we are that the
    intended effect will occur.  Knowing this will allow the {\sc tamp} planner, as far as possible given the problem-solving
    context, to use the skill in a way that it is confident will succeed.
  \item Learning should be {\em sample efficient}:
    that is, it should require relatively few trial executions of the
    skill in different situations 
    to acquire the models needed for planning.  This is critical
    because of the high cost of running trials on a physical robot:
    not only must the robot execute the skill on each trial, it must
    set up the initial conditions appropriately for the next. 
\end{enumerate}

\begin{figure}
\centering
\includegraphics[width=\columnwidth]{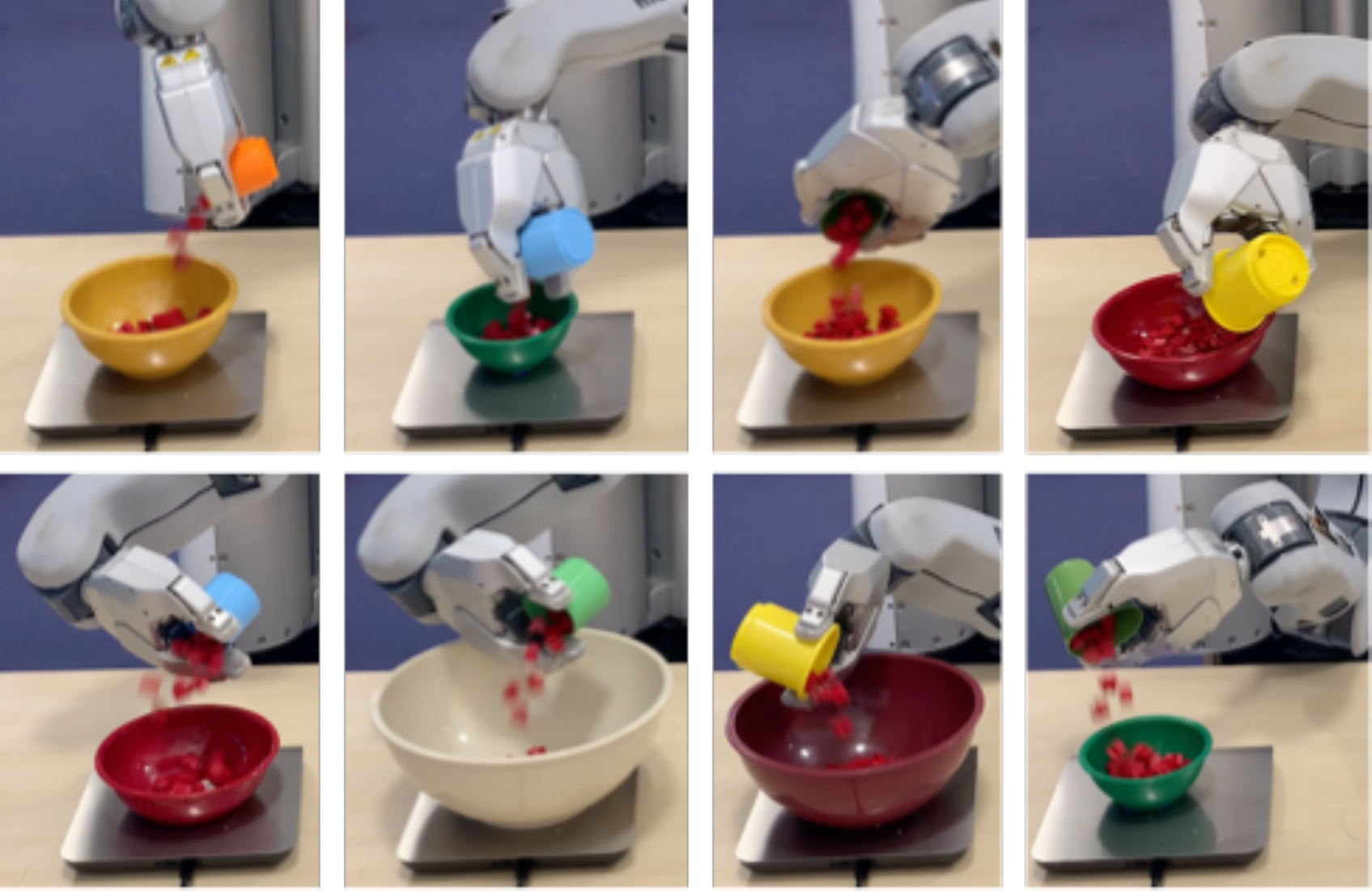}
\caption{Examples of a real-world robot executing a trained pouring primitive in {\em KitchenPR2} for several contexts parameter values (cup dimensions) and control parameters values (relative cup poses).}
\label{fig:pouring}
\end{figure}

\begin{figure}
\centering
\includegraphics[width=\columnwidth]{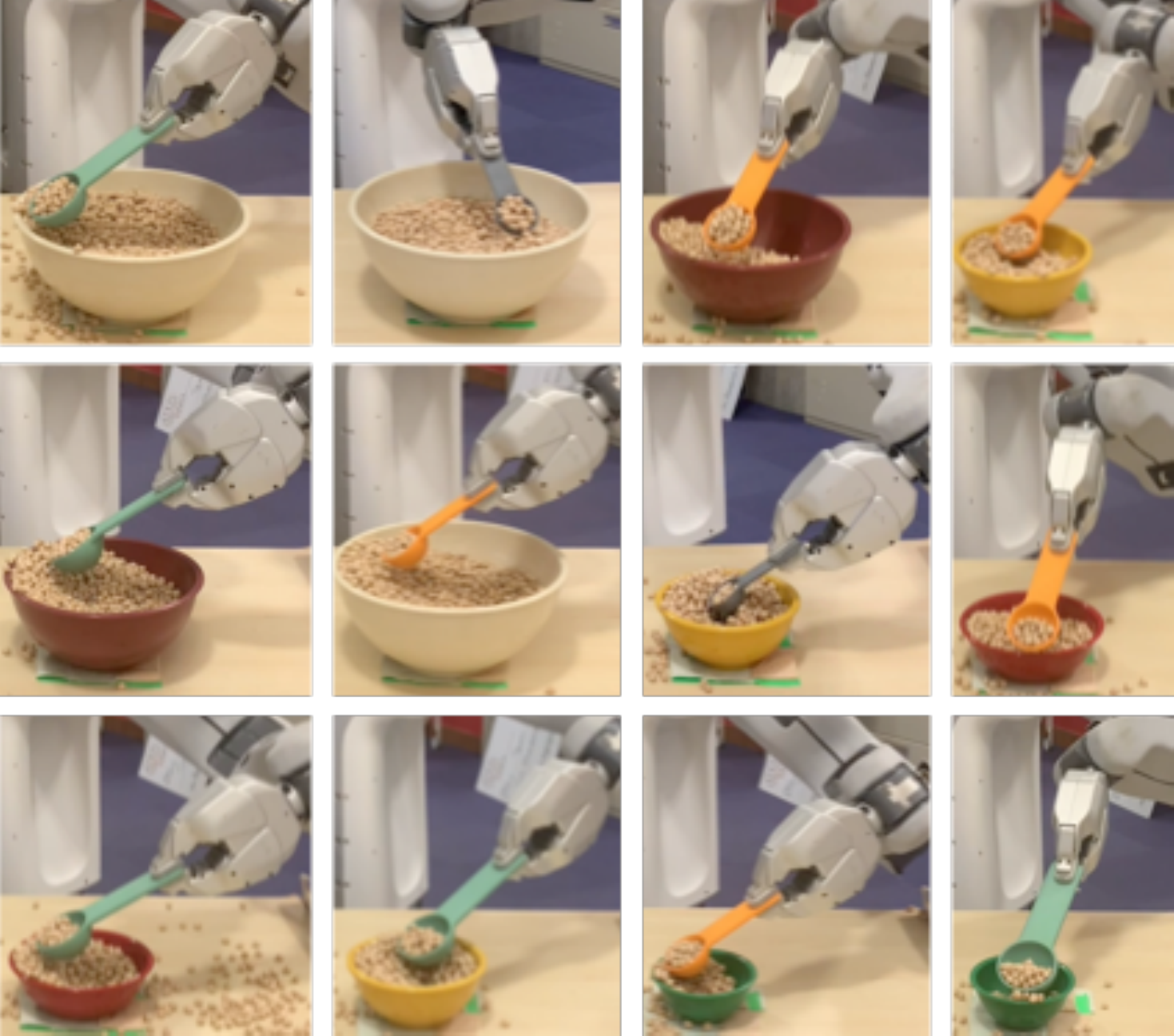}
\caption{Examples of a real-world robot executing a trained scooping primitive in {\em KitchenPR2} for several contexts (bowl and spoon dimensions) and control parameters (relative spoon poses).}
\label{fig:scooping}
\end{figure}

Figure~\ref{fig:pouring} illustrates several instances of a
parameterized sensorimotor policy for pouring with a real-world PR2.
\highlight{Context parameters for the skill encode the approximate dimensions of both the cup and bowl.}
Control parameters specify the cup's rotation about a coordinate frame, the final pitch of the cup in this frame, and the pose of this frame relative to the bowl.
See figure~\ref{fig:pour-parameters} for a visualization of these parameters.
Figure~\ref{fig:scooping} displays the robot executing a sensorimotor policy for scooping.
The objective of our work is to learn the set of sufficiently successful pours and scoops across a wide range of objects.

This paper is an extended version of~\cite{Wang2018ActivePlanning}.
\highlight{Our new contributions} include an in-depth discussion of how the learners, motion planners, and {\sc tamp} planner interact in our integrated system (section~\ref{sec:formulation} and appendix~\ref{sec:pddlstream}),
the application of our approach to a high-dimensional robot manipulator operating in a 3D simulated dynamic manipulation environment (section~\ref{sec:domains}), 
extensive experimentation within this environment that compares different learners, learning strategies, and sampling strategies  (section~\ref{sec:experiments}),
and finally real-world validation of the efficacy of learning
individual primitives and deployment of the full system to solve multi-step manipulation problems (section~\ref{sec:real-world}).

In the rest of this paper, we 1) formulate a precise learning problem and
explore algorithms based on Gaussian processes for efficiently
learning models for and robustly applying robot skills (section~\ref{sec:estpre}); 2) formalize
the overall problem of generating behavior involving these new skills
using the \stripstream{} {\sc tamp} planner~\citep{garrett2020PDDLStream} (section~\ref{sec:planning}); and finally
3) present extensive empirical results both in
physics simulations (sections~\ref{sec:domains} and~\ref{sec:experiments}) and on a real robot (section~\ref{sec:real-world}), demonstrating efficient model
acquisition and robust use of new skills in complex problems.

%% file: formulation.tex
\section{\highlight{Approach}}
\label{sec:formulation} 


\hide{
We begin by describing the high-level design choices, subproblems, and components of our integrated learning and planning system.

\subsection{Task and motion planning}

We take a model-based planning approach to robotic manipulation. 
When a reliable model is available, planning approaches often outperform model-free control approaches due to their compositionality and generality.
Planning fluidly supports modularly reasoning using diverse action primitives such picking, pushing, pouring, and scooping.
Additionally, planning is able to generalize across a vast set of problems involving varying object shapes, sizes, affordances, and initial states as well as varying goal conditions.

Because manipulation problems involve hybrid state variables, hybrid action parameters, and continuous robot motions, they are well-modeled as {\sc tamp} problems.
{\sc tamp} problems are specified by a hybrid initial state, set of parameterized actions, and a goal condition.
In manipulation environments with many objects, the state of a particular object often only significantly changes when the robot acts on it.
As a result, action effects are {\em sparse} because only a small fraction of the state variables are affected by the action.
When this is true, actions can be modeled more compactly, {\it i.e.} using fewer parameters, by describing the difference between current and subsequent state. 
To ensure that an action will be correctly (and usefully) applied in a state, the state must satisfy an action's {\em precondition}, which often is a logical conjunction over individual conditions.
Upon application, the action's {\em effect} specifies the new value of any variable that is changed.
Finally, each action has a {\em constraint} defined on its action parameters that parameter values must satisfy in order to perform the action.

Action descriptions have free parameters, so a {\sc tamp} planner must ultimately assign values to these parameters.
Because some action parameters are continuous, there generally are an infinite number possible parameter values. 
One strategy for generating candidate values, is to specify {\em samplers}, which enumerate an infinite sequence of values. 
In order to produce parameter values that jointly satisfy an action constraint with values given in, for example, the initial state, samplers must be {\em conditional}, meaning they take some values an {\em input} and, conditioned on those values, generate {\em output} values that together with the input values satisfy a constraint.
Through compositionality, the output values of a sampler can serve as input values to another sampler, enabling a planner to generate values that collectively satisfy a constraint with many parameters.
Some samplers that are useful for robot manipulation include placement and grasp samplers, inverse kinematic samplers, and motion planners.

\subsection{Modeling a new skill}


Critically, planning requires a model that specifies the structure of each action, the constraint on its parameter values, and a method for generating parameter values that satisfy this constraint.
We will assume that the structure of each action is given, but see work by \cite{Xia} for one strategy for learning the structure.
Some constraints, such as geometric constraints involving kinematics and collisions, are well understood and can be straightforwardly specified by a roboticist.
Furthermore, there often are a repertoire of existing techniques for satisfying these constraints, such as inverse kinematics solvers for kinematic constraints.
However, other constraints, such as the constraint that a pour or scoop is sufficiently successful, are less well understood due to the complex dynamics of the entities involved.
Thus, we seek to use machine learning to estimate these constraints from real-world data.
Namely, we learn a classifier for whether a tuple of action parameter values satisfies the constraint.

Planning can be harmful when the model is substantially inaccurate.
For example, a poor pour prediction over porous planes leaves PhDs poor.
As a result, we seek to be risk-adverse and avoid taking actions with effects that we are not sufficiently confident in.
This requires modeling our own uncertainly in our predictions, which obtain through using Gaussian Processes.
Uncertainty quantification is not just useful for safety but also can be used to expedite the learning process.
More specifically, active learning that selects the next training example using the current model's uncertainty can identify regions of the parameter space that would benefit most from another training example.

Once we have learned a constraint we need to provide a sampler that produces values that are likely to satisfy it.
Again, we need samples that are useful given the object properties in the current problem instance.
Thus, we construct conditional samplers by partitioning the parameters into {\em context} (input) and {\em control} (output) parameters where intuitively context parameters are properties of objects, such as their dimensions, and control parameters are commands taken by the robot.
Conditioned on context parameter values, a tempting strategy is to select a sample that maximizes the likelihood of satisfying constraint; however, this strategy is insufficient for planning.
Sampled values for action must not only respect the action's constraints but also constraints arising from the actions preceding and succeeding the action.
These additional plan constraints may cause any particular action parameter assignment to be not viable.
For example, a high quality pour action may not be performable if the initial robot configuration for the pour is not reachable via move action.
Thus, we must sample from the set of {\em all} sufficiently likely-to-succeed parameter values to ensure that the planner has many options in case its initial candidates fail to admit a full plan.

The simplest strategy for sampling this set is to apply rejection sampling, namely sampling the parameter space uniformly at random and yielding only the samples that satisfy the learned constraint.
But when the set of satisfying values has a much smaller volume than the full parameter space, vanilla rejection sampling can be computationally burdensome.
Instead, we adopt an {\em adaptive} strategy that biases the sampling distribution towards control parameter values that previously were successful through sampling from a Gaussian Mixture Model defined on these points.

The sampler has the freedom to select the order in which samples are considered and the chosen ordering affects the overall planning time.
For example, parameter values that numerically are almost identical likely have similar impacts on other constraints under the assumption that constraints in general are locally often smooth.
Under this assumption, the viability of these samples is correlated.
Intuitively, we would instead prefer to propose samples that are less correlated with previously attempted samples for which we were unable to incorporate into a plan.
To instead encourage variety with respect to sample viability, we first define {\em diversity} using determinantal point processes ({\sc DPPs}). 
Then, we apply greedy submodular function optimization to select among a set of candidate samples the most diverse sample given our prior choices.

\subsection{Integration}

Ultimately, we take a middle ground between engineering and learning models.
Constraints vary in how they are most effectively modeled, so we introduce a system that can make the best of both in combination.
For example, a spoon scooping path generated by a learned sampler can be the input to a robot motion planner that solves for robot joint trajectories that realize this tool movement.
From an engineering perspective, only a characterization of the component, whether it be a conventional algorithm or a learned solution, is needed to consider it in the context of other actions.
}


\hide{


Section~\ref{sec:estpre}

Section~\ref{sec:planning}

Section~\ref{sec:activeLearn}

Section~\ref{ssec:adaptive}

Section~\ref{sec:adaptive}

Section~\ref{ssec:diverse}

Section~\ref{sec:related}

Section~\ref{sec:domains}

Section~\ref{sec:experiments}

Section~\ref{sec:real-world}

Appendix~\ref{sec:pddlstream}
}




We model the {\sc tamp} problem as one of controlling a robot operating in a deterministic discrete-time hybrid system.
The state $s$ of the system is comprised of a set of discrete and continuous state variables, which describe the properties and configuration of the robot as well as the objects in the environment.
At each time step, the robot executes an action $a$, which corresponds to applying low-level motor torques.
Let ${\cal S}$ be the state space of the system and ${\cal A}$ be the action space of the robot.
The initial state of the system is $s_0 \in {\cal S}$, and the objective of the robot is to control the system to a state $s_*$ contained with a specified set of goal states $s_* \in S_* \subseteq {\cal S}$.
Let ${\cal T}: {\cal S} \times {\cal A} \to {\cal S}$ be the system's transition function. 

We assume that a set of parameterized skills ${\cal O}$ has already been programmed or learned, and it is our objective to learn a characterization of each skill that can be used by a task-and-motion planner.
Each skill $O(\omega) \in {\cal O}$ is specified as an {\em option}~\citep{sutton1999between} with initiation set $I_{O(\omega)} \subseteq {\cal S}$, policy $\pi_{O(\omega)}: {\cal S} \to {\cal A}$, 
and termination set $G_{O(\omega)} \subseteq {\cal S}$ where $\Omega_O$ is the parameter space for skill $O$ and $\omega \in \Omega_O$ is a particular parameter value.
Let $\Gamma_{O}: {\cal S} \times \Omega_O \to {\cal S}$ be the option transition function for skill $O$ where $\Gamma_O(s, \omega) = s'$ if and only if executing option $O(\omega)$ from state $s \in I_{O(\omega)}$ terminates in state $s' \in G_{O(\omega)}$, which is the result of recursively applying $s \gets {\cal T}(s, \pi_{O(\omega)}(s))$ until reaching a termination state $s \in G_{O(\omega)}$.
Controlling the system can now be viewed as a planning problem over skill instances where the objective is to find a finite sequence of $k$ skill instances $O_1(\omega_1), ..., O_k(\omega_k)$ such that the corresponding sequence of states $s_0, s_1, ..., s_k$ satisfies $\Gamma_{O_i}(s_{i-1}, \omega_i) = s_i$ for $i \in \{1, ..., k\}$ as well as $s_k \in S_*$.
Importantly, we consider the setting in which the robot {\em does not} have full knowledge of $\Gamma_O$, {\it i.e.} it does not have have a complete model of the effects of skill $O$.
Without a model, the robot is unable to plan over sequences of different skill instances.
Thus, we seek to learn $\Gamma_O$ for each skill $O$ from data in order to combine and apply them during planning.

We learn $\Gamma_O$ through estimating a constraint $\chi_O: {\cal S} \times \Omega_O \times {\cal S} \to \{0, 1\}$ where $\chi_O(s, \omega, s') = 1$ if and only if $\Gamma_O(s, \omega) = s'$.
We will learn from training examples in the form of $(s, \omega, s')$ triplets, which represent a present state $s$, skill parameter value $\omega$, and future state $s'$.
In many systems, $\chi_O$ can be naturally expressed as a conjunction defined over a set of atomic constraints, each of which might only involve a small subset of the state variables and skill parameters present within $(s, \omega, s')$.
Additionally, we often have at least a partial model of $\Gamma_O$, namely we might know the analytic form of some constraints, such as geometric constraints involving kinematics and collisions, and thus need not redundantly learn these constraints from scratch.
Finally, we will assume that the structure of each atomic constraint is given, meaning that we know which state variables and skill parameters are relevant for predicting the effects of the skill (but see work by \cite{Xia} for one strategy for learning the structure).

As an example, suppose the robot is given a skill whose intended effect is to pour the contents of a cup into a bowl, and we would like to learn the conditions under which executing that skill will transfer a sufficiently large fraction of the cup's initial contents into the target bowl.
These conditions can be articulated as a constraint representing a relation among the initial poses of the cup and bowl, some aspects of their shapes, and the trajectory of the cup, defined in terms of waypoints and a final pose relative to the bowl.
During planning, some of these state variable values, such as the dimensions of objects, are given by the problem and thus cannot be chosen by the robot.
Thus, it is convenient to define constraint $\chi$ generically on $(\theta, \alpha)$ parameter pairs instead of $(s, \omega, s')$ triplets where $\alpha$ are fixed {\em context} parameters and $\theta$ are {\em control} parameters that the robot can choose.  
These control parameters can include both parameters $\omega$ passed directly to the skill as well as additional aspects of the state $s$ that the robot can control indirectly through the execution of other skills.
It is important to note that the learned constraint does not apply directly to robot configurations: all operations will include default constraints on kinematic path existence and lack of collision that are based on the system's prior knowledge of robot motion.   This means that the robot does not have to re-learn this foundational knowledge every time it acquires a new skill.

Once the set of constraints for each skill $O$ are determined, a task-and-motion planner can
construct a plan in the form of a sequence of these skill instances that achieves a desired goal condition.  In order for a plan to be correct, a planner must ensure that the accompanying sequence of parameters and induced sequence of states satisfies the constraint $\chi_O$ for each skill $O$.
Namely, a sequence of $k$ skills $O_1, ..., O_k$ has an associated alternating sequence of states and skill parameters $s_0, \omega_1, s_1, ..., s_{k-1}, \omega_k, s_k$.
Each contiguous triplet $(s_{i-1}, \omega_i, s_i)$ must satisfy $\chi_{O_i}(s_{i-1}, \omega_i, s_i) = 1$. 
A planner must search over sequences of skills as well as parameter values that satisfy these constraints.
Finding values that satisfy these constraints is a nontrivial problem; however, existing work in {\sc tamp} has shown that a variety of methods can be effective~\citep{Garrett2021}.
In this work we take a sampling-based approach using the PDDLStream {\sc tamp} planner~\citep{garrett2020PDDLStream}, which we describe in appendix~\ref{sec:pddlstream}.


\hide{
We assume a skill $\pi_\tau: {\cal S} \to {\cal A}$ that is parameterized by parameter $\tau \in {\cal T}$ has already been learned or programmed, and it is our objective to learn a characterization of it
that can be used by a task-and-motion planner.
A parameterized skill will, in general, be a policy that runs for some time and then terminates.
Namely, we wish to learn the set of triplets $(s, \tau, s')$ where $s \in {\cal S}$ is a state, $\tau \in {\cal T}$ is a policy parameter, and $s' \in {\cal S}$ is the resulting state after executing policy $\pi_\tau$ starting from state $s$.
Many quasi-static task and motion planning domains are highly underactuated due to the fact that the robot only changes the state of the objects it acts upon.
As a result, the set of skill transitions $(s, \tau, s')$ is generally a low-dimensional submanifold ${\cal M}_\pi$ of ${\cal S} \times {\cal T} \times {\cal S}$.

To cope with this during planning, task-and-motion planners typically reason in 
a lower-dimensional space $\Omega_\pi$ that parameterizes ${\cal M}_\pi$.
The set of legal transitions is often described as a set of individual equality and inequality constraints defined on $\Omega_\pi$, for example involving kinematics constraints, collision constraints, and tool constraints.
In this work, we consider the setting where an analytical form for some of these constraints is not known {\it a prior}, so we instead use statistical learning to estimate these constraint from data.
Let $\chi: \Omega_\pi \to \{0, 1\}$ denote a constraint that we wish to learn.
Ultimately, we begin with a {\em partial} model of the system's dynamics and apply learning to complete to the model for use during planning.

Given a set of known and learned constraints describing legal transitions of a skill $\pi$, a task-and-motion planner must produce parameter values $\omega \in \Omega_\pi$ that satisfy these constraints.
In this work, we take a sampling-based approach to generating satisfying values.
In order to produce parameter values that jointly satisfy an action constraint with values present in, for example, the initial state, samplers must be {\em conditional}, meaning they take some values an {\em input} and, conditioned on those values, generate {\em output} values that together with the input values satisfy a constraint~\citep{GarrettRSS17, garrettIJRR2018}.
Through compositionality, the output values of a sampler can serve as input values to another sampler, enabling a planner to generate values that collectively satisfy a set of constraints defined on many parameters.
Along the same lines, we require conditional samplers for each learned constraint $\chi$.
Namely, we partition parameters $\omega = (\alpha, \theta)$ into {\em context} (input) parameters $\alpha$ and {\em control} (output) parameters $\theta$ where intuitively context parameters are properties of objects, such as their dimensions, and control parameters are commands taken by the robot.
During planning, our learned sampler will consume values for context parameters $\alpha$ and sample values for control parameters $\theta$ such that, with high confidence, $\chi(\theta, \alpha) = 1$.
Ultimately, sampled control parameter values will become inputs to other conditional samplers that perform operations such as full-body robot motion planning.
Because of this, learned samplers may need produce many control parameter values in the event that a downstream sampler fails to produce any values conditioned on the predictions.
In the next two sections, we focus on learning these constraints and then on sampling from them for use during planning.
}


\hide{
\caelan{@TLPK - consider revising according to the reviewer feedback. Looking over it again, it is a little disconnected from the subsequent sections. For example, $\theta, \phi, \psi$ mean something else (and are extensively used) in the later sections}
\zw{and $T$ was used for number of samples later.}

We assume that a parameterized skill $\pi_A(\omega_a)$ has already been learned or programmed, and it is our objective to learn a characterization of its preconditions and effects that can be used by a task-and-motion planner.
A parameterized skill will, in general, be a policy that runs for some time and then terminates, potentially yielding an observation characterizing its result. 
\caelan{Do you mean the effects, score, or something else?}

For the purposes of planning, we describe states of the world in terms of {\em conditions} characterizing particular aspects, such as the pose of an object or a relationship between objects, for example, that one object is on top of another.
Using this same vocabulary of conditions, we formulate {\em operator} descriptions to describe the preconditions and effects of parameterized skills, so that we can use a {\sc tamp} planning algorithm to construct sequences of actions, characterized as instances of the available operators.

In our domains of interest, operator descriptions are ``lifted'' in the sense that they are parameterized by the particular objects they are intended to operate on.
An operator description comprises:
\begin{itemize}
\item a set of {\em variables} naming objects over which the operation is lifted; 
\item a {\em skill}, with parameters that govern the details of its operation ({\it e.g.}, how far to move or how hard to grasp);
\item a primary {\em result condition}, which typically names some objects and parameter values;
\item a {\em precondition}, in the form of a conjunction of conditions defined on the variables and possibly introducing additional parameters $\omega_p$;
\item a possibly complex {\em constraint} on the values of all the parameters that occur in the result, the skill, and the preconditions; and
\item a {\em sampler} that, given some values of the variables in the constraint, produces a stream of possible value for the other variables, such that the variable values collectively satisfy the constraint.
\end{itemize}

The learning problem we address is: given a parameterized skill $\pi_A(\omega_a)$ and effect condition $\phi(O, \omega_e)$, where $\omega_a$ and $\omega_e$ are parameter vectors and $O$ is a vector of object names, learn an operator description with parameterized action $\pi_A(\omega_a)$, effect $\phi(O, \omega_e)$, and precondition $\psi(O, \omega_p)$, that are correct and as weak as possible.
An operator description is {\em correct} if its precondition constraint (viewed as a set of world states in which it holds) is a subset of the true precondition constraint;  a precondition constraint representing a larger set is weaker than one representing a smaller set.  

We will learn such operator descriptions from training examples of the form: $(s, a, s')$.
The $s$ and $s'$ are detailed descriptions of the world state before and after the action is executed and $a$ is a parameterized instance of the operator with known parameters.
We will assume that the structure of the operator description is given (but see work by \cite{Xia} for one strategy for learning the structure), meaning that we know which objects and their properties are relevant for predicting the effects of the operation.
Our focus, then, is on learning a constraint on the real-valued parameters that guarantees the effectiveness of an action.

Learning the constraint in the preconditions can be framed as a classification problem, mapping from a vector of values of the parameters $\omega = (\omega_p, \omega_a, \omega_e)$ characterizing the preconditions, action, and effect to a Boolean value.
However, we will want to use the resulting classifier in a non-standard way.
To implement the sampler for an operator, we need to be able to, given bindings of some of the parameters, sample from the set of legal bindings of the other parameters.
This is because we may need to consider many different instances of this operator when planning, in case some states in the preconditions of this operator are not achievable, due to obstacles or other constraints present in the planning problem.  

For this same reason, it is not sufficient simply to predict a single ``best" $\omega$ given some of the parameters, since that value may not be feasible due to collisions or kinematics.
Furthermore, it is critical to quantify our uncertainty about which settings of $\omega$ will result in a successful operation, and to learn it from as little data as possible.
For these reasons, we will take a Bayesian approach, formulating the classification problem instead as a regression from $\omega$ to a continuous {\em score} of how well the operation worked, and using Gaussian process~\citep{rasmussen2006gaussian} (\gp{}) regression to characterize our uncertainty, for use during data acquisition and during planning.
In the next two sections, we focus on learning the constraint, and then on using it during planning.
}


\hide{
\caelan{What follows is my initial attempt to more precisely set up the problem. Maybe some of the ideas are useful. I think we shouldn't stress the PDDL-like form except for lifting/parameterization of operators}

We consider the problem of controlling an agent acting in a deterministic and observable environment to achieve a specified goal.
The state of the agent and the environment is $s \in {\cal S}$ where ${\cal S}$ is the state space, the set of all possible states.
The agent sequentially takes actions $a \in {\cal A}$ where ${\cal A}$ is the action space, the set of all possible commands.
Upon taking action $a$ from state $s$, the state is now $s' = T(s, a)$ where $T: {\cal S} \times {\cal A}_A \to {\cal S}$ is a deterministic transition function.
The goal to control the world to be at a state $s_*$ within a set of goal states $S_* \subseteq {\cal S}$.

We seek to control the agent through planning by repeated applying the transition function $T(s, a)$ for different $a$ to explore possible futures.
The challenge is that the transition function is only {\em partially} known, in that we know its structure in terms of its parametric form and the components of the state (state variables) it affects, as well as possibly the resulting value for some of these variables; however, some of these effects are unspecified.
We aim to use statistical learning to produce these unspecified effects. 

We are adverse to planning with an woefully incorrect $T(s, a)$ predictions and would rather intentionally prevent the agent from reasoning about taking action $a$ in state $s$. 
Additionally, many actions $a$ are prohibitively costly or unproductive when executed from $s$.
Thus, we seek to learn $T(s, a)$ for a subset $D \subseteq {\cal S} \times {\cal A}$ of the {\em domain} of $T$, where we get to choose $D$.
We would like to maximize the size of $D$ subject to the constraint that our predictions $T(s, a)$ are sufficiently accurate.

Rather than learn the functional relationship $T(s, a)$ directly, we learn the {\em graph} of $T$ over the subset $D$
\begin{equation*}
    G_D = \{(s, a, s') \mid (s, a) \in D, s' = T(s, a)\}.
\end{equation*}
This results in a classification problem $\chi: D \times {\cal S} \to \{0, 1\}$,
enabling us to take a Bayesian approach where we reason about the likelihood that this transition is correct, and on top of that, our own uncertainty in this prediction. 

The state spaces and action spaces that we wish to operate mixed discrete-continuous ({\em hybrid}) and are very high-dimensional.
To cope with these challenges during learning and planning, we apply ideas from AI planning by factoring states into a collection of variables and describing transition $(s, a, s')$ triplets using operators $o$, which describe sets of transitions.
Furthermore, we consider {\em lifted} (parameterized) operators $O(\omega)$ where $o = O(\omega)$ for parameter values $\omega$.
Thus, our objective is to learn a set of parameter values $\Omega$ that encodes $G_D$. 
Each lifted operator is endowed with a {\em precondition} that encodes $D_O$, here expressed as a logical formula, an {\em effect} that describes $T(s, a)$, which specify using a logical conjunction, a {\em skill} that executes $a$, and a learned {\em constraint} that characterizes $\Omega$.

w.r.t sampling
\begin{itemize}
    \item Some parameter values appear in the state/goal
    \item Sampling these values randomly is ineffective
    \item Need to condition on them and produce completing values
    \item Outputs of learner are inputs to a motion planner that fully-parameterizes the operator
\end{itemize}

We learn from $(s, a, s')$ instances.
We score based on $s'$ whether $(s, a)$ should be a component of $D_T$.
}


%% file: estimation.tex
\newcommand{\context}{\alpha}
\newcommand{\free}{\theta} 
\newcommand{\dcontext}{d_\alpha}
\newcommand{\dfree}{d_\theta}


\section{\highlight{Estimating the constraint}} 
\label{sec:estpre}


Our primary technical problem, then, is to learn a constraint representing the success criteria for executing a skill, represented as a relation among fixed context parameters $\context$ and free control parameters $\free$.  For reasons outlined in the introduction, we seek to characterize the entire space of successful control parameters for any given context, using an explicit characterization of uncertainty in the learned relation to guarantee robust parameter selection at planning time.

\subsection{\highlight{Contextual super-level set estimation}}

We will focus on the formal problem of learning a function from values of the 
context parameters $\context\in \mathbb{R}^{\dcontext}$ to sets of control parameters $\free\in B$.
We assume that the domain of $\free$ is a
hyper-rectangular space ${B=[0,1]^{\dfree}\subset \mathbb{R}^{\dfree}}$,
but generalization to other topologies is possible.
We are interested in learning a Boolean function $\chi: B \times \mathbb{R}^{\dcontext} \to \{0, 1\}$ for a skill of interest, 
where $\chi(\free , \context) = 1$ if and only if executing the skill 
with control parameter $\free$ and context parameter $\context$ results in the desired effect. 

We assume that $\chi$ can be expressed in the 
form of an inequality constraint $\chi(\free,  \context) = \indicator{g(\free,
\context) > 0}$, where ${g: B \times \mathbb{R}^{\dcontext} \to \R}$ is a real-valued scoring function with arguments
$\free$ and $\context$. We denote the conditional {\em super-level set} of the scoring
function given $\context$ by 
\begin{equation*}
    A_{\context} \equiv \{\free \in B\mid g(\free, \context) > 0\}.
\end{equation*}
For example, the scoring function $g(\free, \context)$ for pouring might
be the proportion of poured liquid that actually ends up in the target
cup, minus some target proportion.  So, given any new situation with context parameters $\alpha$, we know that any value of control parameters $\theta \in A_{\context}$ will result in success with high probability.  This strategy relies on the 
availability of real-valued values of this score function during
training rather than just binary labels of success or failure.

\subsection{Active sampling for learning}
\label{sec:activeLearn}

Our objective in the learning phase is to efficiently gather data
to characterize the conditional super-level sets $A_{\context}$ with high
confidence.  We use a Gaussian process (\gp{}) on the score function $g$ to select
informative queries using a level-set estimation 
approach.
In order to implement the constraint sampler, we must be able to sample from the super-level set $A_{\context}$
for any given context $\context$
During training, we select $\context$ values from a
distribution reflecting naturally occurring contexts in the underlying
domain, for example, the dimensions of cups and bowls in a pouring operation.
In the event that the agent can initialize its environment, for example by picking the objects for an experiment, some ``context parameters'' can be viewed as control parameters that can be selected in the process of active learning.
Note that learning an accurate description of the boundaries of the level-set is
a different objective from learning all of the function values
well and also different from finding the maximum function value, and so it must be handled differently from typical \gp{}-based active learning.

For each $\context$ value in the training set, we apply the {\em
  straddle} algorithm~\citep{bryan2006active} to actively select
samples of $\free$ for evaluation by running the skill policy.
After each new evaluation of $g(\free, \context)$ is obtained, the
data-set $\mathcal D$ is augmented with pair
$\langle (\free, \context), g(\free, \context) \rangle$, and used to
update the \gp{}.  Given the mean function $\mu(\cdot)$ and the variance function $\sigma^2(\cdot)$ for the posterior \gp{}, the straddle algorithm selects $\free$ that maximizes the {\em acquisition function}
\begin{equation*}
\psi_{\mu, \sigma}(\free, \context) = -|\mu (\free,\context)|+1.96\sigma(\free,\context).
\end{equation*}

It has a high value for values of $\free$ that
are near the zero boundary for the given $\context$ or for which the score
function is highly uncertain.  
The parameter $1.96$ is selected such that if $\psi_{\mu, \sigma}(\free, \context)$ is negative, $\free$ has less than 5 percent chance of being in the level set. 
In practice, this heuristic has
been observed to deliver state-of-the-art learning performance for
level set estimation~\citep{bogunovic2016truncated,gotovos2013active}.
After each new evaluation, we retrain the Gaussian process by
maximizing its marginal data-likelihood with respect to its
hyper-parameters.  Algorithm~\ref{alg:bo} specifies the algorithm;
$\proc{gp-predict}({\cal D})$ computes the posterior mean and variance, which is explained in appendix~\ref{ssec:gp}.

\begin{algorithm}[H]
  \begin{small}
  \caption{Active Bayesian Level Set Estimation}\label{alg:bo}
  \begin{algorithmic}[1]
    \State Given initial data set $\mathcal D$, context $\alpha$,
    number of samples $T$ 
      \For{$t \in \{1, ..., T\}$}
      \State $\mu, \sigma$ $\gets$ \proc{gp-predict}($\cd$)
      \State $\theta \gets \argmax_{\theta}{\psi_{\mu, \sigma}(\theta, \alpha)}$ 
      \State $y\gets g(\theta, \alpha)$ 
      \State $\cd \gets \cd \cup \{\langle (\theta,\alpha), y\rangle\}$
      \EndFor
     \State \Return{$\mathcal D$}
  \end{algorithmic}
  \end{small}
\end{algorithm}

%% file: planning.tex
\section{Planning with a new skill} 
\label{sec:planning}


We have shown how to take a controller for a new motor skill and use active-learning strategies to estimate a constraint representing the conditions under which executing that skill will have a desired effect.   In this section, we describe our strategies for integrating that new skill into a {\sc tamp} system.  

\subsection{The need for sampling during planning}
Planning for {\sc tamp} problems is difficult, because it requires integrating aspects of motion planning through continuous robot configuration space, AI-style planning through discrete choices of operations and objects, and the selection of real-valued parameters, such as object grasps and placements as well as robot configurations that enable the execution of manipulation operations.

In our work, we use the \stripstream{} planning framework~\citep{garrett2020PDDLStream}, which is discussed in more detail in appendix~\ref{sec:pddlstream}.  
In this framework, a skill description must specify the constraint on parameter values and a {\em sampler} that can, given values of context parameters $\context$, produce a stream of assignments to the control parameters $\free$. 
In this section, we focus on the construction and use of this sampler.

The reason for sampling values of $\free$, rather than simply selecting the one that maximizes the likelihood of success given $\context$, is that there may be other considerations that make $\free$ infeasible in broader planning context.  For example, a particular grasp of the cup to be poured from might acceptable for pouring, but unreachable for the robot given the current placement of the object on the table.



Our objective in the planning phase is to select
a diverse set of samples $\{\free_i\}$ for which it is likely that
$(\context, \free_i)$ \highlight{satisfy both the learned constraint $\chi$ and the rest of the constraints in the planner.} We do this in two steps:  first, we use a
novel risk-aware sampler to generate $\free$ values 
that satisfy the learned constraint with high probability; second, we
integrate this sampler with \stripstream{}, where we generate
samples from this set that represent its diversity, in order to 
expose the full variety of choices to the planner.

\subsection{Risk-aware sampling} 
\label{ssec:adaptive}
We can use our Bayesian estimate of the scoring function
$g$ to select action instances for planning.  Given a
new context $\context$, which need not have occurred in the training
set---the \gp{} will provide generalization over contexts---we would
like to sample a sequence of $\theta\in B$ such that with high
probability, $g(\theta, \context) > 0$. In order to guarantee this, we
adopt a concentration bound and a union bound on the predictive scores
of the samples. Notice that by construction of the \gp{}, the predictive
scores are Gaussian random variables.  
Letting $\phi_{\mu, \sigma}(\theta, \context)$ be the ratio of the predicted mean and standard deviation, 
\begin{equation*}
    \phi_{\mu,\sigma}(\theta, \context) = \mu(\theta, \context) / \sigma(\theta,\context), 
    \label{eqn:ratio}
\end{equation*}
the following is a direct corollary of lemma 3.2 of~\citet{wang2016est}:
\begin{cor}\label{lem:pbound}
Let $g(\theta, \context) \sim \GP(\mu, \sigma)$, and for $\delta\in(0,1)$ set $\beta^*_i = \sqrt{2\log(\pi_i / 2\delta))}$, where
$\sum_{i=1}^T \pi_i^{-1} \leq 1$, $\pi_i > 0$. \par
If $\forall  i \in \{1, ..., T\}\; \phi_{\mu, \sigma}(\theta_i, \context) > \beta^*_i$, \par
then $ {\Pr [ g(\theta_i, \context) > 0, \forall i ] \geq 1-\delta}$.
\end{cor}  
Corollary~\ref{lem:pbound} \highlight{enables us to properly construct the set of parameters that satisfy the inequality constraint $g(\theta, \context) > 0$ with high probability.} 
Here we define the {\em high-probability super-level set} of $\theta$ for context
$\alpha$ as 
\begin{equation*}
\hat A_\context=\{\theta \mid \phi_{\mu, \sigma}(\theta, \context) > \beta^*\}
\end{equation*}
where $\beta^*$ is picked according to corollary~\ref{lem:pbound}.  If we draw $T$ samples from $\hat A_\context$,
then with probability at least $1-\delta$, all of the samples will
satisfy the constraint $g(\theta, \context) > 0$.

In practice, however, for a given $\context$ and using the definition of $\beta^*$ from corollary~\ref{lem:pbound}, the set $\hat A_\context$ may be empty. 
To account for this, we relax our criterion to include the set
of $\theta$ values whose score is within 5\% of the value
of the most confident parameter, and define an alternative score threshold
$\beta = \Phi^{-1}(0.95 \Phi(
\phi_{\mu, \sigma}(\theta^*, \context))$ 
where $\Phi$ is
the cumulative density function of a normal distribution and 
$\theta^*$ is the {\em most confident} parameter, {\it i.e.} 
\begin{align*}\theta^* = \argmax_{\theta\in B}\phi_{\mu, \sigma}(\theta, \alpha).
\end{align*}
\highlight{Although we can obtain the derivatives of function} $\phi_{\mu, \sigma}(\cdot)$, we may not be able to solve the optimization problem due to the multi-modality of this function. However, we can approximate the solution to the global optimization of function $\phi_{\mu, \sigma}(\cdot)$ over domain $B$ by restarting gradient-based optimization at a few locations within domain $B$.
Alternatively, we may estimate $\theta^*$ by sampling a set of $n$ parameters $\{\theta_1, ..., \theta_n\} \in B$, and returning the value $\theta^* = \argmax_{\theta_i}\phi_{\mu, \sigma}(\theta_i, \alpha)$.

\begin{figure}
\centering
\vskip 0.1in
\includegraphics[width=1.\columnwidth]{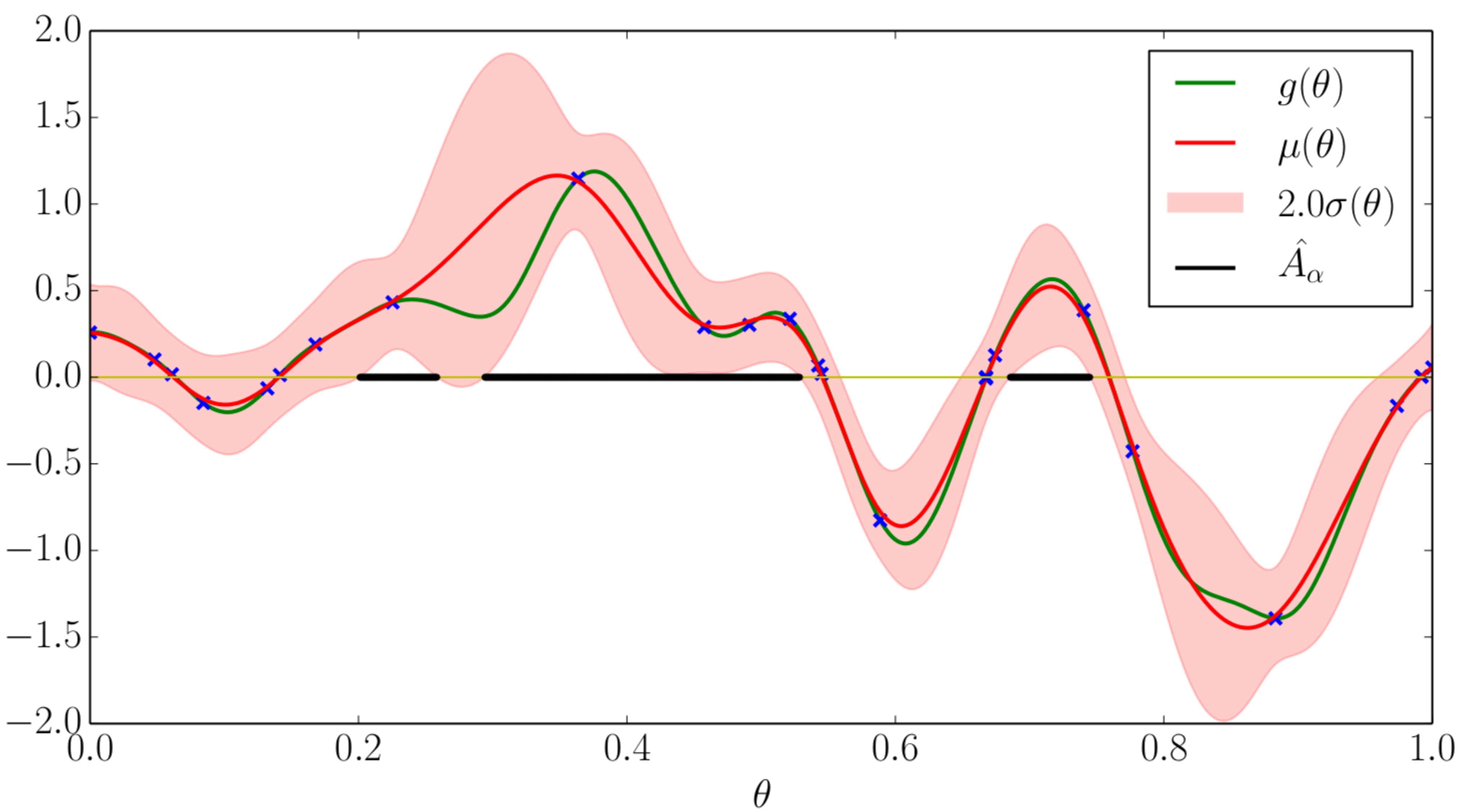}
\caption{High-probability super-level set in black.}
\label{fig:hpsls}
\vskip -0.2in
\end{figure}

Figure~\ref{fig:hpsls} illustrates the computation of $\hat A_\alpha$.
The green line is 
the true hidden $g(\theta)$; the blue $\times$ symbols are the training
data, gathered using the straddle algorithm in $[0,1]$; the red
line is the posterior mean function $\mu(\theta)$; the pink
regions show the two-standard-deviation bounds on $g(\theta)$ based
on $\sigma(\theta)$; and the black line segments are the
high-probability super-level set $\hat A_\context$ for $\beta = 2.0$.
We can see that sampling has concentrated near the boundary, that 
$\hat A_\context$ is a subset of the true super-level set, and that as
$\sigma$ decreases through experience, $\hat A_\context$ will approach
the true super-level set.



\subsection{Efficient adaptive sampling} 
\label{sec:adaptive}

To sample from $\hat A_{\context}$, one simple strategy is to do
rejection sampling with a proposal distribution that is uniform on the
search bounding-box $B$.  However, in many
cases, the feasible region of a constraint is much smaller
than $B$, which means that uniform sampling will have a very low
chance of drawing samples within $\hat A_\context$, and so rejection
sampling will be very inefficient.
We address this problem using a novel adaptive sampler, which 
draws new samples from the neighborhood of
the samples that are already known to be feasible with high probability
and then re-weights these new samples using importance 
weights. 

The algorithm \proc{AdaptiveSampler} in algorithm~\ref{alg:adaptive} takes as input the posterior \gp{}
parameters $\mu$ and $\sigma$ and context vector $\context$, and yields
a stream of samples.
It begins by computing $\beta$, then sets $\Theta_{\it init}$ to contain
the $\theta$ that is most likely to satisfy the constraint.
It then maintains a buffer $\Theta$ of at least $m/2$ samples and
yields the first one each time it is required to do so;  it
technically never actually returns, but yields 
a sample each time it is queried. 

\begin{algorithm} 
  \begin{small}
  \caption{Super-level Set Adaptive Sampling}\label{alg:adaptive}
  \begin{algorithmic}[1]
  \Function{AdaptiveSampler}{$\mu, \sigma, \context$}
  \State $\Theta \gets \emptyset$
 \State $\beta \gets \Phi^{-1}(0.95 \Phi(\max_{\theta\in B}
 \phi_{\mu, \sigma}(\theta, \context)))$
     \State $\Theta_{init}\gets \{\argmax_{\theta\in
       B} \phi_{\mu, \sigma}(\theta, \context)\}$
     \While{{\bf True}}
          \If{$|\Theta| < m/2$}
          \State $\Theta \gets \textsc{SampleBuffer}(\mu, \sigma, \context, \beta, \Theta_{init}, n, m)$
          \EndIf
          \State $\theta \gets \Theta[0]$
          \State \textbf{yield} $\theta$
          \State $\Theta \gets \Theta\setminus \{\theta\}$
          \EndWhile
  \EndFunction
  \end{algorithmic}
  \end{small}
\end{algorithm} 

The main work is done by \proc{SampleBuffer} in algorithm~\ref{alg:buffer}, which constructs a
mixture of truncated Gaussian distributions ({\sc tgmm}), specified by
mixture weights $p$, means $\Theta$, circular variance with parameter
$v$, and bounds $B$. 
Parameter $v$  indicates how far from known good $\theta$ values it is reasonable
to search;  it is increased if a large portion
of the samples from the {\sc tgmm} are accepted and decreased otherwise.
%
The algorithm iterates until it has constructed a set of at least $m$
samples from $\hat A_\context$.  It samples $n$ elements from the {\sc
  tgmm} and retains those that are in $\hat A_\context$ as $\Theta_a$.
Then, it computes ``importance weights'' $p_a$ that are inversely
related to the probability of drawing each $\theta_a \in \Theta_a$
from the current {\sc tgmm}.  This will tend to spread the mass of the
sampling distribution away from the current samples, but still keep it
concentrated in the target region.  A set of $n$ uniform
samples is drawn and filtered, again to maintain the chance of
dispersing to good regions that are far from the initialization.  The
$p$ values associated with the old $\Theta$ as well as the newly
sampled ones are concatenated and then normalized into a distribution,
the new samples added to $\Theta$, and the loop continues.  When at
least $m$ samples have been obtained, $m$
elements are sampled from $\Theta$ according to distribution $p$,
without replacement.


\begin{algorithm} 
  \begin{small}
  \caption{Sampling From a Truncated Gaussian Buffer}\label{alg:buffer}
  \begin{algorithmic}[1]
  \Function{SampleBuffer}{$\mu, \sigma, \context, \beta, \Theta_{init}$}
  \State $\Theta\gets \Theta_{init}$
  \State $v\gets [1]_{d=1}^{d_{\theta}}$; $p \gets [1]_{i=1}^{|\Theta|}$
      \While{True}
       \State $\Theta' \gets$ \proc{SampleTGMM}$(n ; p, \Theta,v, B)$
      \State $\Theta_{a} \gets\{\theta \in\Theta' \mid 
      \phi_{\mu, \sigma}(\theta, \context) > \beta\}$
      \State $p_{a} \gets 1/p_{\text{TGMM}}(\Theta_{a}; p, \Theta, v, B)$
      \State $v \gets v / 2  \;{\bf if}\; |\Theta_{a}| < |\Theta'|/2 \;{\bf else}\; 2v$ 
      
      \State $\Theta'' \gets$ \proc{SampleUniform}$(n; B)$
      \State $\Theta_{r} \gets$ $\{\theta \in\Theta'' \mid
      \phi_{\mu, \sigma}(\theta, \context) > \beta\}$
      \State $p_{r} \gets [Vol(B)]_{i=1}^{|\Theta_{r}|}$ 
      \State $p\gets$ \proc{Normalize}$([p, p_{r}, p_{a}])$ 
      \State $\Theta \gets $ $[\Theta, \Theta_{r}, \Theta_{a}]$
      \If{$|\Theta| > m$}
      \State \Return \proc{Sample}$(m; \Theta, p)$
      \EndIf
      \EndWhile
   \EndFunction
  \end{algorithmic}
  \end{small}
\end{algorithm}
It is easy to see that as $n$ goes to infinity, by sampling from the
discrete set according to the re-weighted probability, we are
essentially sampling uniformly at random from $\hat A_{\context}$.
This is because $\forall \theta \in
\Theta,\; p(\theta) \propto
\frac{1}{p_{sample}(\theta)}p_{sample}(\theta) = 1$. For
uniform sampling, $p_{sample}(\theta) = \frac{1}{Vol(B)}$, where $Vol(B)$ 
is the volume of $B$; and for
sampling from the truncated mixture of Gaussians, $p_{sample}(\theta)$
is the probability density of $\theta$. In practice, of course, $n$ is finite,
but this method is much more efficient than rejection sampling.

 \subsection{Diversity-aware sampling for planning}
\label{ssec:diverse}
Now that we have a sampler that can generate approximately
uniformly random samples within the region of values that satisfy the
constraints with high probability, we can use it inside a planning
algorithm to explore continuous action spaces. 
A planner 
may need to consider multiple
different parameterized instances of a particular action before
finding one that satisfies all the constraints in the overall context of the planning problem.
For example, some good pours may not be kinematically reachable given the robot's current configuration, so sampling a single pour might be insufficient for solving the task.

The efficiency of this planning process depends on the order in which
samples are generated.
Intuitively, when previous samples for a context parameter
have failed to contribute to a successful plan, it would be wise to
try new samples that, while still having high probability of
satisfying the constraint, are as different as possible from those that were
previously tried. 
We need, therefore, to consider diversity when generating samples; but
the precise characterization of useful diversity depends on the domain
in which the method is operating.   We address this problem by
adapting a kernel that is used in the sampling process, based on
experience in previous planning problems.


Diversity-aware sampling has been studied extensively with
determinantal point processes ({\sc
  dpp}s)~\citep{kulesza2012determinantal}. We begin with similar ideas
and adapt them to our planning domain, quantifying the diversity of a 
set of samples $S$ using the determinant of a Gram matrix
\begin{equation*}
D(S) = \log \det(\Xi^S\zeta^{-2} + \mI), 
\end{equation*}
where
$\Xi^{S}_{ij} = \xi(\theta_i, \theta_j)$ for $\theta_i,\theta_j\in
S$, $\xi$ is a covariance function, and $\zeta$ is a free parameter
(we use $\zeta=0.1$). In {\sc dpp}s, the quantity $D(S)$ can be interpreted
as the volume spanned by the feature space of the kernel
$\xi(\theta_i, \theta_j)\zeta^{-2} + \textbf{1}_{\theta_i\equiv
  \theta_j}$ assuming that $\theta_i=\theta_j \iff
i=j$. Alternatively, one can interpret the quantity $D(S)$ as the
information gain of a \gp{} when the function values on $S$ are
observed~\citep{srinivas2009gaussian}. This \gp{} has kernel $\xi$ and
observation noise $\mathcal N(0,\zeta^2)$. 
Because of the submodularity and
monotonicity of $D(\cdot)$, we can maximize $D(S)$ greedily with the
promise that
\begin{equation*}
D([\theta_i]_{i=1}^N)\geq (1-\frac1e)\max_{|S|\leq N}D(S)
\end{equation*}
$\forall N=1,2, ...$ where $\theta_i = \argmax_{\theta} D(\theta \cup
\{\theta_j\}_{j=1}^{i-1})$. In fact, maximizing $D(\theta \cup S)$ is
equivalent to maximizing
\begin{equation*}
\eta_S(\theta) = \xi(\theta,\theta) - \vxi^{S}(\theta)\T(\Xi^{S}+\zeta^2\mI)^{-1}\vxi^{S}(\theta)
\end{equation*}
 which is exactly the same as the posterior variance for a \gp{}.

The \proc{DiverseSampler} procedure is very similar in structure to the
\proc{AdaptiveSampler} procedure, but rather than
selecting an arbitrary element of $\Theta$, the buffer of good
samples, we track the set $S$ of samples that have already
been returned and select the element of $\Theta$ that is most diverse
from $S$ as the sample to yield on each iteration.  In addition, we
yield $S$ to enable kernel learning as described in
Algorithm~\ref{alg:learn}, to yield a kernel $\eta$.

\begin{algorithm}[H]
  \begin{small}
  \caption{Super-level Set Diverse Sampling}\label{alg:diverse}
  \begin{algorithmic}[1]
  \Function{DiverseSampler}{$\mu, \sigma, \context, \eta$}
  \State $\Theta \gets \emptyset$; $S\gets \emptyset$
  \State $\theta \gets \argmax_{\theta\in B} \phi_{\mu, \sigma}(\theta, \context)$
  \State $\beta \gets \lambda(\phi_{\mu, \sigma}(\theta, \context))$
      \While{planner requires samples}
      \State \textbf{yield} $\theta$, S
      \If{$|\Theta| < m/2$}
      \State $\Theta \gets \textsc{SampleBuffer}(\mu, \sigma, \context, \beta, \Theta_{init})$
      \EndIf
      \State $S\gets S\cup\{\theta\}$
      \Comment $S$ contains samples before $\theta$
      \State $\theta \gets \argmax_{\theta\in\Theta} \eta_{S} (\theta)$
      \State $\Theta \gets \Theta\setminus \{\theta\}$
      \EndWhile
  \EndFunction
  \end{algorithmic}
  \end{small}
\end{algorithm}

It is typical to learn the kernel parameters of a \gp{} or \dpp{} given
supervised training examples of function values or diverse sets, 
but those are not available in our setting;  we can only observe whether the set of samples is sufficient for the planner to identify a solution. 
We derive our notion of similarity by assuming that all samples that
fail to lead to a solution are similar.
Under this assumption, we develop an online learning
approach that adapts the kernel parameters to learn a good diversity
metric for a sequence of planning tasks.
We use the \proc{focused} algorithm (appendix~\ref{sec:incremental}) 
as our \stripstream{} planner in order to more precisely determine which sampled values failed to satisfy a downstream plan constraint for a particular plan skeleton.

We use the squared exponential kernel of the form
$\xi(\theta, \gamma; l) = \exp(-\sum_d r^2_d)$, where
$r_d = |l_d(\theta_d - \gamma_d)|$ is the rescaled ``distance''
between $\theta$ and $\gamma$ on the $d$-th feature and $l$ is the
inverse length scale. Let $\theta$ be the sample that failed and the
set of samples sampled before $\theta$ be $S$. We define the
importance of the $d$-th feature as
\[
\tau_{S}^{\theta}(d) = \xi(\theta_d, \theta_d; l_d) - \vxi^{S}(\theta_d; l_d)\T(\Xi^{S}\!+\!\zeta^2\mI)^{-1}\vxi^S(\theta_d; l_d),
\]
which is the conditional variance if we ignore the distance
contribution of all other features except the $d$-th; that is,
$\forall k\neq d, l_k=0$. Note that we keep $\Xi_i+\zeta^2\mI$ the
same for all the features so that the inverse only needs to be
computed once.

The diverse sampling procedure 
is analogous to the weighted majority
algorithm~\citep{foster1999regret} in that each feature $d$
is seen as an expert that contributes to the conditional
variance term, which measures how diverse $\theta$ is with respect to
$S$. The contribution of feature $d$ is measured by
$\tau^\theta_S(d)$. If $\theta$ was rejected by the planner,
we decrease the inverse length scale $l_d$ of feature
$d=\argmax_{d\in[d_{\theta}]}\tau_S^{\theta}(d)$ to be
$(1-\epsilon)l_d$, 
because feature $d$ contributed the most to the
decision that $\theta$ was most different
from $S$. 
 
\begin{algorithm}[H]
  \begin{small}
  \caption{Task-level Kernel Learning}\label{alg:learn}
  \begin{algorithmic}[1]
  \For{task in T}
    \State $S\gets \emptyset$
    \State $\context \gets $ current context
    \State $\mu, \sigma$ $\gets$ \proc{gp-predict}($\context$)
      \While{plan not found}
      \If{$|S| > 0$}
      \State $d\gets \argmax_{d\in[d_{\theta}]}\tau_S^{\theta}(d)$
      \State $l_d \gets  (1-\epsilon)l_d$
      \EndIf
      \State $\theta,S  \gets \textsc{DiverseSampler}(\mu, \sigma, \context, \xi(\cdot, \cdot; l))$
      \State Check if a plan exist using $\theta$
      \EndWhile
  \EndFor
  \end{algorithmic}
  \end{small}
\end{algorithm}

Algorithm~\ref{alg:learn} depicts a scenario in which the kernel is
updated during interactions with a planner; it is simplified in that it
uses a single sampler, but in our experimental applications there are
many instances of action samplers in play during a single execution of
the planner. 
Given a sequence of tasks presented to the planner, we can continue to
apply this kernel update, molding our diversity measure to the demands
of the distribution of tasks in the domain.
This simple strategy for kernel learning may lead to a significant
reduction in planning time, as we demonstrate in Section~\ref{sec:experiments}. 


%% file: related.tex
\section{Related work}
\label{sec:related}

Our work draws ideas from model learning, probabilistic modeling of
functions, and task and motion planning. 

There is a large amount of work on learning individual motor
primitives such as pushing~\citep{kroemer2016meta, hermans2013learning},
scooping~\citep{schenck2017learning}, and
pouring~\citep{pan2016robot,tamosiunaite2011learning,brandi2014generalizing,yamaguchi2016differential,schenck2017visual}.
We focus on the task of learning models of these primitives suitable
for multi-step planning.  We extend a particular formulation of
planning-model learning~\citep{kaelbling2017learning}, where
constraint-based preimage models are learned for parameterized action
primitives, by giving a probabilistic characterization of the
preimage and using these models during planning.

There are several other approaches for learning precondition and
effect models of sensorimotor skills that are suitable for planning.
\cite{konidaris2018skills} construct a completely symbolic model
of skills that enables purely symbolic task planning.  Our method, on
the other hand, learns hybrid models, involving continuous
parameters. \cite{kroemer2016learning} learn image
classifiers for preconditions but do not support general-purpose
planning. More recently, \cite{wang2019learning} learn state transition models for sequencing low-level motor skills that perform manipulation tasks.

We use \gp-based level-set estimation~\citep{bryan2006active, gotovos2013active,
  rasmussen2006gaussian, bogunovic2016truncated} to model the feasible
regions (superlevel set of the scoring function) of action
parameters.  We use the {\em straddle} algorithm~\citep{bryan2006active}
to actively sample from the function threshold, in order to estimate
the superlevel set that satisfy the constraint with high probability.
Our methods can be extended to other function approximators that give
uncertainty estimates, such as Bayesian neural
networks and their variants~\citep{gal2016dropout,lakshminarayanan2016simple}.

Alternatively, one can use \gp{} classification methods with active learning~\citep{kapoor2007active} to model our constraints. Active learning of \gp{} classifiers is often used for modeling safety constraints to help perform safe exploration~\citep{schreiter2015safe, englert2016combined}. The focus of this work, however, is to present a suite of approaches to address not only how to actively learn a model but also how to use learned models to solve complex long-horizon manipulation tasks. In this work, we only focus on one setting of the active model learning problem (level set estimation with \gp{} regression) but other active learning approaches can certainly be used.

Determinantal point processes ({\dpp}s)~\citep{kulesza2012determinantal} are typically used for diversity-aware sampling.
However, both sampling from a continuous \dpp{}~\citep{hafiz2013approximate} and learning the kernel of a \dpp{}~\citep{affandi2014learning} are challenging. 

Several approaches to \tamp{} utilize generators to enumerate infinite
sequences of
values~\citep{kaelbling2011hierarchical,srivastava2014combined,GarrettRSS17}.
Our learned samplers can be incorporated into any of these approaches.
Additionally, some recent papers have investigated learning effective
samplers within the context of \tamp{}.  
\cite{chitnis2016guided} frame learning plan parameters as a
reinforcement-learning problem and learn a randomized policy that
samples from a discrete set of robot base and object poses.  
\cite{kimICRA17} proposed a method for selecting from a discrete
set of samples by ranking new samples based on their correlation with
previously attempted samples.  In subsequent work, they instead train
a generative adversarial network ({\sc gan}) to directly generate a distribution of
satisfactory samples~\citep{kimAAAI2018}.









%% file: exp.tex
\section{Experimental domains} 
\label{sec:domains}

We analyze the effectiveness and efficiency of each component of our
system independently and then demonstrate their collective
performance in the context of planning for long-horizon tasks in a
simulated high-dimensional manipulation domain. 
We have carried out experiments in three settings:
\begin{itemize}
    \item {\em Kitchen2D}: a simulated 2D kitchen domain implemented in Box2D~\citep{box2d}; a description of the simulation and results can be found in appendix~\ref{sec:kitchen2D} and in our earlier paper~\citep{Wang2018ActivePlanning}.
    \item {\em Kitchen3D}: a new simulated 3D kitchen domain implemented in PyBullet~\citep{coumans2019}; a description of the simulation and results are given in this section.
    \item {\em KitchenPR2}: a real-world experiment with a PR2 robot; a description of the implementation and results are given in section~\ref{sec:real-world}.
\end{itemize}



\subsection{Implementation of Kitchen3D} \label{sec:kitchen3D}

\begin{figure*}
    \vskip 0.1in
    \centering
    \includegraphics[width=0.315\textwidth]{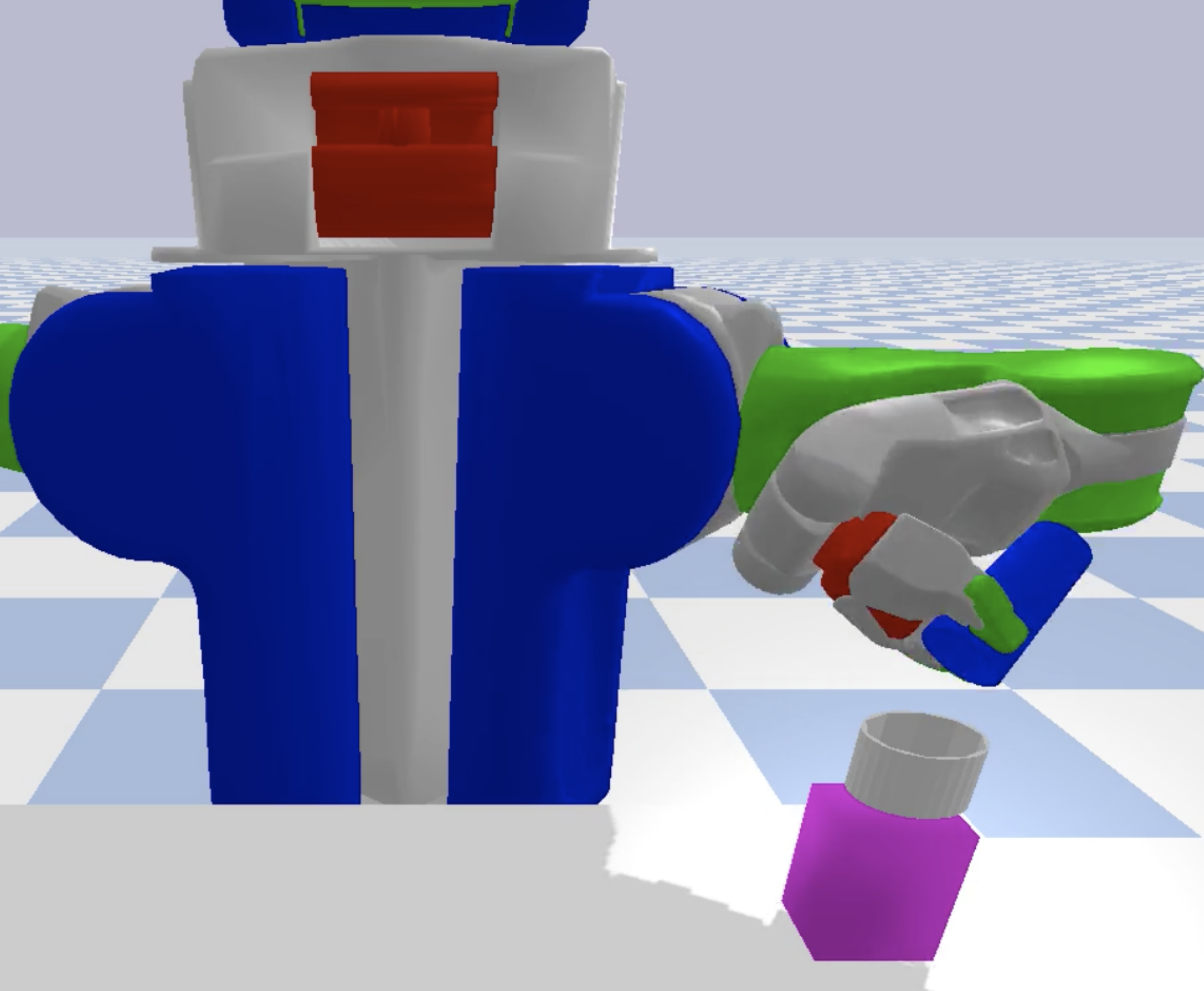}
    \includegraphics[width=0.32\textwidth]{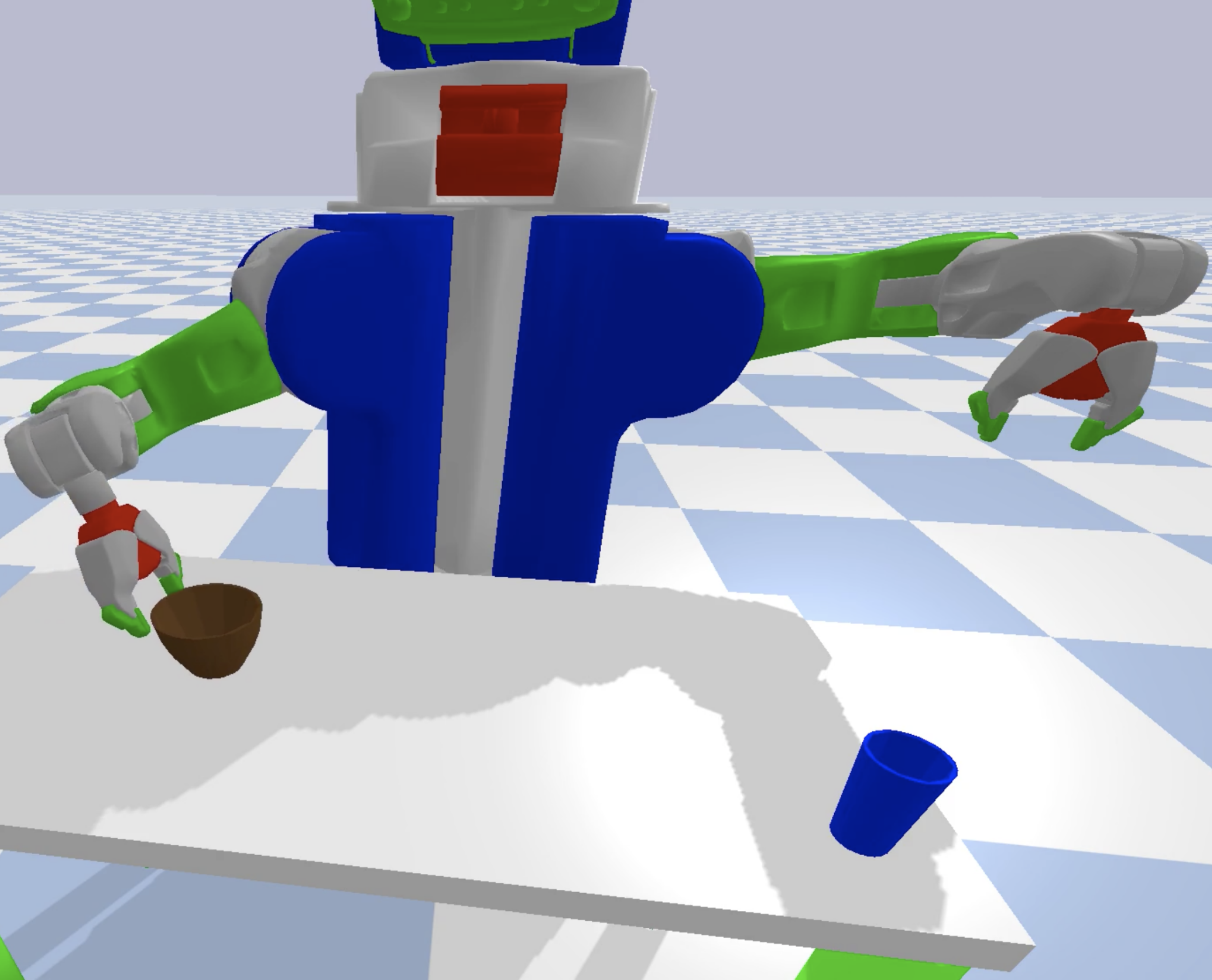}
    \includegraphics[width=0.35\textwidth]{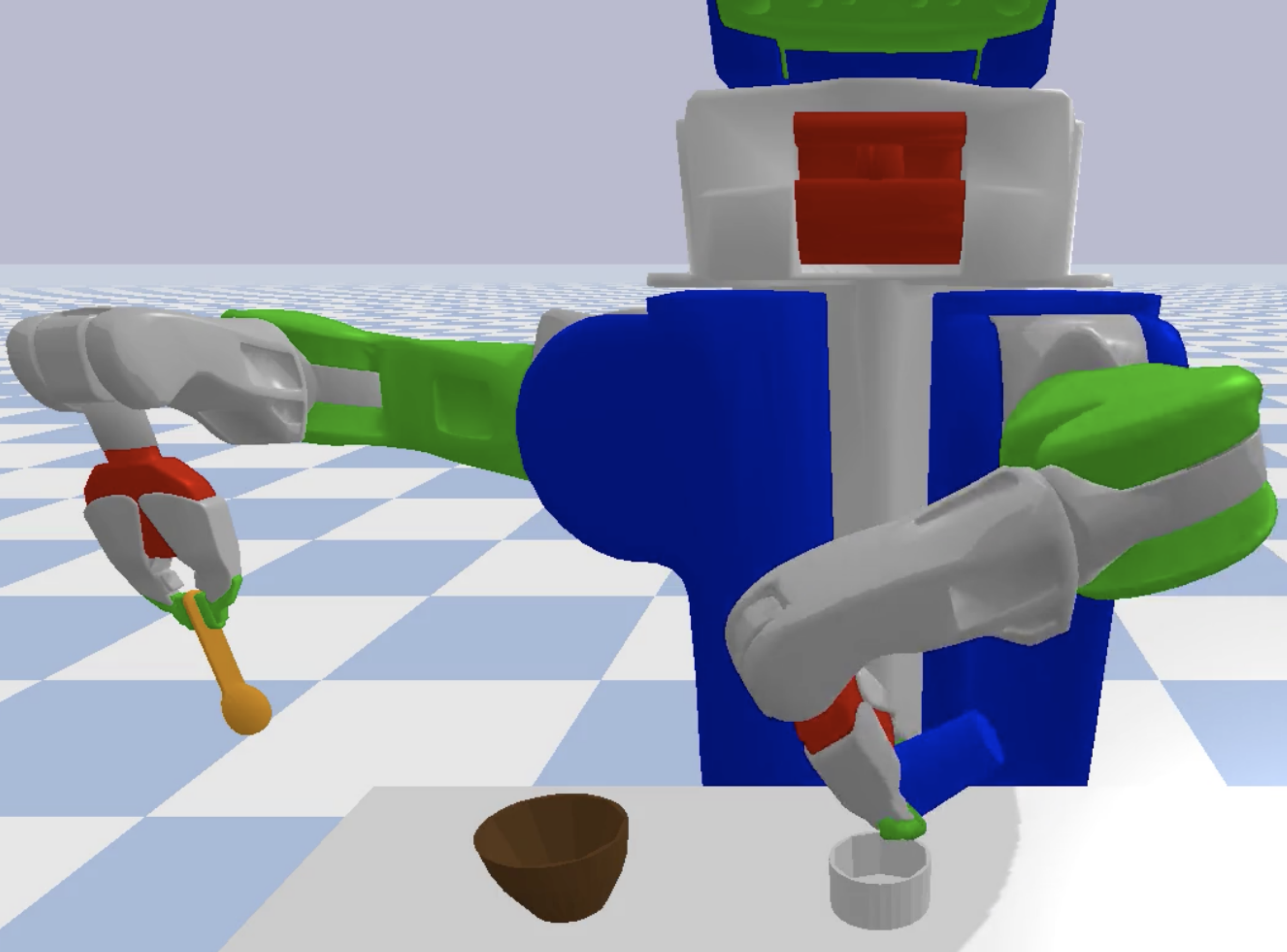}
    \caption{Scenes of a simulated PR2 robot solving planning tasks requiring pouring as well as stacking ({\em left}), pushing ({\em center}), and making coffee ({\em right}).}
    \label{fig:pr2simu}
    \vskip -0.1in
\end{figure*}

To investigate how well our approach scales to high-dimensional robots interacting with 3D objects, we implemented a simulated 3D tabletop environment with a dual-arm PR2 robot.
The 3D environment serves to bridge the gap between our previous 2D domain and a real-world robot operating scenario.
Our 3D simulation uses the PyBullet~\citep{coumans2019} physics engine. 
An illustration of the robot performing pouring and scooping skills is shown in figure~\ref{fig:pr2simu}. 

We experimented using objects created by randomly adapting meshes from our real-world data set of bowls, cups, and spoons, illustrated in figure~\ref{fig:objects}.
We uniformly-at-random and independently scale the diameter and height of each bowl and cup, but do not geometrically alter the three spoons.
We randomly sample mass, inertial, damping, and frictional properties for all objects according to a truncated Gaussian distribution. 
Finally, we randomly sample the number, radius, and density of the spherical liquid particles. 
This randomization process ensures that with probability one, each training or testing trial is unique.

We use PyBullet not only during simulation but also during planning for forward kinematics, collision checking, and visualization. 
We plan for each of the PR2's two 7 degree of freedom robot arms independently.
We use IKFast~\citep{openrave,diankov2010automated} for inverse kinematics.
We use RRT-Connect~\citep{KuffnerLaValle} to plan free-space arm motions.
Finally, we use Randomized Gradient Descent (RBD)~\citep{yao2005path,stilman2010global}, a constrained motion planner for planning robot joint motions that follow a Cartesian gripper path.

In order to transport and dump particles from a cup or spoon into a bowl, the robot must ensure that the cup or spoon does not spill any of the particles during transit.
To enforce this, we impose constraints $|\rho(q)| \leq \pi/6, |\phi(q)| \leq \pi/6$ that the grasped object's orientation remain within a safe region whenever the robot is carrying an object that contains particles, where $\rho(q)$, $\phi(q)$ give the roll and pitch of the tool at configuration $q$.
This constraint can be easily incorporated into robot motion planning by adding an additional check within the configuration ``collision'' function.


The robot executes actions by following planned sequences of arm or gripper configurations using a position controller.
We apply a rigid attachment constraint whenever the robot intentionally grasps an object to better model the real world, where the robot can reliably move without the held object deviating significantly relative to its hand.

We focus on learning conditional samplers for pouring and scooping because they are the most challenging to learn due to particle dynamics.
Similar to our work in {\em Kitchen2D}, we score pours and scoops relative to the filled capacity of the involved bowl or spoon.
We approximately compute the total number of particles that successfully ended up in a bowl, cup, or spoon by counting the number of particles contained within the 3D axis-aligned bounding box of these objects at the end of simulation.
We use a piecewise linear scoring function with threshold hyperparameter $\tau \in (0, 1)$ defined on the fraction of particles filled $x \in [0, 1]$:
\[g(x; \tau) = \begin{cases} 
    -1 + x/\tau  & 0 \leq x\leq \tau \\
    (x - \tau) / (1-\tau) & \tau < x \leq 1
   \end{cases}\;\;.\]
This function chosen for the following properties:  $g(0; \tau) = -1$, $g(\tau; \tau) = 0$, and $g(1; \tau) = +1$.
For pouring, $x$ is the final number of particles in the bowl over the initial number of particles in the cup, and we used $\tau = 0.9$.
For scooping, $x$ is the final number of particles in the spoon relative to the capacity of the spoon, and we used $\tau = 0.7$.

We assume that bowls and cups are approximately cylindrically symmetric, allowing us to parameterize contexts and controls using radial ($r$) and $z$ coordinates.
During learning, we use min-max normalization to scale each parameter value to within the interval $[-1, +1]$.

\subsection{Parameterization} \label{sec:parameterization}

\begin{figure}
    \centering
    \includegraphics[width=0.65\columnwidth]{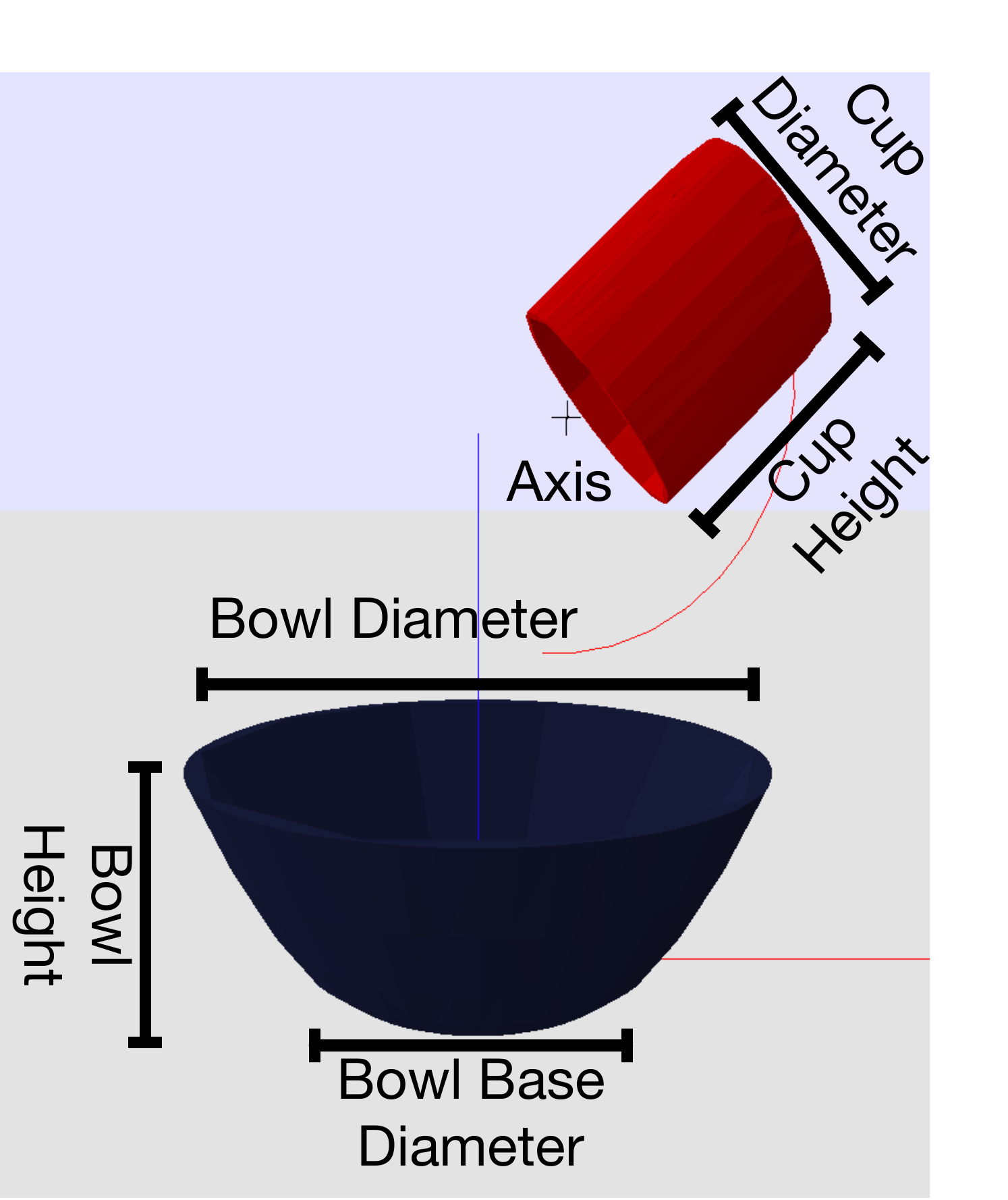}
    \caption{A visualization of the context parameters (the bowl and cup dimensions) and control parameters (the axis of rotation, the cup rotation frame, and the final pitch), for a pour. The red curve is the path of the cup base during the pour.}
    \label{fig:pour-parameters}
\end{figure}

For both pouring and scooping, we derive context parameters from the base diameter, top diameter, and height of a bowl or cup.
For scooping, we consider an additional context parameter for the length of a spoon.
As a result, pouring has 6 context parameters, and scooping has 4 context parameters.

\begin{figure}[ht]
    \centering
    \includegraphics[width=0.49\columnwidth]{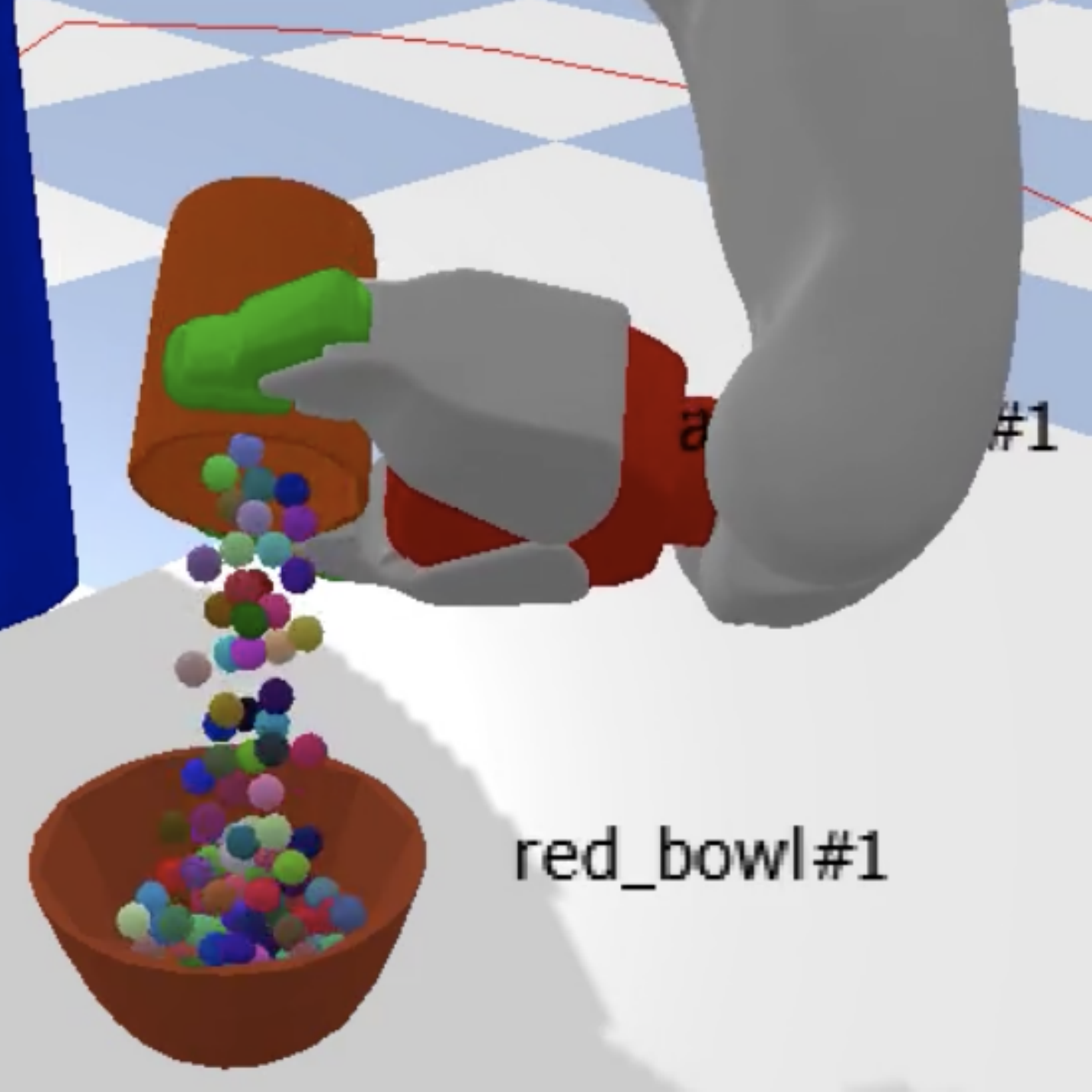}
    \includegraphics[width=0.49\columnwidth]{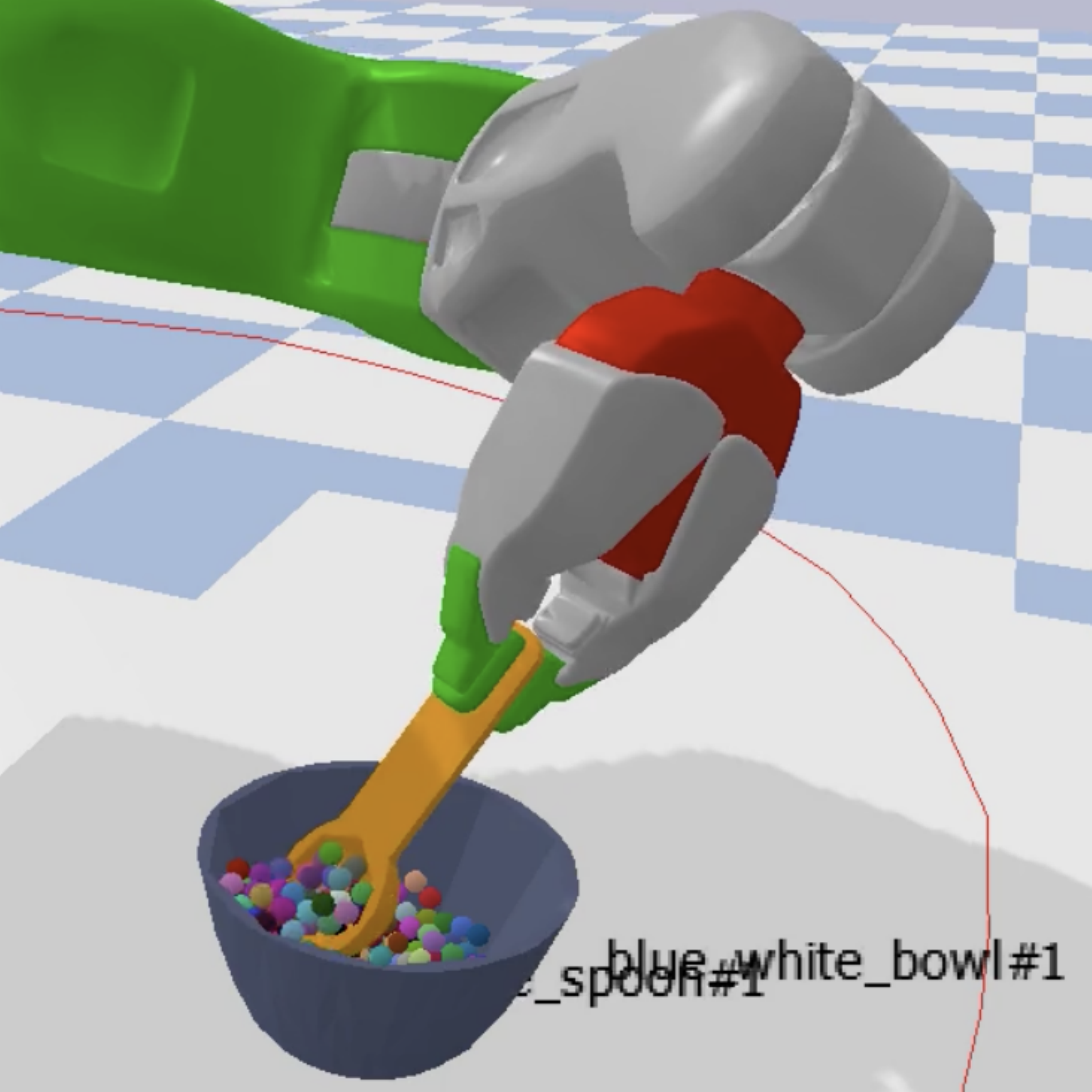}
    \caption{The simulated robot executing simulated pour ({\em left}) and scoop ({\em right}) actions.} 
    \label{fig:pybullet}
\end{figure}

Our control parameterization determines a sequence of waypoints that the cup or spoon moves through. 
Then, we interpolate though these waypoints to obtain the full path of the moving object.
We parameterized controls to be relative to the center of the base of a bowl or cup.
The pouring control parameters are the initial upright $r, z$ position of the cup relative to the bowl, the $r, z$ point for the cup to rotate about relative to its initial position, and the final pitch of the cup.
The scooping control parameters are the initial downward-facing $r, z$ position of the spoon relative to the bowl, the $r$ scoop distance, and the final $r, z$ upright position of the spoon.
Thus, pouring and scooping both have 5 control parameters.
We normalize distance-related control parameters relative to a bowl context parameter defined on the same coordinate in order to make the parameter space invariant to size of the involved bowl.
This ensures that uniform exploration of the prediction space produces roughly the same frequency of successful pours across different bowl sizes.
Figure~\ref{fig:pour-parameters} visualizes the context and control parameters for a pour.
Figure~\ref{fig:pybullet} demonstrates the robot executing sampled pouring and scooping actions in simulation.

We specify additional constraints per skill that enforce that execution does not knowingly cause any undesirable consequences.
In our application, we prohibit any unsafe contact between objects; however, this function can be any general-purpose test.
For pouring, this constraint enforces that full cup trajectory must not collide with the bowl.
For scooping, this constraint enforces that the final spoon pose must not collide with the bowl.
Satisfying the hard constraint function does not guarantee that the planner will able to find a full collision-free robot path to execute the path specified by the control parameters.
For example, a proposed pour in the interior of a bowl might not collide with the bowl; however, it might not admit any collision-free grasps of the cup.
Because this failure can be evaluated during planning, the control parameter during training should not receive the same negative score as a pour whose low quality can only be deduced after execution.
Thus, we weakly penalize learner predictions for which we failed to find plans with a small negative score, reflecting the computational time wasted by considering that sample.


\section{Experiments in simulation} 
\label{sec:experiments}

We evaluated the performance of our approach in the {\em Kitchen3D} environment. 
See appendix~\ref{ssec:app-exp_active} for several additional experiments performed in our {\em Kitchen2D} environment~\citep{Wang2018ActivePlanning}.

\subsection{Supervised learning} \label{fig:kitchen3D}

\begin{figure*}[ht]
    \centering
    \includegraphics[width=0.99\columnwidth]{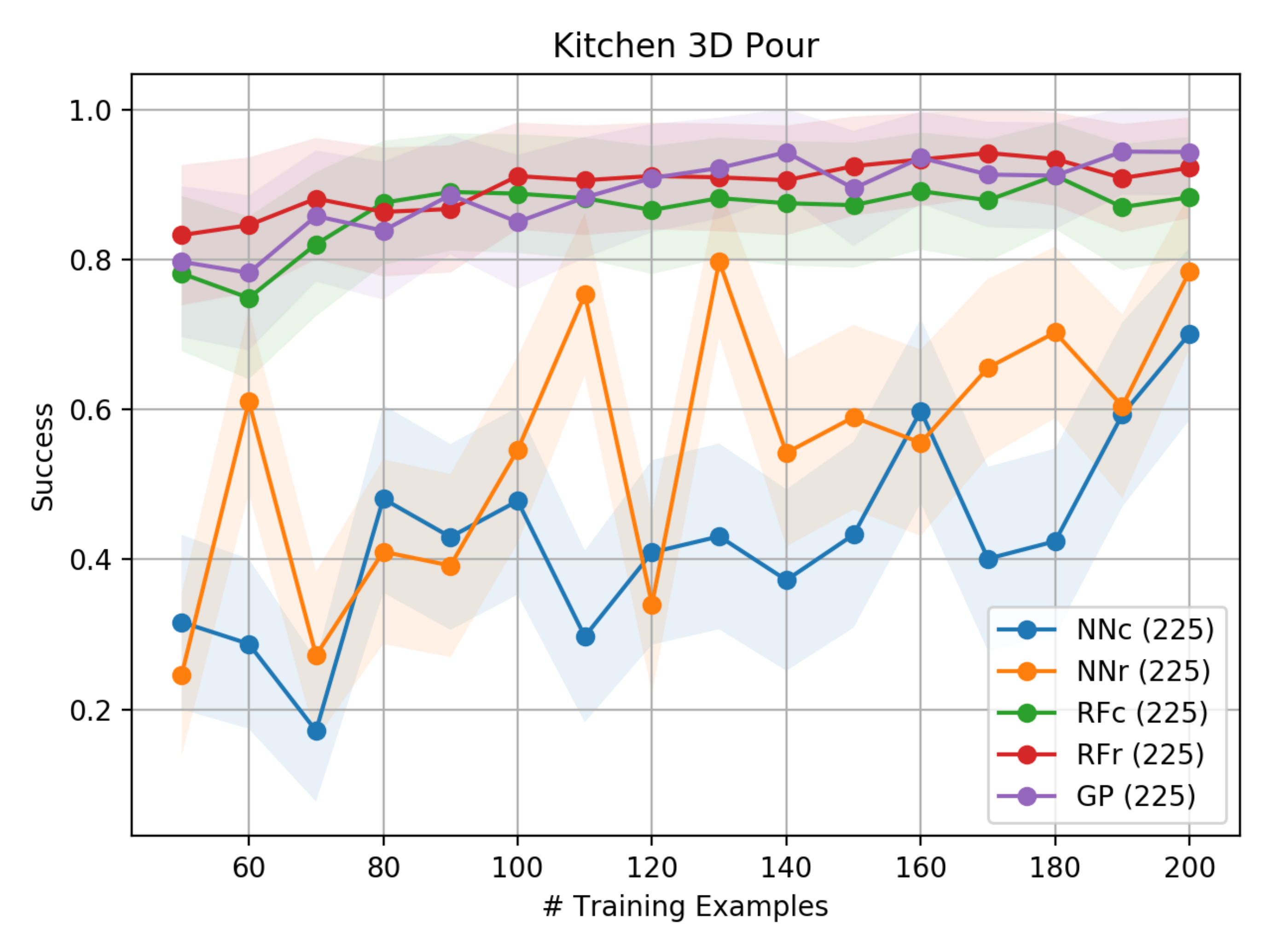}
    \includegraphics[width=0.99\columnwidth]{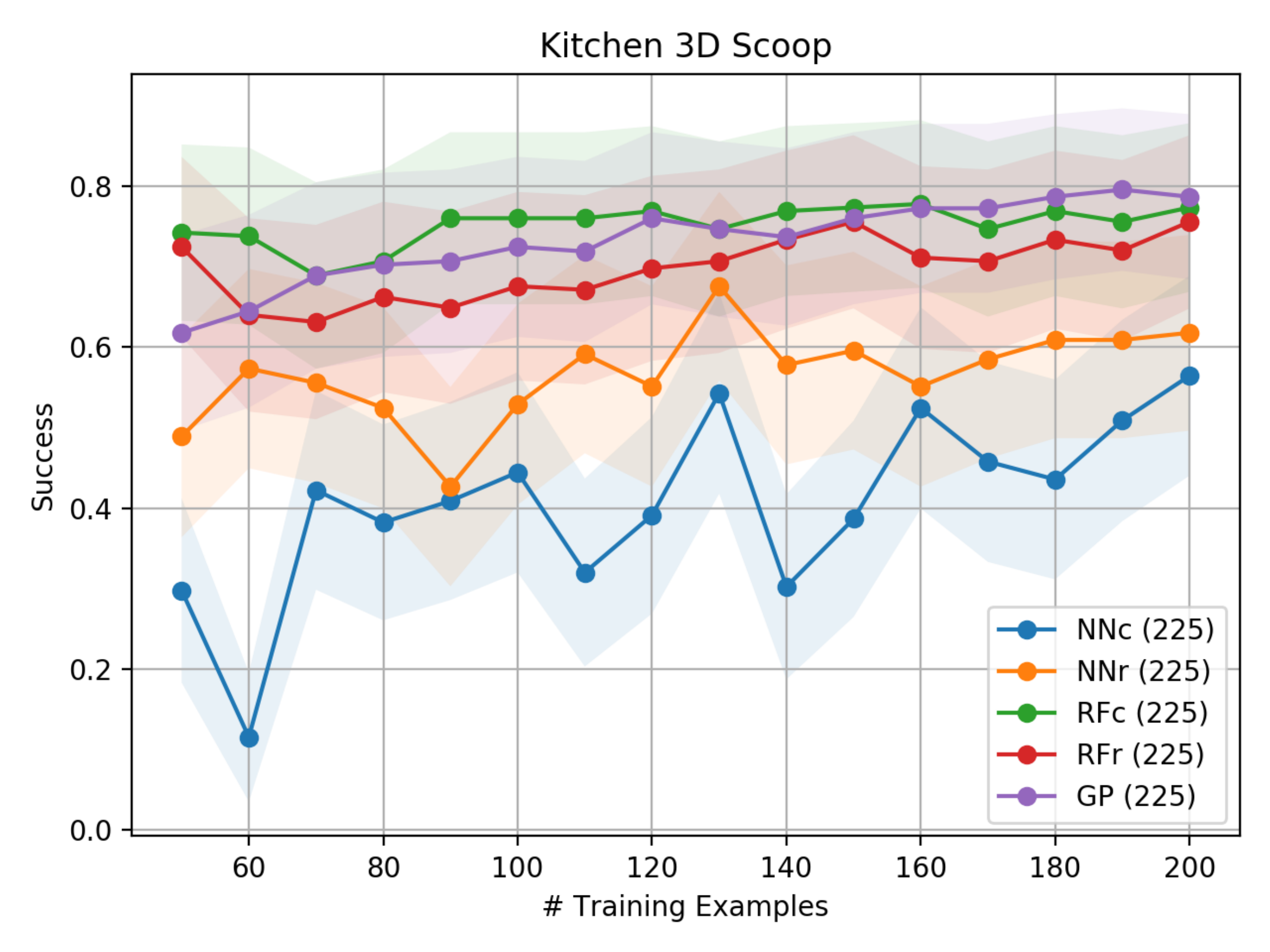}
    \includegraphics[width=0.99\columnwidth]{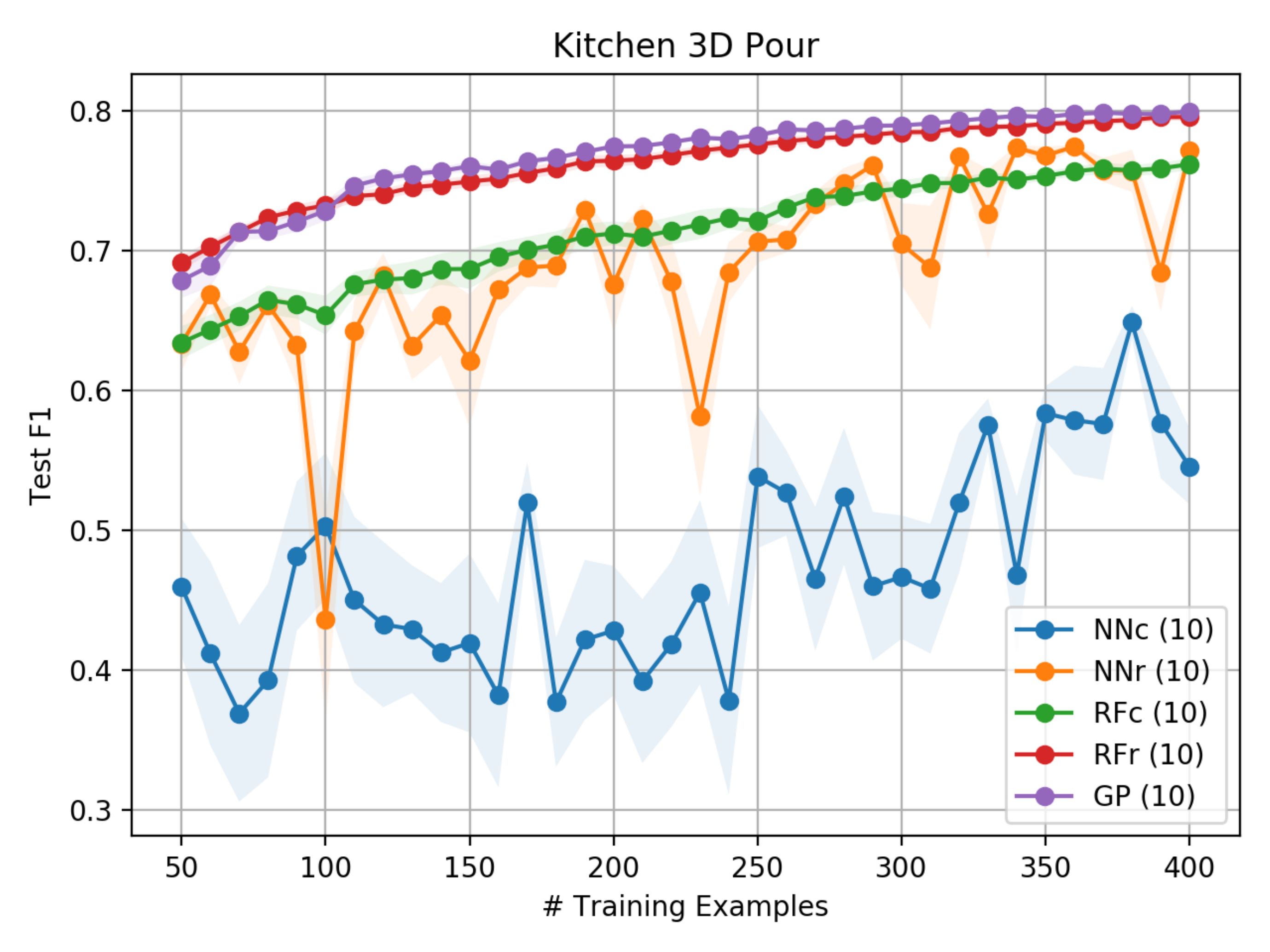}
    \includegraphics[width=0.99\columnwidth]{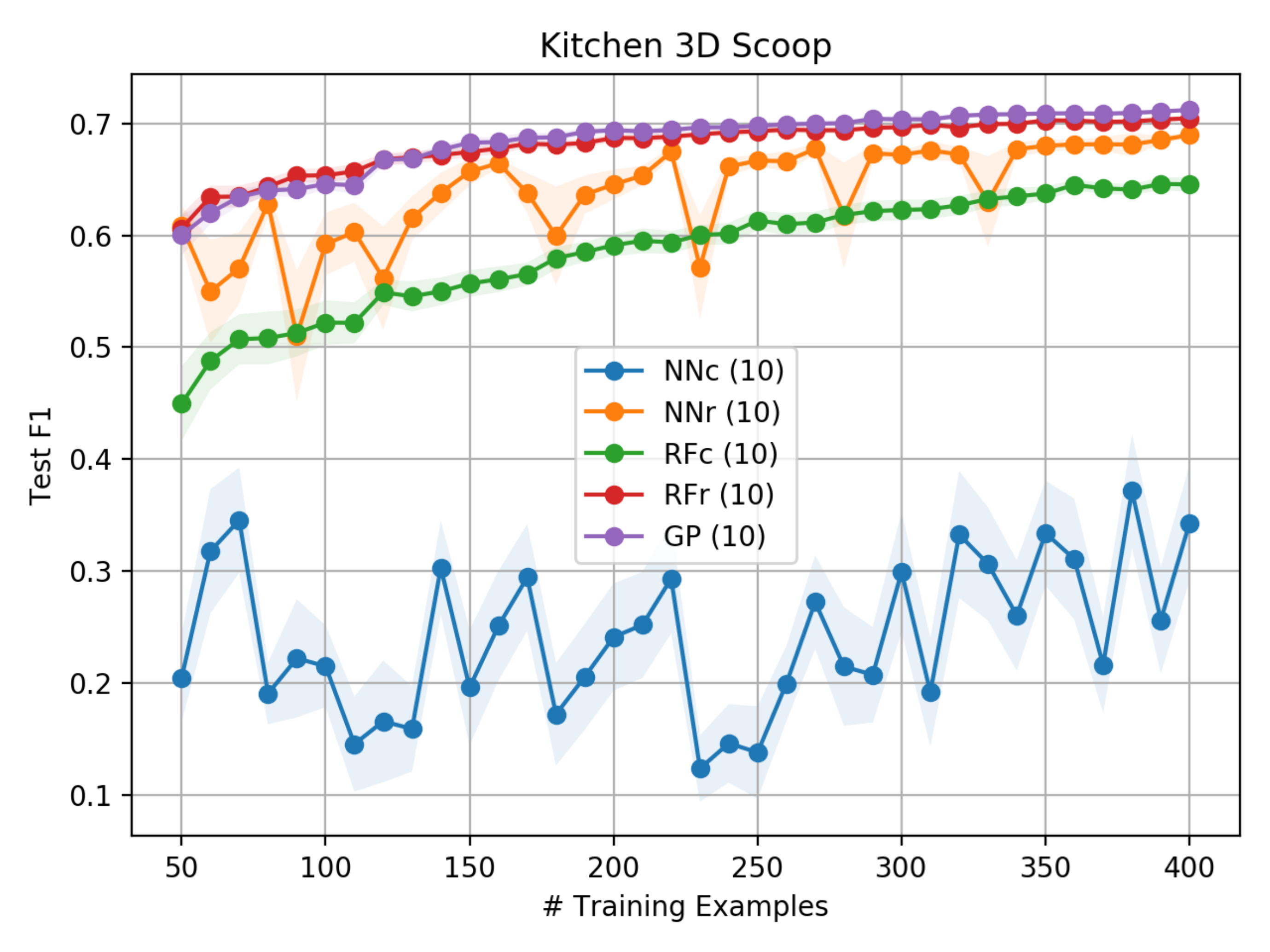}
    \caption{Pouring and scooping learning curves comparing a neural network classifier (NNc), a neural network regressor (NNr), a random forest classifier (RFc), a random forest regressor (RFr), and a \gp{} using the multi-layer perceptron kernel. {\em Top row}: the success rate of the most confident control parameter produced by each learner. {\em Bottom row}: the test F1 score for each learner.} 
    \label{fig:nn-rf-gp}
\end{figure*}

To aid with training and evaluating models in {\em Kitchen3D}, we first collected 10,000 pour and scoop trials by sampling a context and control parameter uniformly at random.
These examples are used for training traditional (non-active) learners, for holdout test evaluation, and for efficiently approximating active learning (as described in section~\ref{sec:active}).


We first compared the performance of a \gp{} using the multi-layer perceptron kernel trained {\em without} active learning with four baselines available through SKLearn~\citep{scikit-learn}:
(1) a neural network classifier (NNc), (2) a neural network regressor (NNr), (3) a random forest classifier (RFc), (4) and a random forest regressor (RFr).
The following section describes two experiments per skill, which compare the likelihood of success and classification coverage across the decision space.

\subsubsection{Success rate:} \label{sec:success}

First, we compared the average success rate of the single  best prediction per learner.
We trained each learner on 5 randomly shuffled sequences of 200 training examples.
We evaluated the success rate after every 10 examples by sampling 45 contexts, optimizing for the best control parameter per context, simulating the control parameter, and scoring the outcome.
For the SKLearn classifiers, the best control parameter was obtained by maximizing the probability that the parameter is successful.
For the SKLearn regressors, the best control parameter was obtained by maximizing the predicted score for the parameter.
To optimize these scores, we randomly sampled 1,000 control parameters, sorted them in order of decreasing score, and returned the first control parameter that respects the hard constraints (described in section~\ref{sec:kitchen3D}).
Finally, we treated the high-probability parameter that maximizes equation~\ref{eqn:ratio} as the best control parameter, which incorporates both the predicted mean and standard deviation.

Figure~\ref{fig:nn-rf-gp} ({\em top}) shows the success rate learning curve\footnote{This and all other {\em Kitchen3D} plots have 1/4 standard error shaded.}.
The random forest and \gp{} methods greatly outperform the neural network methods.
Additionally, the \gp{} ultimately achieves the best average success rate, likely due to its awareness of its own uncertainty.

\subsubsection{F1 score:} \label{sec:f1}

Second, we compared the F1\footnote{The F1 score is the harmonic mean of the precision and recall of a test.} classification score on held-out test data.
We trained each model on 10 randomly shuffled train and test splits, each consisting of 400 training examples and 1000 test examples.
The predicted label for a classifier is simply the most likely class, and the predicted label for a regressor is positive if the expected score is positive.
Figure~\ref{fig:nn-rf-gp} ({\em bottom}) displays the test F1 score learning curves.
The regressors outperform the classifiers, despite the fact the models were evaluated using the F1 score, a classification metric.
This is likely because the underlying score functions are real-valued.
When near the zero level set, small changes in score, which may be due to simulation noise caused by latent physical properties, can change the binary label of the example.
As a result, these regions may have high variance due to the strict thresholding.
Thresholding the score would be particularly detrimental when estimating uncertainty using a \gp{}, as these regions have substantial uncertainty that is not derived from the underlying stochastic process but rather from the nature of thresholding.
An active learner trained on thresholded score might indefinitely select examples near the zero level set because the process noise there is much larger than the rest of the space.

\subsubsection{Kernel selection:} \label{sec:kernel}

\begin{figure*}[ht]
    \centering
    \includegraphics[width=0.99\columnwidth]{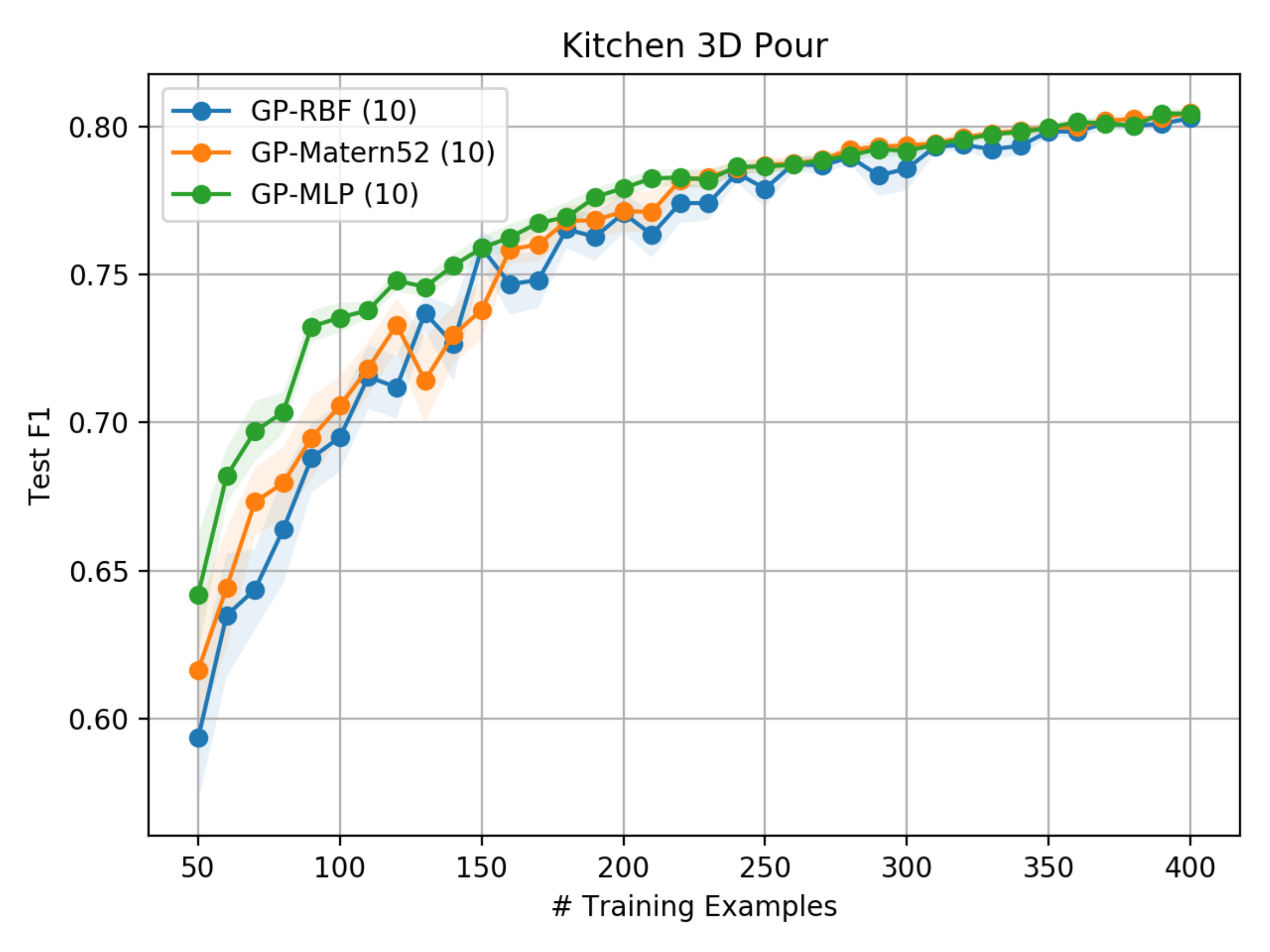}
    \includegraphics[width=0.99\columnwidth]{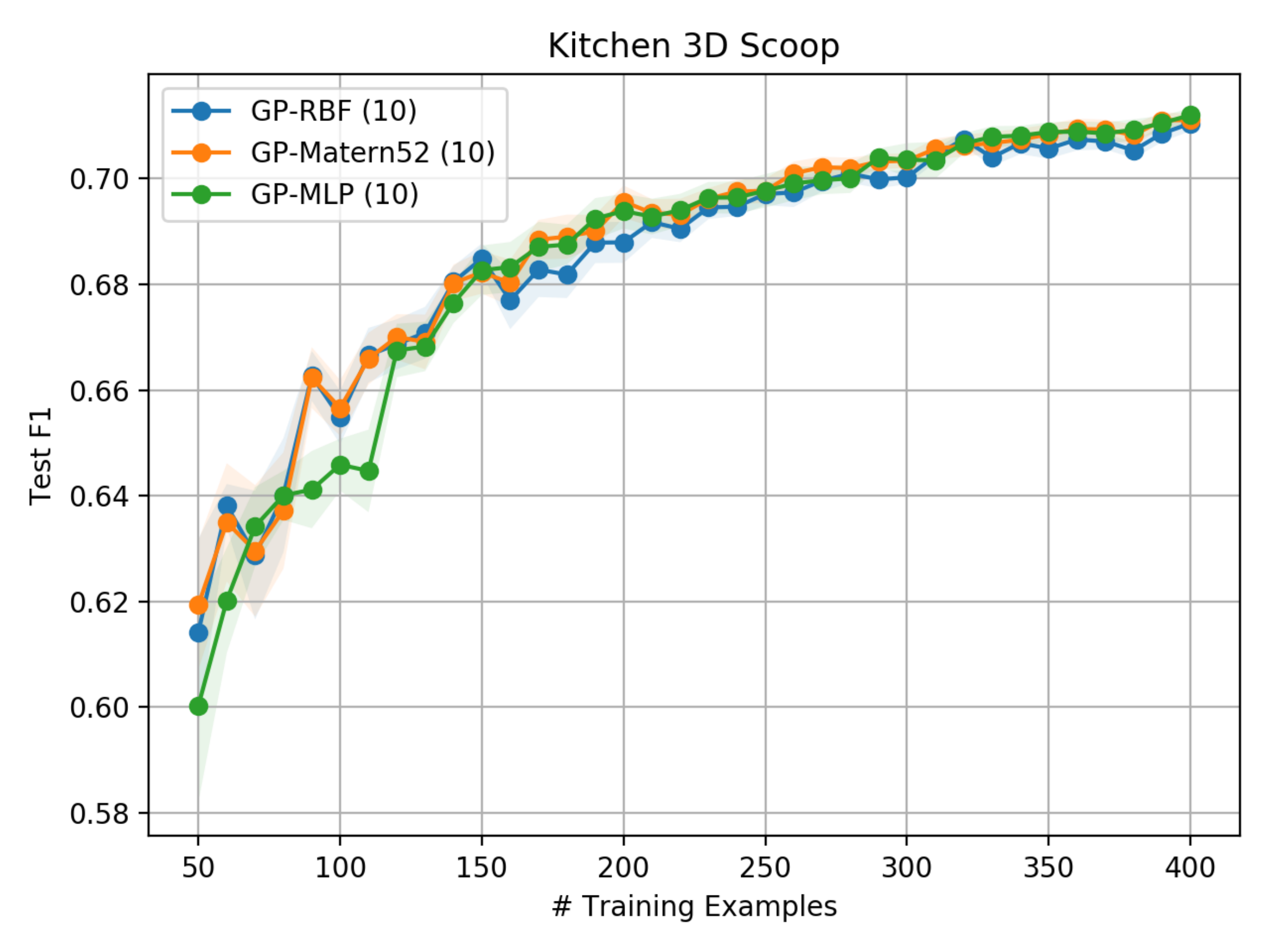}
    \caption{The test F1 score of the \gp{} when trained with the squared exponential, Mat\'ern, and multi-layer perceptron kernels (section~\ref{ssec:gp}).} 
    \label{fig:kernel}
\end{figure*}

Finally, we compared the \gp{} performance when trained on the three kernels described in appendix~\ref{ssec:gp}: the squared exponential radial basis kernel (GP-RBF), the Mat\'ern kernel (GP-Matern52), and the less commonly used multi-layer perceptron kernel (GP-MLP).
Figure~\ref{fig:kernel}, compares the F1 test performance when experimenting with each of kernels, using the same conditions as described in section~\ref{sec:f1}.
The multi-layer perceptron kernel slightly outperforms both the squared exponential and Mat\'ern kernels.
We hypothesize that is due to the discontinuous nature of the pouring and scooping scoring functions; the score of a pour or scoop can vary dramatically when, for instance, the pour ejects particles near an edge of the bowl.

\subsubsection{Visualizing predictions:} \label{sec:visualization}


\begin{figure}[ht]
    \centering
    \includegraphics[width=0.315\columnwidth]{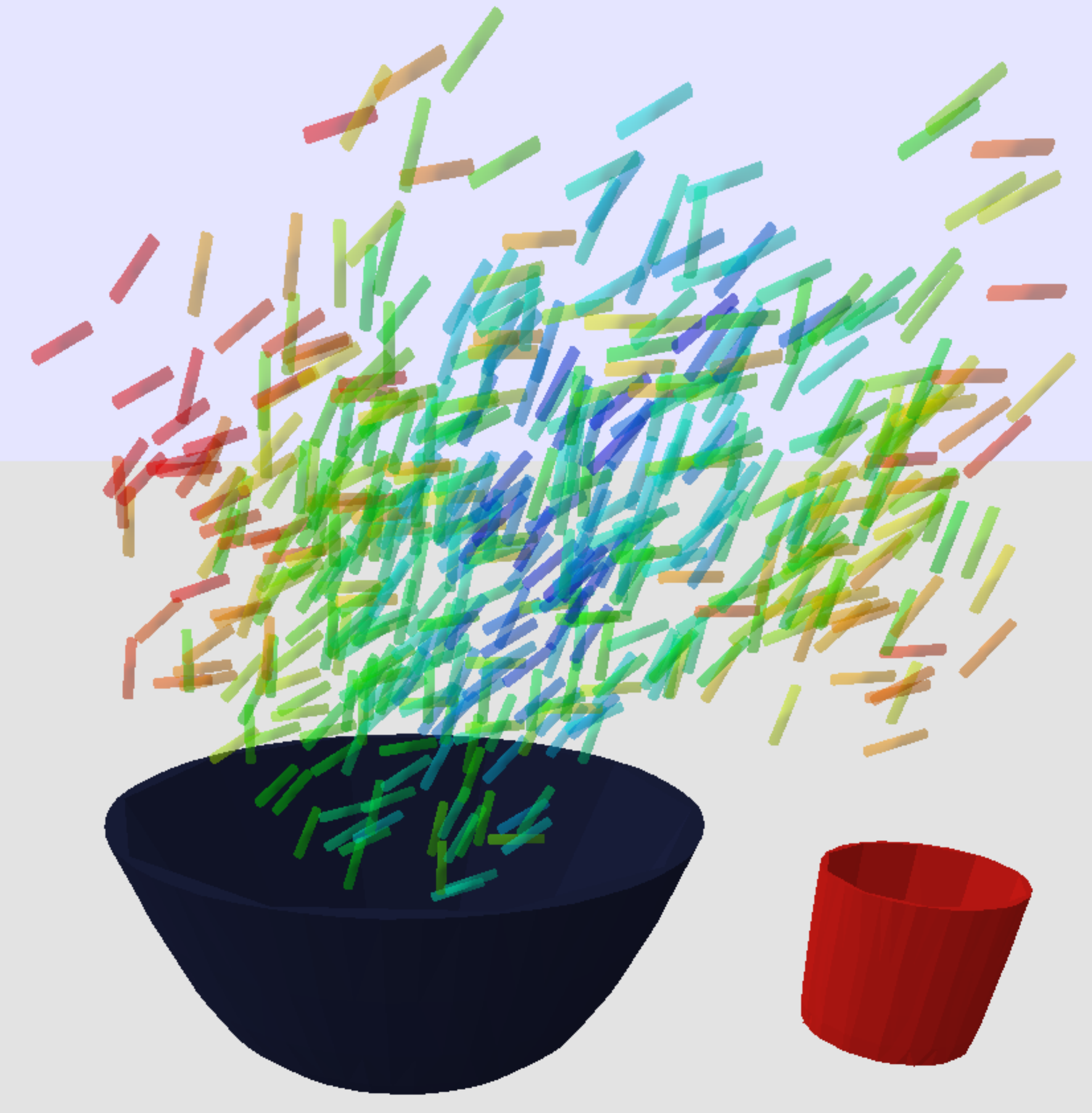}
    \includegraphics[width=0.355\columnwidth]{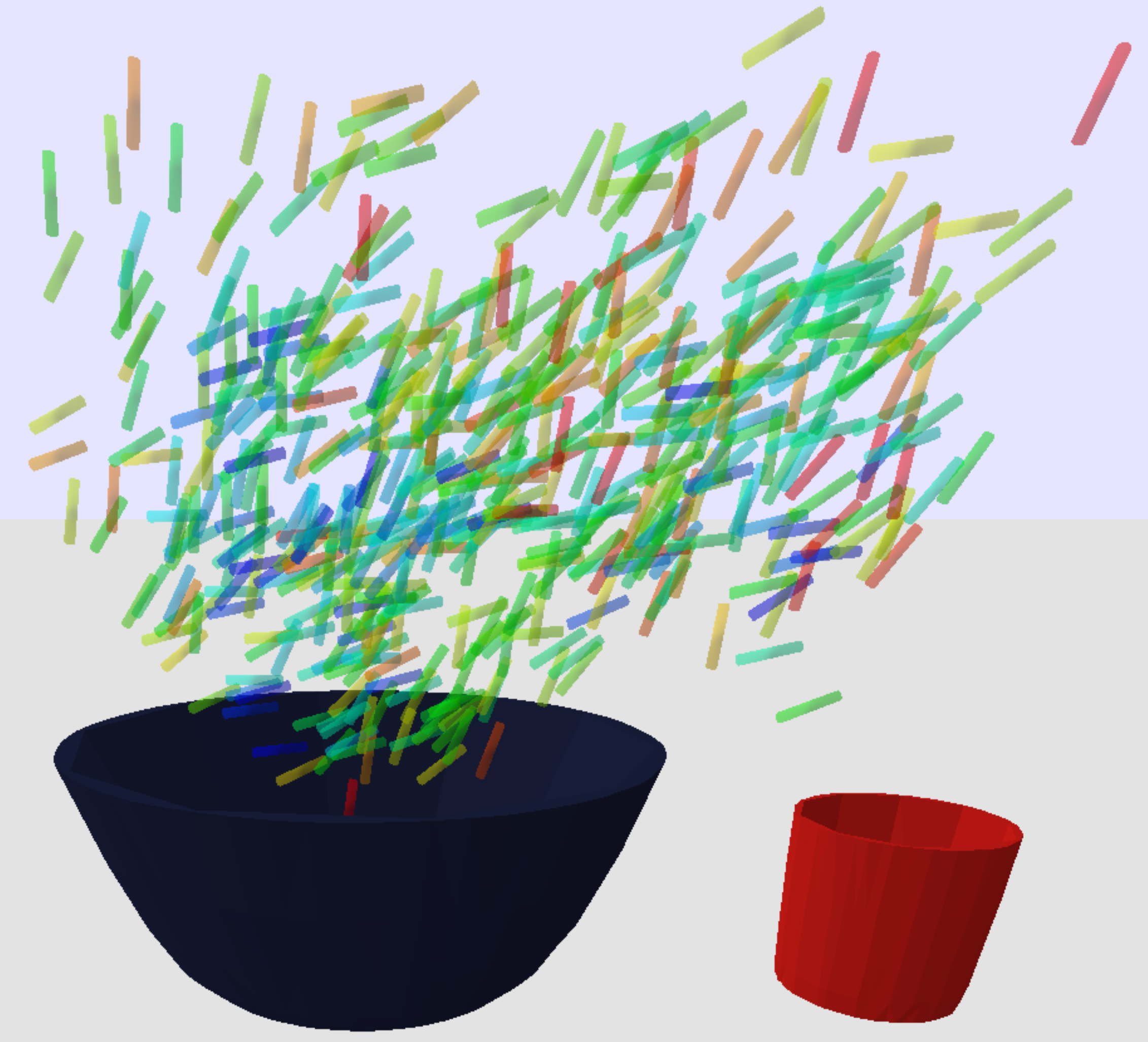}
    \includegraphics[width=0.31\columnwidth]{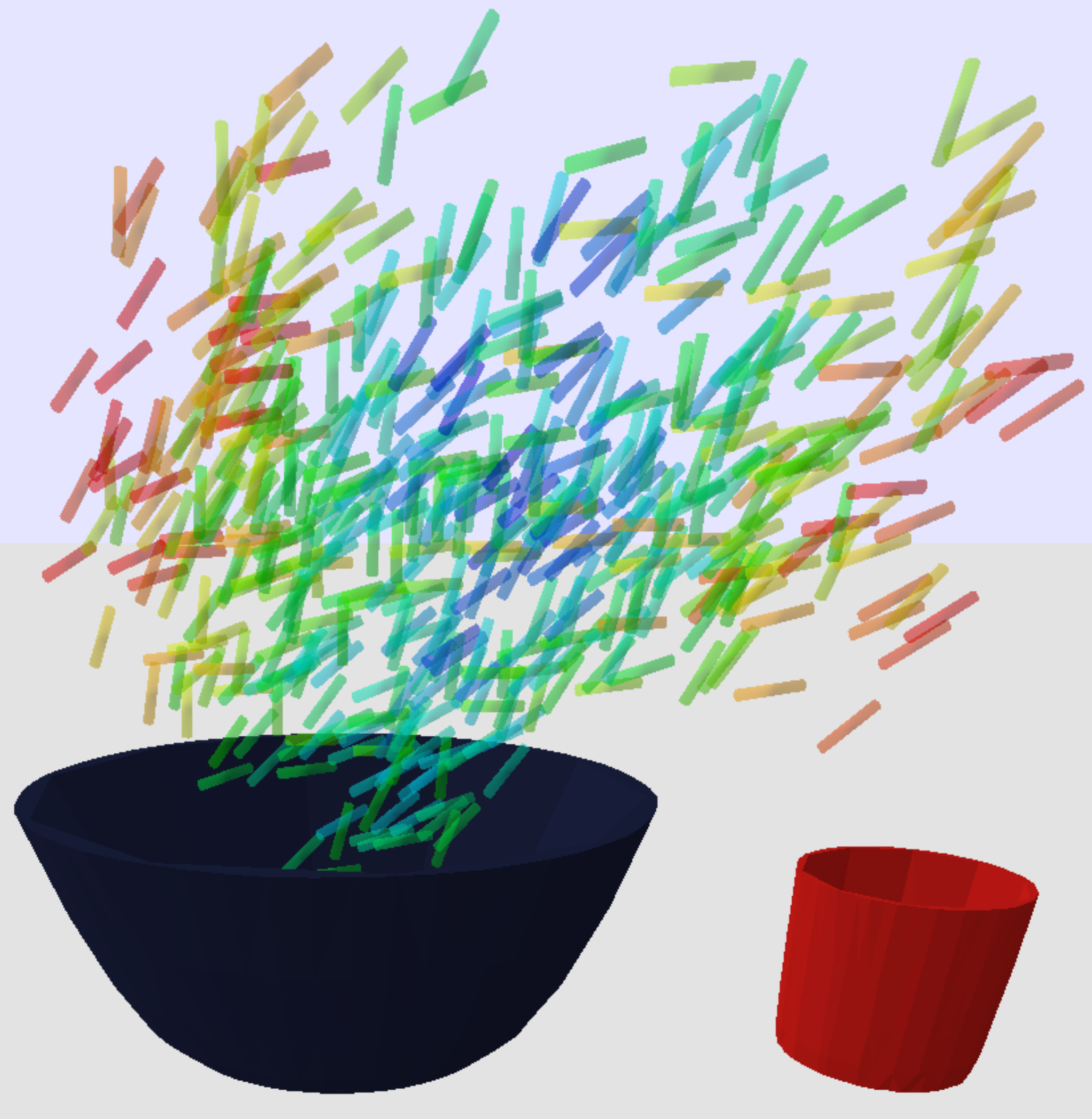}
    \caption{A visualization of the final cup pose for 500 pour valid control parameters. Poses are colored according to normalized mean ($\mu$), inverse standard deviation ($1/\sigma$), and best probability ($\mu/\sigma$) \gp{} predictions, where red poses have the smallest values and blue poses have the largest values.} 
    \label{fig:pour-dist0}
\end{figure}

We created a geometric visualization for the trained \gp{}'s mean and standard deviation score predictions across the space of legal control parameters.
Figure~\ref{fig:pour-dist0} renders a data set of pour control parameters for a single bowl and cup pair by displaying the final pose of the red cup.
It visualizes the \gp{}'s mean, inverse standard deviation, and most confident predictions by coloring small values red and large values blue.
The mean predictions ({\em left}) demonstrate that the model learns that the z-axis of the cup must roughly intersect with the interior of the bowl for a pour to be successful.
The standard deviation predictions ({\em center}) suggest that the more negative the cup pitch is, the higher variance in the outcome.
We hypothesize that this is because the longer rotation ejects the particles at larger velocities, making particles more likely to bounce out of the bowl.
The most confident prediction ({\em right}) combines the mean and standard deviation predictions.
Incorporating the standard deviation biases the learner towards high scoring pours that are closer to level.
These results suggest that the \gp{} is capturing intuitively relevant information for a successful pour.

\subsection{Active learning} \label{sec:active}

\begin{figure*}[ht]
    \centering
    \includegraphics[width=0.99\columnwidth]{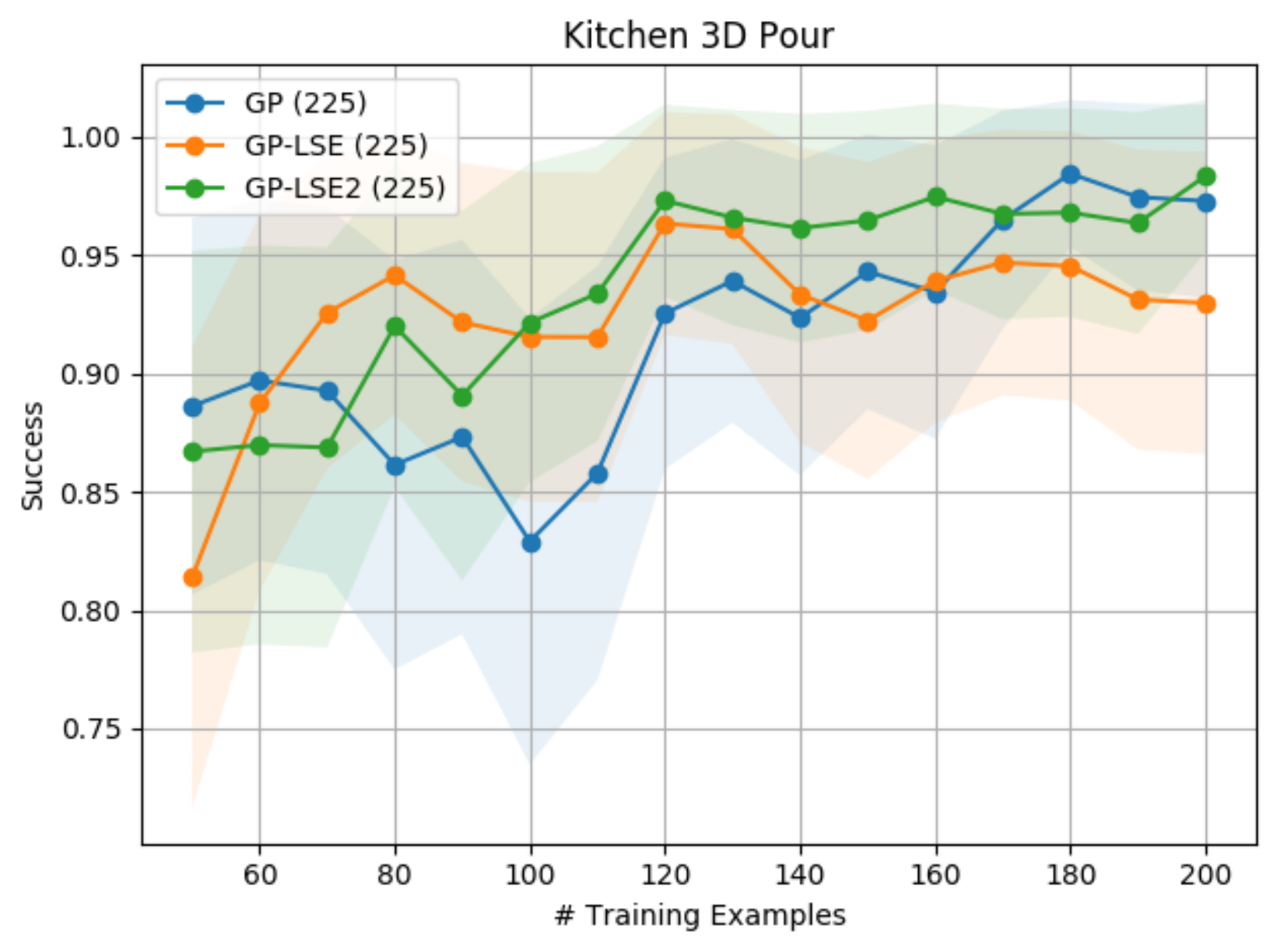}
    \includegraphics[width=0.99\columnwidth]{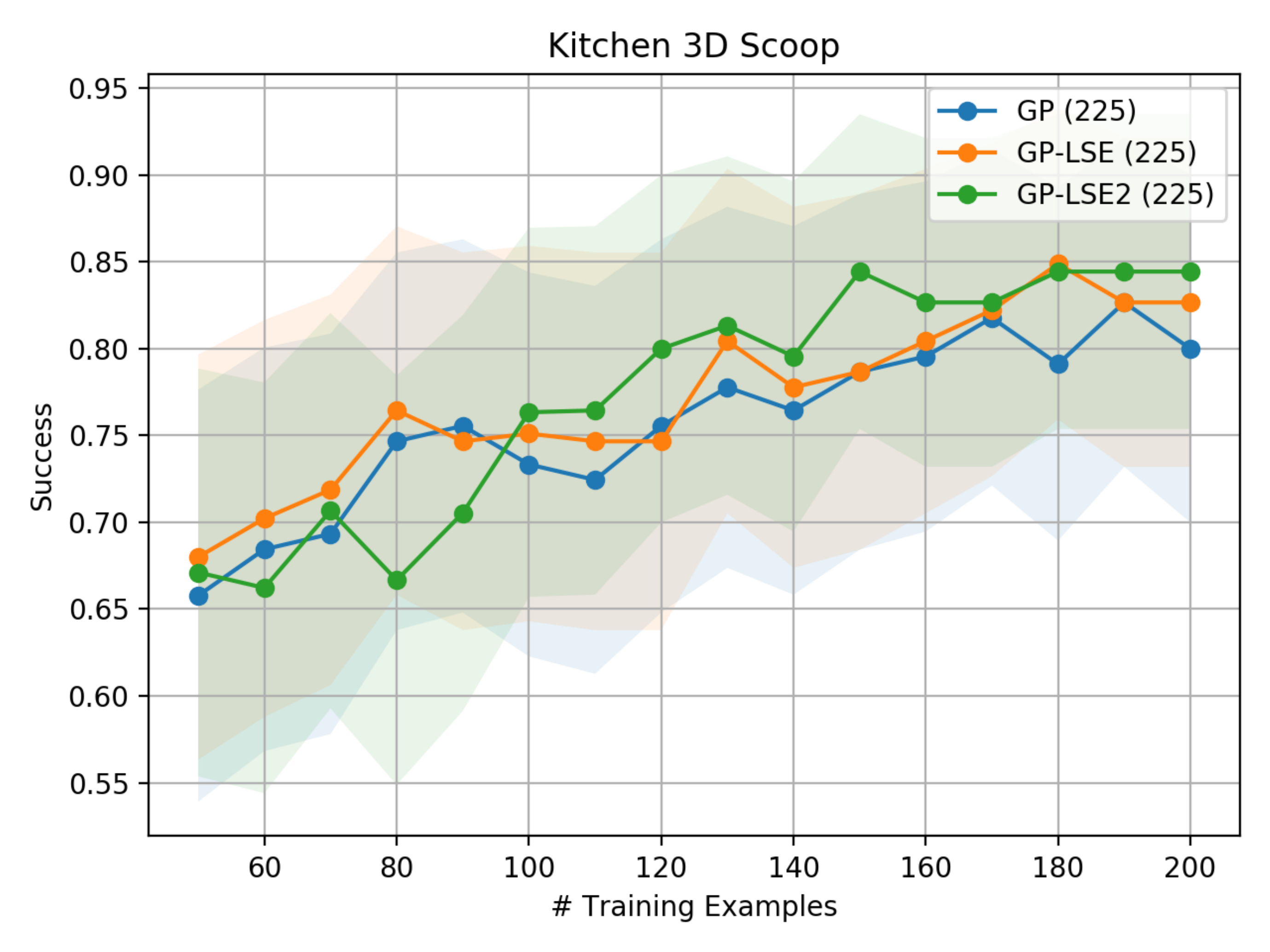}
    \includegraphics[width=0.99\columnwidth]{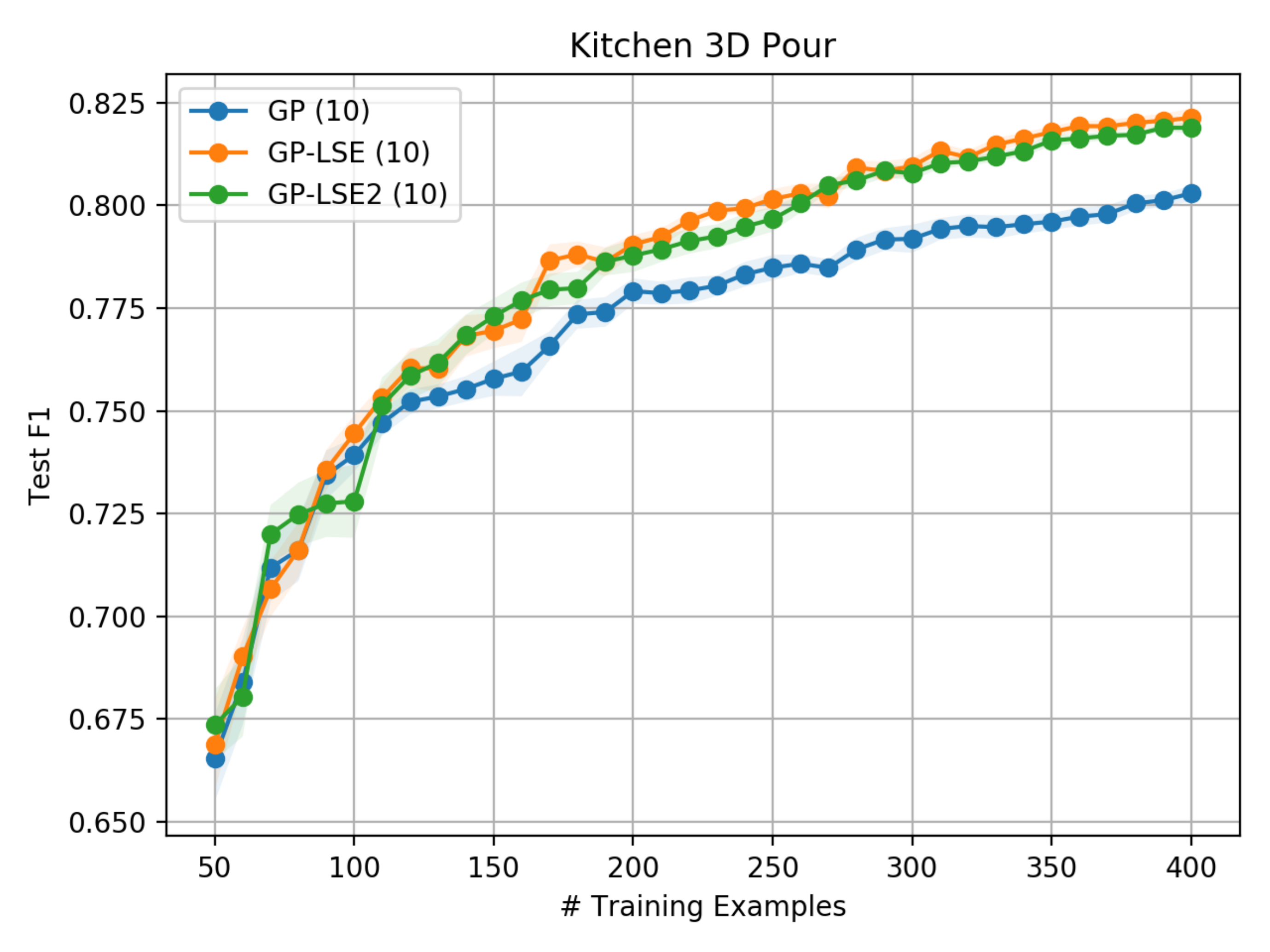}
    \includegraphics[width=0.99\columnwidth]{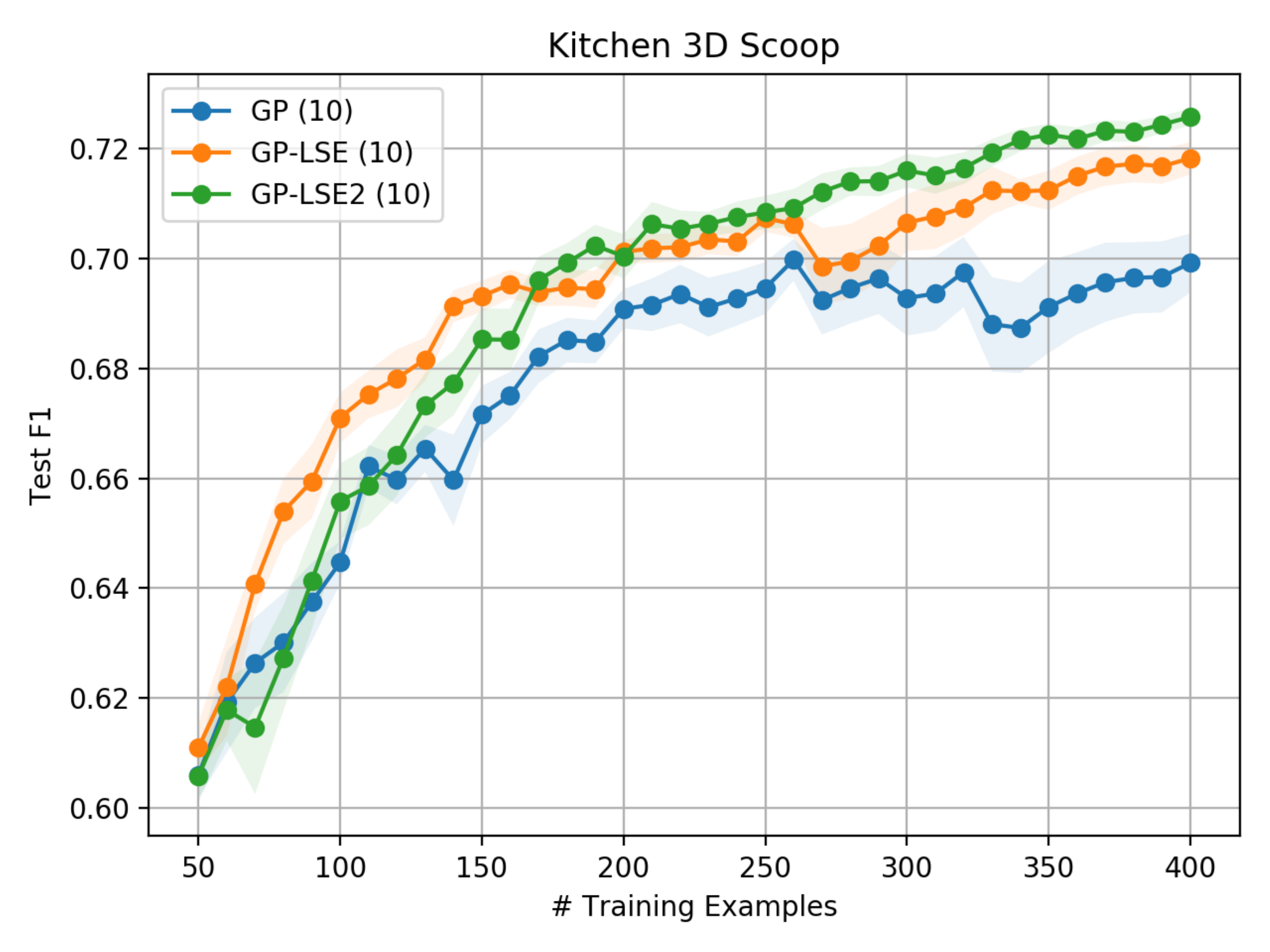}
    \caption{Pouring and scooping learning curves comparing a \gp{} trained without active learning (GP), a \gp{} that actively selects both the context and control parameters (GP-LSE), and a \gp{} that only actively selects the control parameter (GP-LSE2). {\em Top row}: the success rate of the most confident control parameter produced by each learner. {\em Bottom row}: the test F1 score for each learner.} 
    \label{fig:context-400}
\end{figure*}

We also evaluated the impact of training a \gp{} using active learning on the success rate and F1 score learning curves in this setting.
We compare a \gp{} trained {\em without} active learning (GP) with two \gp{}s trained {\em with} active learning strategies, both of which use the straddle algorithm (GP-LSE, GP-LSE2). 
Each \gp{} uses the multi-layer perceptron kernel.

When actively training a model in the real world, the learner can fairly quickly evaluate any control parameter.
However, this is not necessary true for context parameters because they often involve properties of physical objects.
If we applied the straddle algorithm to perform a {\em continuous} optimization over context parameters, we would need to fabricate objects with the selected sizes in order to faithfully score the trial.
While this could be possible through, for example, 3D printing, the real-world time and resource overhead would make it prohibitive.
However, given a finite set of contexts derived from a fixed set of existing objects, it is possible to perform a {\em continuous} optimization over control parameters per {\em discrete} context parameter select the best parameter pairs.
Still, this assumes that the robot can select the next context, which might not be true for a robot learning online in the wild. 

Motivated by the semantic differences between context and control parameters, we experimented with three partitions of parameters into those that are sampled uniformly at random and those that are actively optimized in some manner. 
Specifically, we compared sampling all parameters (GP), actively optimizing all parameters (GP-LSE), and sampling the context parameters but actively optimizing the control parameters with respect to the context parameters (GP-LSE2).

In order to faithfully train an active learner, training must be performed in series because 
every new training example modifies the learner's posterior and thus influences the selection of the next trial.
As a result, active learning must be performed serially while trials selected independently and randomly can be collected massively in parallel.
Because planning and simulating each trial takes at least 30 seconds, training several active learners over hundreds of training examples can be computationally burdensome.
To expedite experimentation, we performed {\em discrete} active learning over  the set of 10,000 training examples that we initially gathered randomly.
The active learners score each example using the straddle acquisition function and extract the example with maximum value without replacement.

Figure~\ref{fig:context-400} ({\em top}) displays the success rate of the three learners using the same experimental conditions as in section~\ref{sec:success}.
The active learners (GP-LSE, GP-LSE2) outperform the non-active learner (GP).
Ultimately, the active learner that randomly samples the context (GP-LSE2) resulted in the best success rate.
Figure~\ref{fig:context-400} ({\em bottom}) displays the F1 score of the three learners using the same experimental conditions as in section~\ref{sec:f1}.
Here, the active learners (GP-LSE, GP-LSE2) more conclusively outperform the non-active learner (GP).
Achieving a high F1 score requires making accurate predictions for most of the decision space, not just a single point per context.
As a result, methodically exploring high-variance regions outperforms random sampling.

\begin{figure*}[t]
    \centering
    \includegraphics[width=0.99\columnwidth]{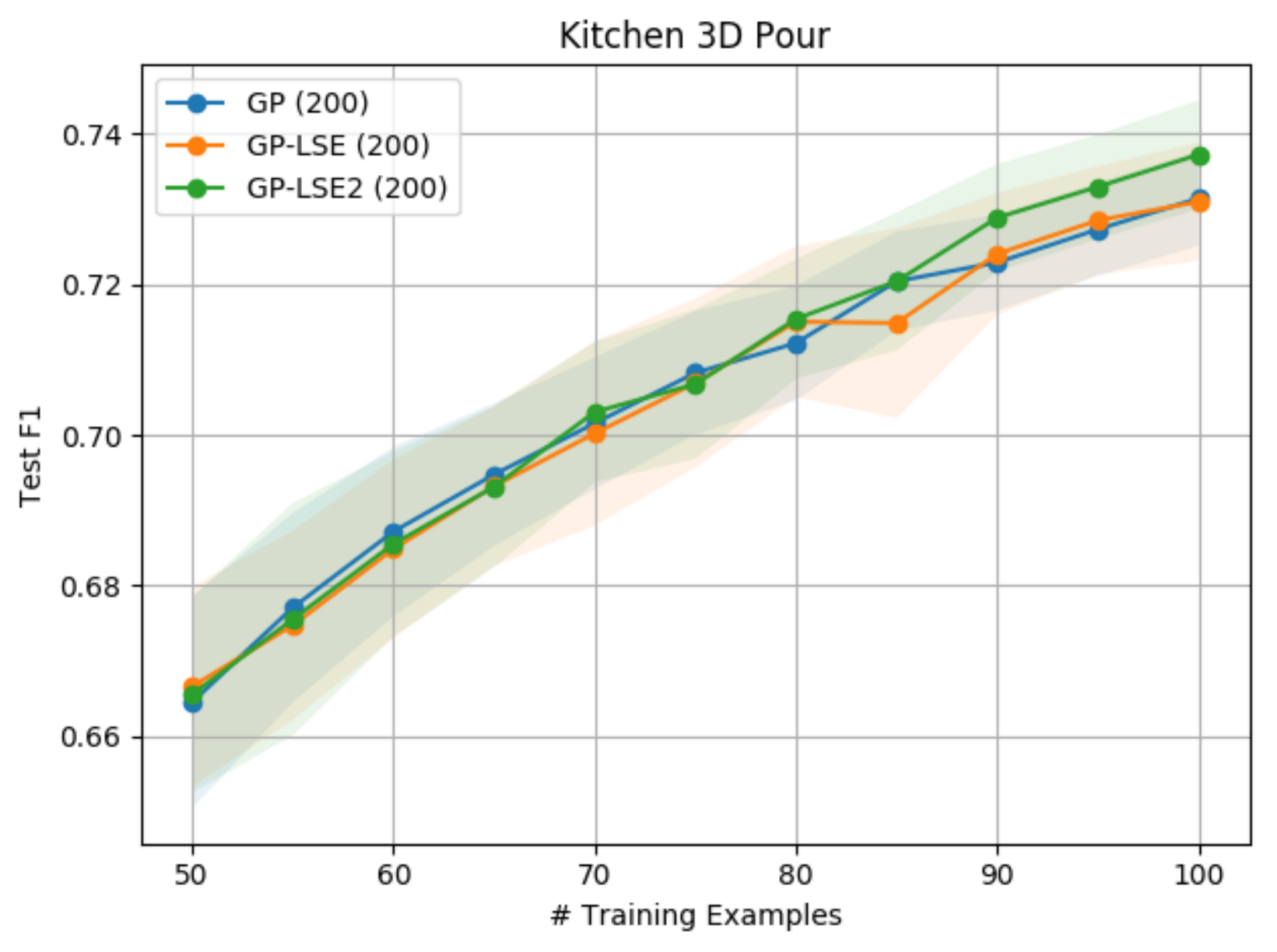}
    \includegraphics[width=0.99\columnwidth]{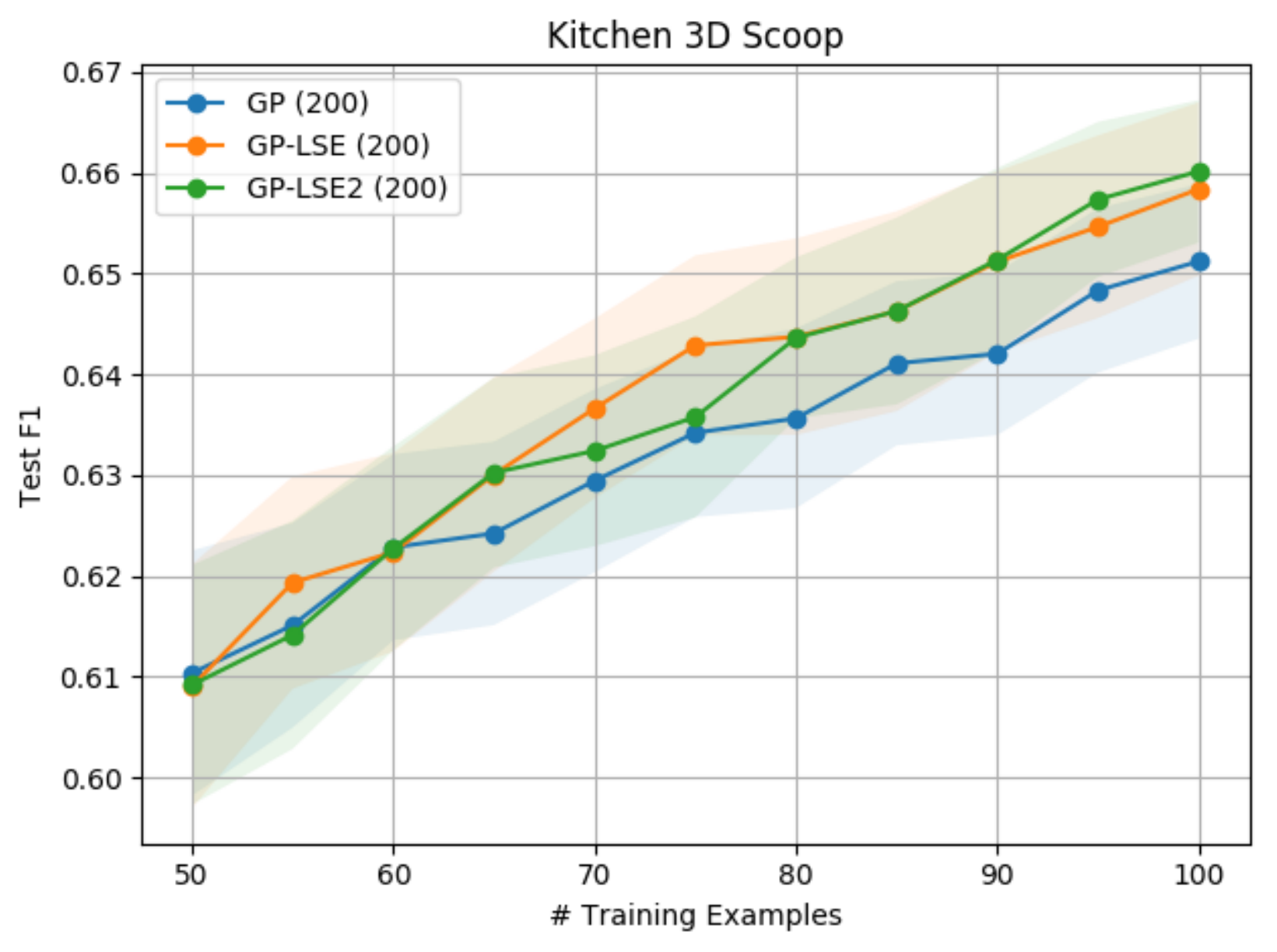}
    \caption{Pouring and scooping test F1-score learning curves from 50 to 100 training examples for the learners described in figure~\ref{fig:context-400}.
    Here, the learner process was repeated 200 times per learner in order to more accurately capture the variance in performance when using a small number of training examples.} 
    \label{fig:context-100}
\end{figure*}

Because gathering real-world data is labor intensive, we desired learning good pouring and scooping models with only around 100 training examples.
Thus, we performed a extensively-repeated experiment over a fewer samples in order to simulate our real-world experiments (described in section~\ref{sec:real-training}).
Instead of training on 400 examples for 10 episodes, we trained on only 100 examples but for 200 episodes.
Figure~\ref{fig:context-100} displays the F1 score for this experiment.
Although the variance is non-trivial, the active learner that randomly samples the context (GP-LSE2) results in the best average performance.
Our hypothesis is that, because control parameters are often more predictive of the score than the context parameters, GP-LSE focuses its attention on reducing uncertainty along the control dimensions, possibly neglecting the context dimensions.

\subsection{Adaptive and diverse sampling}
\label{ssec:exp-adaptive}

Given a probabilistic estimate of a desirable set of $\theta$ values,
obtained by a method such as \lse, the next step is to sample values
from that set to use in planning.  We compare simple rejection
sampling using a uniform proposal distribution (\rej), the basic
adaptive sampler from section~\ref{ssec:adaptive}, and the
diversity-aware sampler from section~\ref{ssec:diverse} with a fixed
kernel:  the results are shown in table~\ref{tb:sampling}. For all the results, we use $\Phi^{-1}(0.99 \Phi(\beta_*))$ to construct the high probability super-level set. 

\begin{table}
\caption{
Effectiveness of adaptive and diverse sampling. 
FP: the false positive rate of $50$ samples. $T_{50}$: the total sampling time of the $50$
samples. $N_5$: number of samples required to
  achieve $5$ positive ones. Diversity: the diversity rate of the 5 positive
  samples. 
} 
\label{tb:sampling}
\begin{center}
\resizebox{\columnwidth}{!}{%
\begin{small}
\begin{tabular}{llccc}
\hline
\abovestrut{0.15in}\belowstrut{0.10in}

&   & \rej & \adapt & \diverse  \\
\hline
\abovestrut{0.10in}
\parbox[t]{0mm}{\multirow{4}{*}{\rotatebox[origin=c]{90}{Pour (3D)}}}& FP (\%) $\downarrow$ & $0.03 \pm 0.10$ & {\color{red}$0.02 \pm 0.07$} & ${\color{red}0.02 \pm 0.08}$ \\
& $T_{50}$ (s) $\downarrow$& $143.56 \pm 176.05$ &  $72.84 \pm 71.26$ & ${\color{red}65.93 \pm 72.93}$ \\
& $N_5$ $\downarrow$& $5.14 \pm 0.45$ & $5.10 \pm 0.58$ & $5.15 \pm 0.71$  \\
& Diversity $\uparrow$&  $15.29 \pm 3.44$ & $15.40 \pm 2.94$ & ${\color{red}18.78 \pm 3.07}$ \\
\hline
\abovestrut{0.10in}
\parbox[t]{0mm}{\multirow{4}{*}{\rotatebox[origin=c]{90}{Scoop (3D)}}}& FP (\%) $\downarrow$& $0.13 \pm 0.17$ & $0.16 \pm 0.16$ & {\color{red}$0.12 \pm 0.10$} \\
& $T_{50}$ (s) $\downarrow$& $265.57 \pm 118.24$ &  $72.84 \pm 71.26$ & ${\color{red}35.11 \pm 18.73}$ \\
& $N_5$ $\downarrow$& $5.77 \pm 1.82$ & $6.11 \pm 1.77$ & $5.66 \pm 1.09$  \\
& Diversity $\uparrow$&  $10.93 \pm 2.50$ & $11.82 \pm 1.63$ & ${\color{red}14.57 \pm 2.13}$ \\
\hline

\end{tabular}
\end{small}
}
\end{center}
\vskip -0.1in
\end{table}

We report the false positive rate (FP)\footnote{The proportion of samples that do not
satisfy the true constraint.} on $50$ samples, the time to sample
these $50$ samples ($T_{50}$), the total number of samples required to
find $5$ positive samples ($N_5$), and the diversity of those $5$
samples. The experiments are repeated over 50 such samplers for each
  method. We do not limit CPU time for gathering $50$ samples for 3D simulated experiments.
The diversity term
is measured by $D(S) = \log \det(\Xi^S\zeta^{-2} + \mI)$ using a
squared exponential kernel with inverse length scale
$l=[1, 1, ..., 1]$ and $\zeta=0.1$. We run the sampling algorithm
for an additional 50 iterations (a maximum of 100 samples in total)
until we have 5 positive examples and use these samples to report the diversity quantity $D(S)$. 


\diverse uses slightly more samples than
\adapt to achieve 5 positive ones, and its false positive rate
is slightly higher than \adapt, but the diversity of the samples
is notably higher.  The FP rate of diverse can be
decreased by increasing the confidence bound on the level set.

\subsection{Learning kernels for diverse sampling} 

\label{ssec:exp_plan}

\begin{table}
\caption{Effect of distance metric learning on sampling.
}
\label{tb:timing}
\vskip -0.3in
\begin{center}
\resizebox{\columnwidth}{!}{%
\begin{small}
\begin{tabular}{lccc}
\hline
\abovestrut{0.15in}\belowstrut{0.10in}
WASH& Runtime (s) $\downarrow$& 60s SR  (\%) $\uparrow$ & 6s SR (\%) $\uparrow$\\
\hline
\abovestrut{0.10in}
\adapt   & $18.41\pm 8.87$  & $42.0\pm 10.3$& $28.0\pm 15.4$ \\
\gk & $18.22\pm 9.70$   &  $48.0\pm 7.5$ & $26.0\pm 16.6 $\\
\lk  & ${\color{red}17.07\pm 9.72}$ & {\color{red} $53.0\pm 6.0$} &  {\color{red}$40.0\pm 11.8$}       \\
\hline
\abovestrut{0.15in}\belowstrut{0.10in}
UNCLOG & Runtime (s) $\downarrow$& 60s SR  (\%) $\uparrow$& 6s SR (\%) $\uparrow$\\
\hline
\abovestrut{0.10in}
\adapt   & $44.20\pm 22.05$  & {\color{red} $23.0\pm 12.5$} & $5.0\pm 3.2$ \\
\gk & $44.85\pm 23.47$   &  $21.0\pm 9.2$ & $5.0\pm 3.2$\\
\lk  & ${\color{red}42.86\pm 23.34}$ & {\color{red} $23.0\pm 12.1$} &  {\color{red}$6.0\pm 5.8$}       \\
\hline
\end{tabular}
\end{small}
}
\end{center}
\vskip -0.2in
\end{table}

In the final set of experiments, we explore the effectiveness of the
diverse sampling algorithm with task-level kernel learning.
We compare \adapt, \gk with a fixed kernel and diverse sampling with
learned kernel (\lk), in every case using a high-probability
super-level set estimated by a~\gp. All the experiments are repeated 5 times
with random scene settings. In \lk, we use $\epsilon=0.3$.

\begin{figure}
    \centering
    \includegraphics[width=0.51\columnwidth]{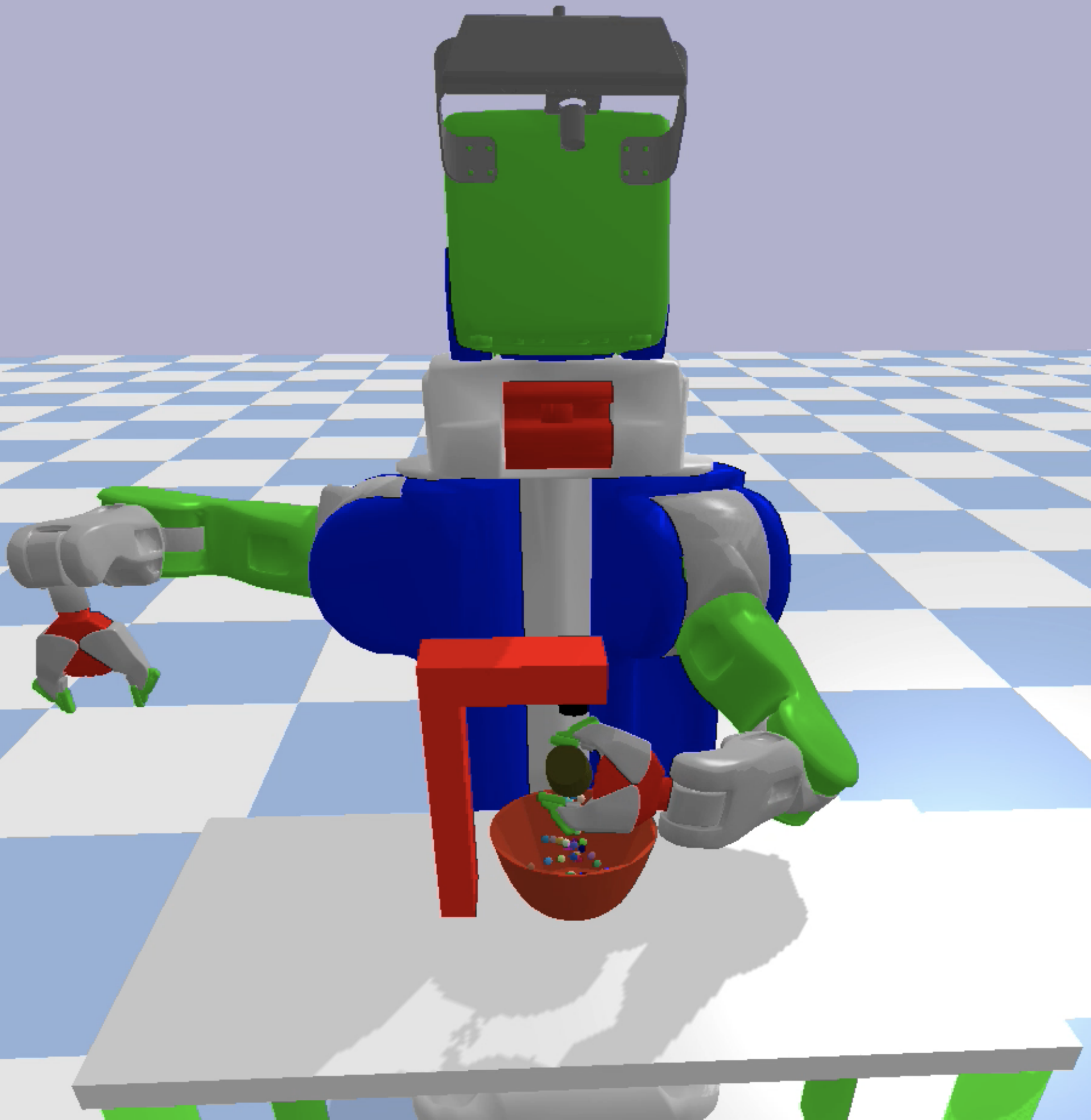}
    \includegraphics[width=0.45\columnwidth]{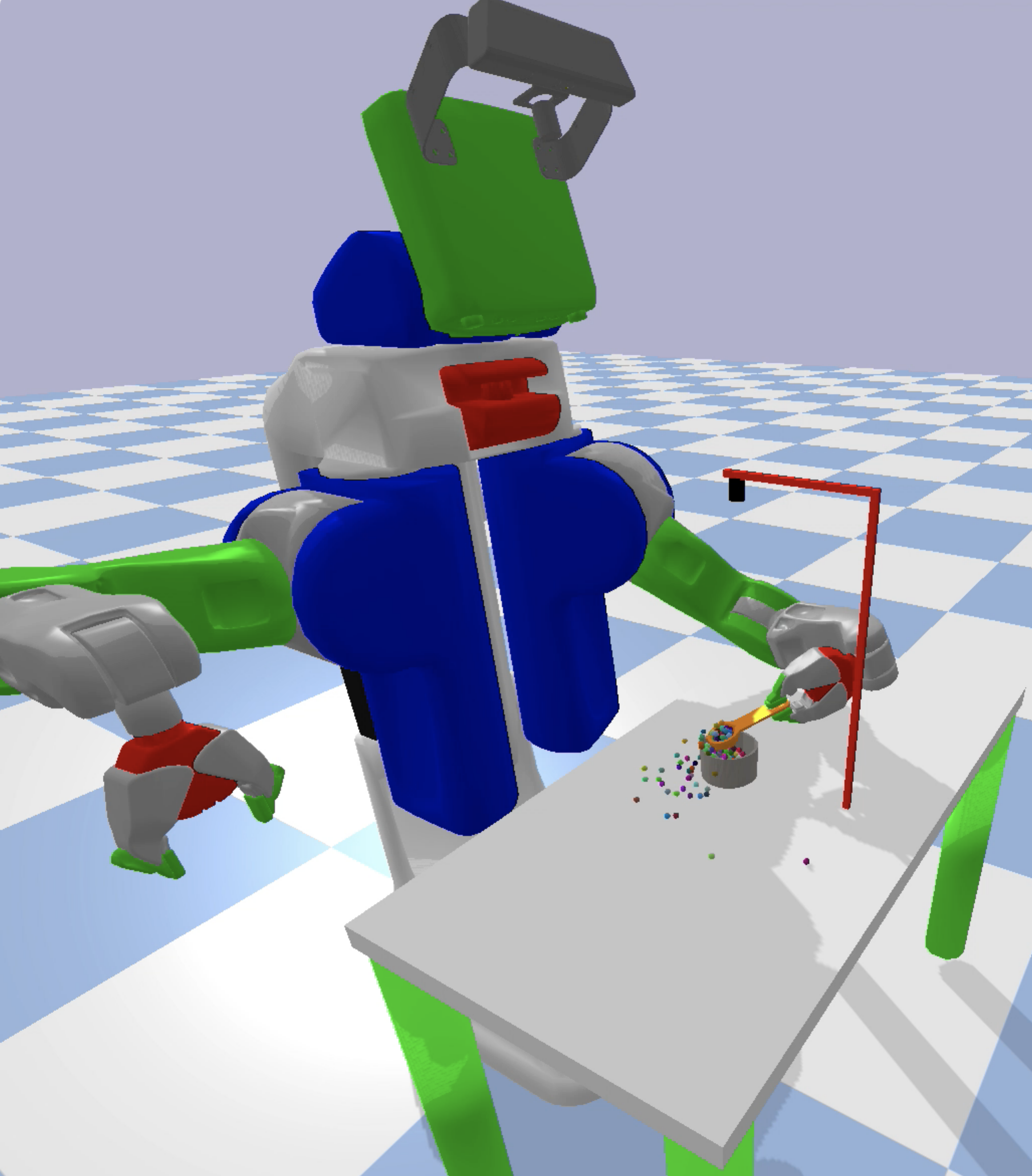}
    \caption{Examples for Task WASH and UNCLOG. Task WASH's goal is to pour from a cup to a bowl while avoiding the faucet next to the bowl.
Task UNCLOG is to scoop from a bowl while avoiding the faucet next to the bowl. } 
    \label{fig:sink}
\end{figure}

To test the performance of kernel learning, we design two tasks that require sophisticated manipulation to accomplish the goals. In the first task, called WASH, the goal is to pour ({\it e.g.} dish liquid) from a cup to a bowl while avoiding collisions with the faucet next to the bowl. The second task, called UNCLOG, aims to scoop ({\it e.g.} food waste)  from a bowl-shaped sink while avoiding collisions with the faucet. We select a fixed test set with 50 task specifications and repeat the evaluation 5 times. Different task specifications have different faucet sizes, bowl shapes, spoon sizes, cup sizes, faucet heights and distances between faucet and bowl. Figure~\ref{fig:sink} shows examples for task WASH and task UNCLOG. 

We show 
the timing and success rate results in table~\ref{tb:timing} (after training). Our empirical results shows that, in general, \lk is able to find a better
solution than the alternatives in both of these tasks. This suggests that the kernel learning approach that we adopted is indeed generating more suitable samples for the planner.
\hide{
Moreover, the two diverse sampling
methods achieve lower variance on the success rate and perform more
stably  after training. 
}



\hide{
\begin{table}[t]
\caption{Timing results and plan success rates (SR) for pouring task. The runtime only includes runs that successfully completed within 60s.}
\label{tb:timingpour_hard}
\vskip -0.2in
\begin{center}
\begin{small}
\begin{tabular}{lccc}
\hline
\abovestrut{0.20in}\belowstrut{0.10in}
Method & Runtime (s) & 60s SR  (\%) & 6s SR (\%) \\
\hline
\abovestrut{0.20in}
Adaptive   & $3.22\pm 6.51$  & $91.0\pm 2.7$& $82.4\pm 5.6$ \\
Diverse-GK & $2.06\pm 1.76$        &  {\color{red} $95.0\pm 1.8 $} & $93.6\pm 2.2 $\\
Diverse-LK  & {\color{red}$1.71\pm 1.23$} & {\color{red} $95.0\pm 1.8$} &  {\color{red}$94.0\pm 1.5$}       \\
\hline
\end{tabular}

\end{small}
\end{center}
\vskip -0.2in
\end{table}

\begin{table}[t]
\caption{Timing results and plan success rates (SR) for pouring task with a holder around the picked cup. The runtime only includes runs that successfully completed within 60s.}
\label{tb:timingpour}
\vskip -0.2in
\begin{center}
\begin{small}
\begin{tabular}{lccc}
\hline
\abovestrut{0.20in}\belowstrut{0.10in}
Method & Runtime (s) & 60s SR  (\%) & 6s SR (\%) \\
\hline
\abovestrut{0.20in}
Adaptive   & $5.79\pm 11.04$  & $51.4\pm 3.3$& $40.9\pm 4.1$ \\
Diverse-GK & $3.90\pm 5.02$        &  $56.3\pm 2.0 $ & $46.3\pm 2.0 $\\
Diverse-LK  & $4.30\pm 6.89$ & {\color{red} $59.1\pm 2.6$} &  {\color{red}$49.1\pm 2.6$}       \\
\hline
\end{tabular}
\end{small}
\end{center}
\vskip -0.2in
\end{table}
}

%
%
%

\subsection{Integrated system} \label{sec:integrated}

Finally, we integrated the learned sampling models for the \pddl{pour} and \pddl{scoop} actions with 7 pre-existing robot operations (\pddl{move}, \pddl{pick}, \pddl{place}, \pddl{fill}, \pddl{push}, \pddl{stir})
in a domain specification for \stripstream.


As a demonstration, we give the robot a goal which is to ``prepare'' a cup of coffee with cream and sugar. To achieve this goal, the robot must pour coffee into the white bowl, scoop sugar from the red bowl and dump it into the white bowl, and stir the while bowl, and return to its initial configuration. While doing this, the robot also needs to plan its path in a way that avoids all obstacles. Figure~\ref{fig:settings} displays the robot solving a {\em Kitchen3D} ({\em left}) and real-world ({\em right}) version of this task. 
See {\small \url{https://tinyurl.com/lis-ltamp}} for a video of a real-world robot solving this task.



These results illustrate the ability to augment the existing
competences of a robotic system (such as moving while avoiding collisions)
with new sensorimotor primitives by learning probabilistic models of
their preconditions and effects and using a state-of-the-art
domain-independent continuous-space planning algorithm to combine them
fluidly and effectively to achieve complex goals.


%% file: real_world.tex






\section{Real-world experiments}
\label{sec:real-world}


We applied our learning and planning framework to several real-world problems to demonstrate its sample efficiency and ability to generalize over a diverse set of planning scenarios.
We use the same set of PyBullet primitive implementations as in the {\em Kitchen3D} simulation. 
An open-source implementation of system is available at {\small \url{https://github.com/caelan/LTAMP}}.

\subsection{Perception}

We assume that the objects rest on a single table and are fully observable from a Kinect RGB-D camera mounted on the robot's head. 
Additionally, we assume that we have an approximate 3D mesh model for each object on the table.
We created crude mesh models for each bowl and cup, derived solely from rough base diameter, top diameter, and height measurements.

\begin{figure}
    \centering
    \includegraphics[width=0.55\columnwidth]{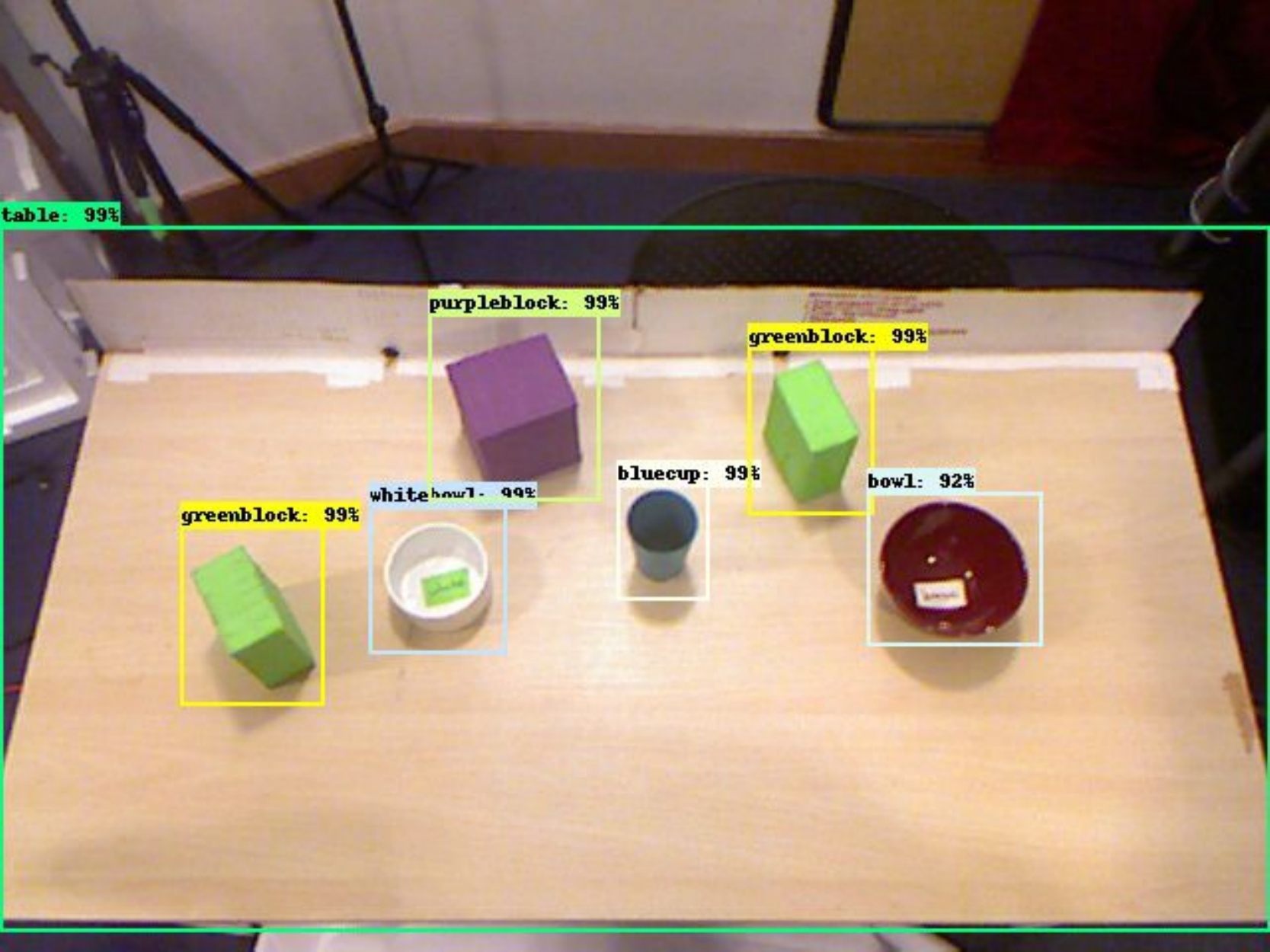}
    \includegraphics[width=0.44\columnwidth]{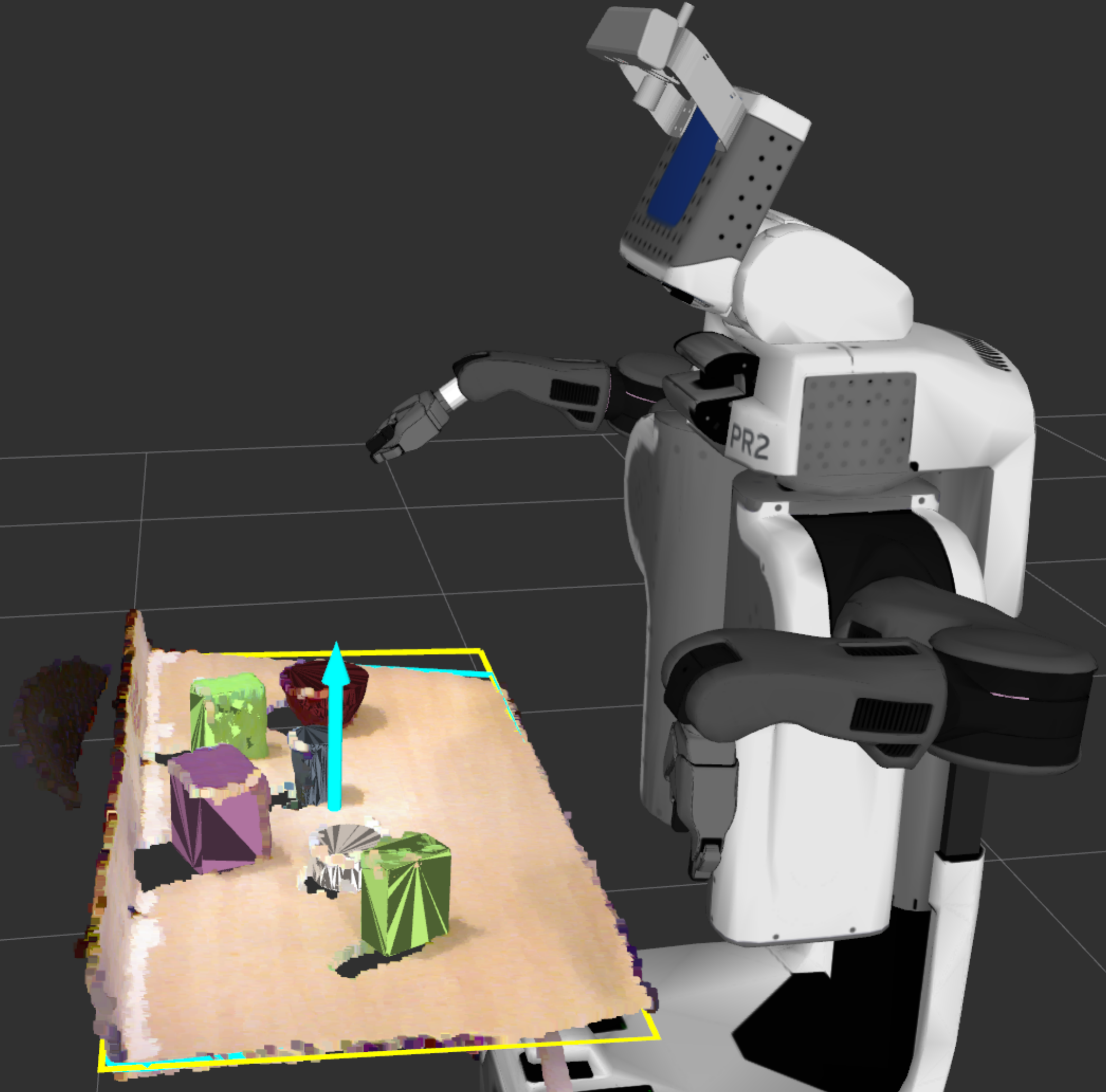}
    \caption{{\em Left:} R-CNN table and object 2D visual bounding box detections. 
    {\em Right:} the estimated table surface and object poses visualized in RViz. 
    The table surface plane normal is the blue vector, the yellow rectangle is the axis-aligned bounding of the surface within the plane, and the blue polygon is the convex hull of the surface within the plane.
    The colored mesh of each registered object pose is overlaid on the point cloud, demonstrating the accuracy of the position and orientation estimates.
    } 
    \label{fig:tensorflow}
\end{figure}

We use visual data to recognize and coarsely locate objects on the table and depth data to identify the table surface and localize objects.
We use the Faster R-CNN~\citep{ren2015faster} visual object detector~\citep{huang2017speed} implemented in TensorFlow~\citep{abadi2016tensorflow} to predict labeled 2D bounding boxes for the table and each object model.
We pretrained the R-CNN on the COCO data set~\citep{lin2014microsoft} and then trained it on 447 hand-annotated image arrangements of our table, blocks, cups, and bowls.
An example set of detections is displayed in figure~\ref{fig:tensorflow} ({\em left}).

We use the detection information to isolate subsets of the point cloud contained within the 3D view cone corresponding to each 2D bounding box.
For each detected table, we use the Point Cloud Library's (PCL)~\citep{Rusu_ICRA2011_PCL} random sample consensus (RANSAC)~\citep{fischler1981random} plane estimator to obtain the equation of its plane as well as the 2D convex hull of its points when projected into the plane.
We filter planes with normal vectors that significantly deviate from the global z-axis.
Then, we prune detected objects that are not supported by the estimated table plane.
For the remaining detected objects, we perform pose registration on the point cloud contained within its cone using the mesh model corresponding to the predicted label.
We use a pose estimator built by~\cite{GloverThesis}, which performs a randomized optimization over object placements resting normal to the plane, minimizing the distance between the observed point cloud and a point cloud derived from the mesh.
Figure~\ref{fig:tensorflow} ({\em right}) shows the estimated table surface plane as well as the detected objects at their estimated poses.

\begin{figure}
    \centering
    \includegraphics[width=\columnwidth]{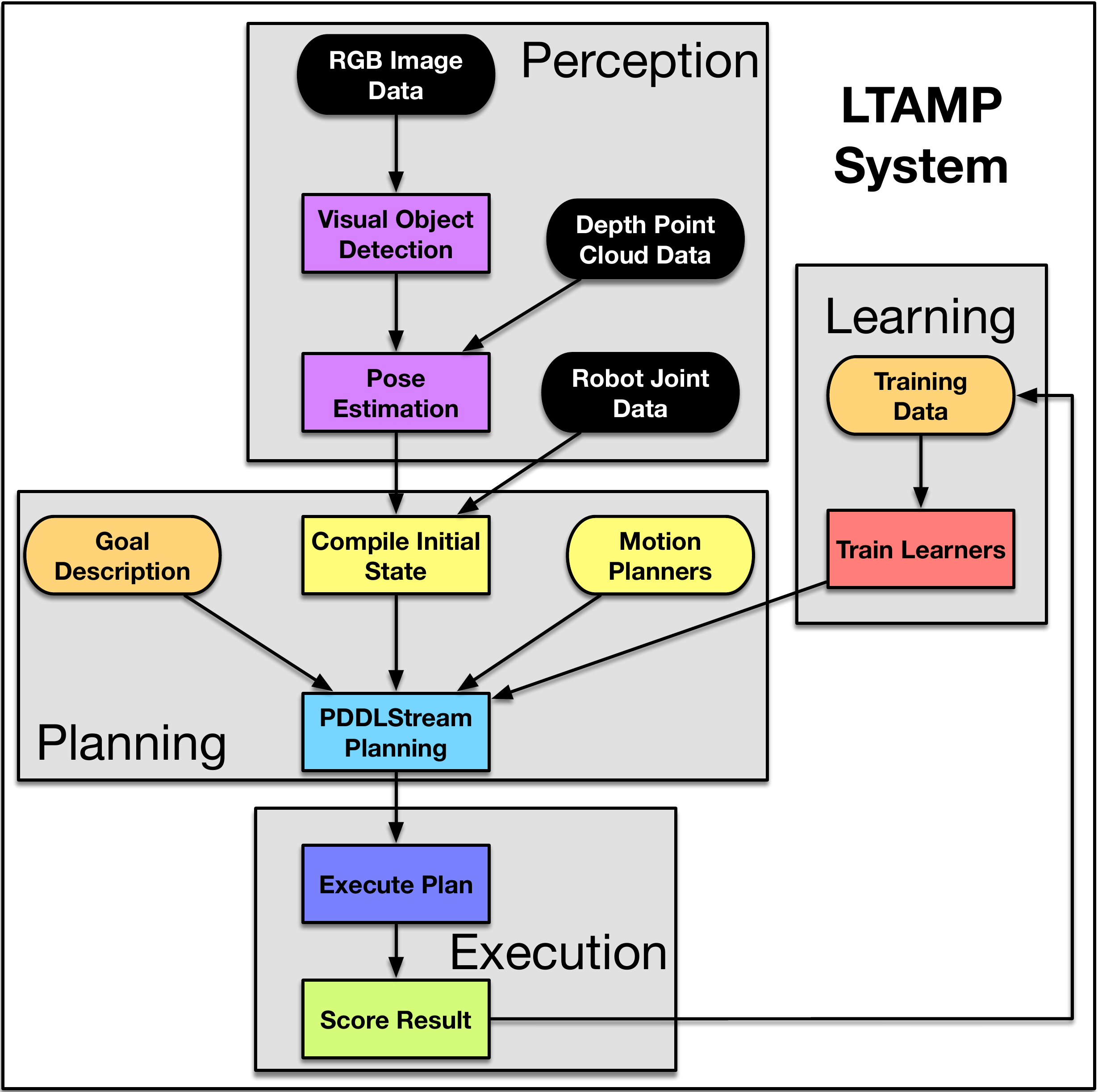}
    \caption{A flowchart that decomposes our real-world system into four components: perception, learning, planning, and execution.}
    \label{fig:system-flowchart}
\end{figure}

Figure~\ref{fig:system-flowchart} provides a flowchart of our system.
We use the Robot Operating System (ROS)~\citep{ROS} to relay plane and pose estimates to the Python planning engine where they are treated as the ground-truth environment.
The inputs are RGB, depth, and joint data as well as a goal description.
The perception subsystem is used to populate an estimate of the initial state.
The planning subsystem consumes this estimate along with the goal description, motion planning primitives, and the current Gaussian Process models.
After receiving a single observation, the planner solves the corresponding problem and outputs a path that specifies a sequence of robot arm joint positions.
After solving for a plan, the execution subsystem performs local feedback control to follow the plan, scores the final world state, and adds the result to the training data set.
After interpolation using cubic splines, the resulting trajectory is executed in an entirely open-loop manner at the high level.

\subsection{Data collection}


We use small bead-like objects as the material to be poured or scooped.
Specifically, we use red wooden objects for pouring and dried chickpeas for scooping. 
In our training setup, we place bowls and cups on USB scales to estimate the particle mass contained within each object both before and after execution.
The USB scales are directly connected to our computer to provide automated real-time mass readings.
We subtract the mass of the bowl or cup in order to obtain the mass of the particle contained within an object. 

In simulation, the world can be directly assigned to be the state prior to executing a skill.
However, in the real world, the robot must also act to set up a skill ({\it e.g.} grasping the cup), act to score the skill ({\it e.g.} dumping the spoon's contents into a bowl), and reset the scene for its next trial.
It is critical that the robot respects kinematic, joint-limit, collision, and spillage orientation constraints in order to ensure that these actions are likely to be successfully executed.
We use our planner to plan paths that respect these constraints and facilitate data collection.
Thus, we are performing both {\em learning for planning} and {\em planning for learning}.

We formulate collecting one trial as a planning problem where the planner is restricted to use a single control parameter that is selected either uniformly at random or by a \gp{} active learner.
Otherwise, the planner has the freedom to select the other plan parameters, such as the grasp used to pick the cup.
For both pouring and scoring, we enforce that the cup finishes at its initial pose and that the robot finishes at its initial configuration.
By planning to reset the scene, we avoid the need to teleoperate the robot or manually extract an object from the robot's gripper.
Additionally, this prevents the robot arm from self-occluding the table during its next observation.
As a result, the only manual actions that a human must perform are cleaning up spilled particles and swapping the placed objects that will be used on next trial.

For pouring, the robot picks up the cup, attempts to pour its contents into the bowl using the sampled control parameter, places the cup back at its initial pose, and returns to the initial configuration. 
This results in the following sequence of operators:
$[\pddl{move}, \pddl{pick}, \pddl{move}, \pddl{pour}, \pddl{move}, \pddl{place}, \pddl{move}]$.
The fraction of particles that were successfully poured is the ratio of the final bowl particle mass to the initial cup particle mass.
For scooping, the spoon starts in the robot's gripper, at an approximate grasp.
The robot scoops the contents of the bowl using the sampled control parameter, dumps the spoon's contents into the measurement bowl, and returns to the initial configuration.
This results in the following sequence of skills: $[\pddl{move}, \pddl{scoop}, \pddl{move}, \pddl{pour}, \pddl{move}]$.
The fraction of particles that were successfully scooped is the ratio of the final measurement bowl particle mass to the mass capacity of the spoon, which is measured offline.
As a result, only one scale is required when scoring a scoop.
Finally, we use the plan constraint compilation procedure of~\cite{garrett2020online} to enforce that each plan exactly executes the prescribed sequence of skills, preventing it from considering plans that, for example, perform two scoops.
Ultimately, the planner is typically able to find a solution in less than 15 seconds.
See the ``Learning to \{Pour, Scoop\}: Data Collection'' videos at {\small \url{https://tinyurl.com/lis-ltamp}} for demonstrations of the robot collecting data using this pipeline.

\begin{figure}
    \centering
    \includegraphics[width=0.58\columnwidth]{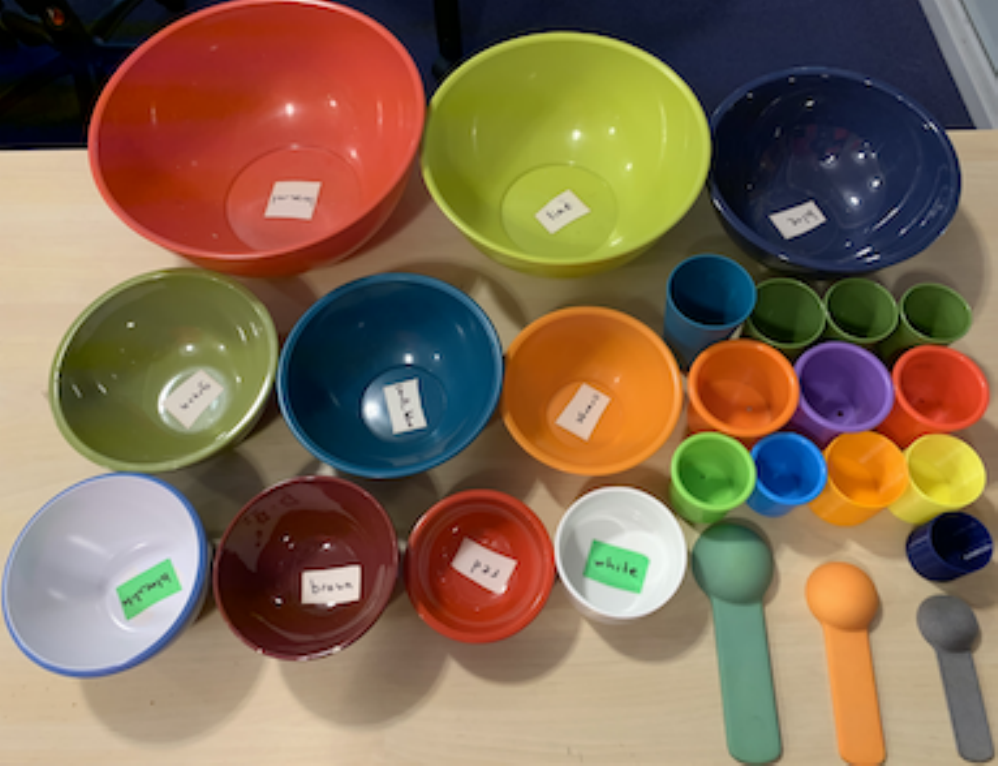}
    \includegraphics[width=0.41\columnwidth]{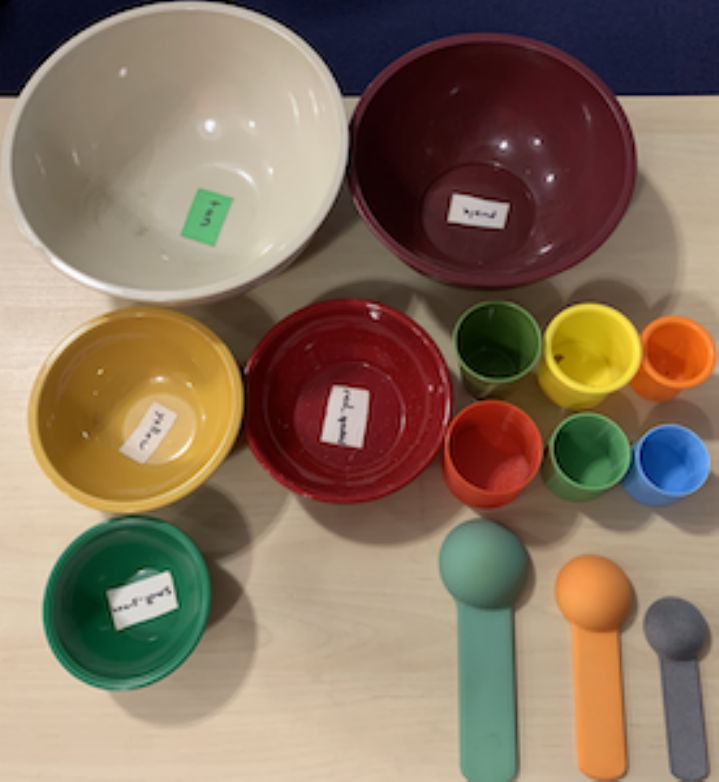}
    \caption{{\em Left}: the training set of 10 bowls, 12 cups, and 3 spoons. {\em Right}: the testing set of 5 bowls, 6 cups, and the same 3 spoons.} 
    \label{fig:objects}
\end{figure}

We trained the learners on a set of training objects and evaluated the learners on a set of unseen testing objects.
Several of the bowls and cups are from the YCB dataset~\citep{calli2015ycb}.
The objects range in both size, mass, and material (ceramic, plastic, and 3D printed).
We trained the learners on 10 bowls and 12 cups of varying sizes.
We tested the learners on 5 bowls and 6 cups of varying sizes.
We used the same set of 3 spoons both during training and testing.
The number of pouring contexts is the number of bowl and cup pairs while the number of scooping contexts is the number of bowl and spoon pairs.
Figure~\ref{fig:objects} displays the set of training and testing objects.
For each trial, we sampled the objects (and as a result the context) uniformly at random.

\begin{figure}[ht]
    \centering
    \includegraphics[width=0.59\columnwidth]{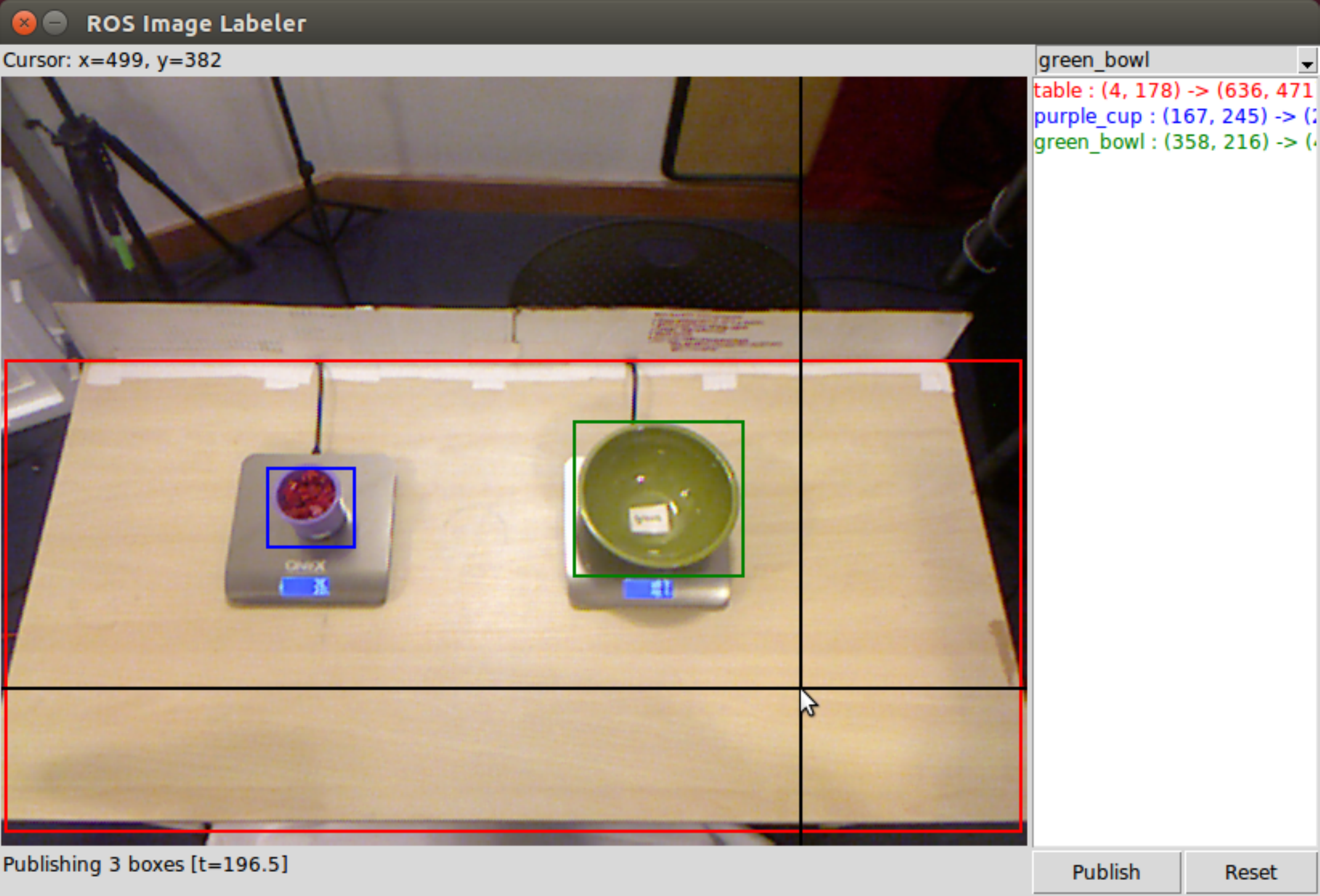}
    \includegraphics[width=0.4\columnwidth]{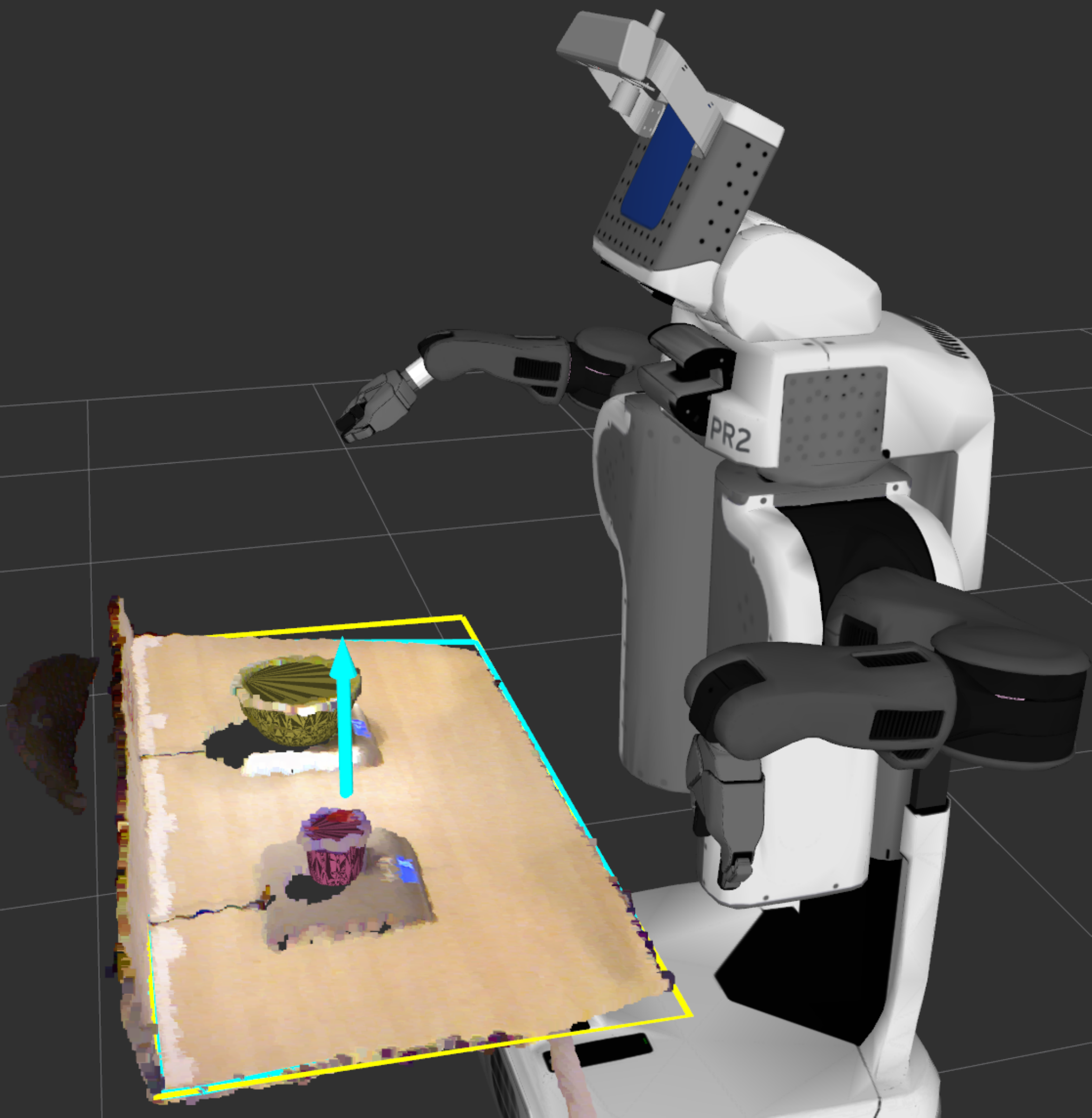}
    \caption{{\em Left:} the manual bounding box labeling tool that replaces the R-CNN predictions during training. {\em Right:} the corresponding estimated table surface and object poses visualized in RViz. See figure~\ref{fig:tensorflow} ({\em right}) for a description of the RViz markers. 
    } 
    \label{fig:pour-collection}
\end{figure}

To quickly collect data incorporating new objects without needing to retrain the R-CNN, we developed a user interface (UI) that allows a user to ``replace'' the object detector by manually annotating object bounding boxes online.
These labeled bounding boxes are then sent to the point-cloud pose-estimation system as normal.
The labels of each bounding box can also be changed programmatically, enabling the data collection program to update their values given the next selected cup and bowl pair.
Figure~\ref{fig:pour-collection} displays the UI tool and visualizes the corresponding table plane, registered bowl mesh, and registered cup mesh for a pouring trial.

\subsection{Training}
\label{sec:real-training}

We compared the sample efficiency of \gp{}s trained both with and without active learning in this real-world setting.
Both \gp{}s used the MLP kernel as well as the parameterization in section~\ref{sec:parameterization}.
We initially seeded each learner with 50 training examples gathered by sampling context and control parameters uniformly at random.

\begin{figure}[ht]
    \centering
    \includegraphics[width=0.45\columnwidth]{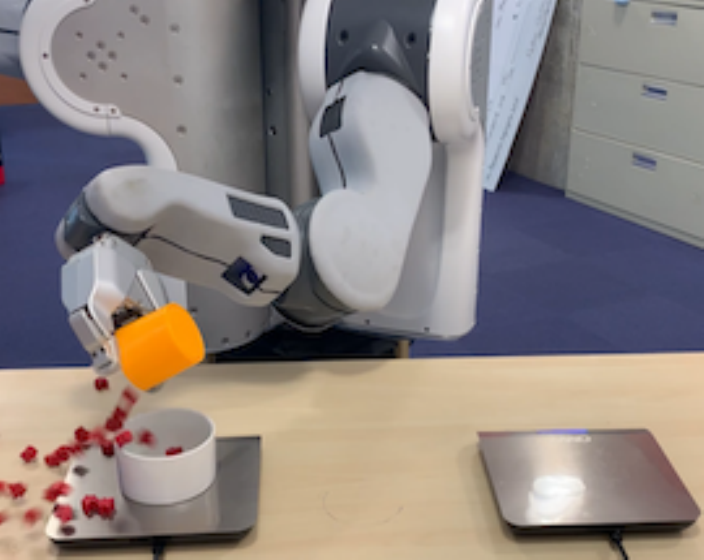}
    \includegraphics[width=0.54\columnwidth]{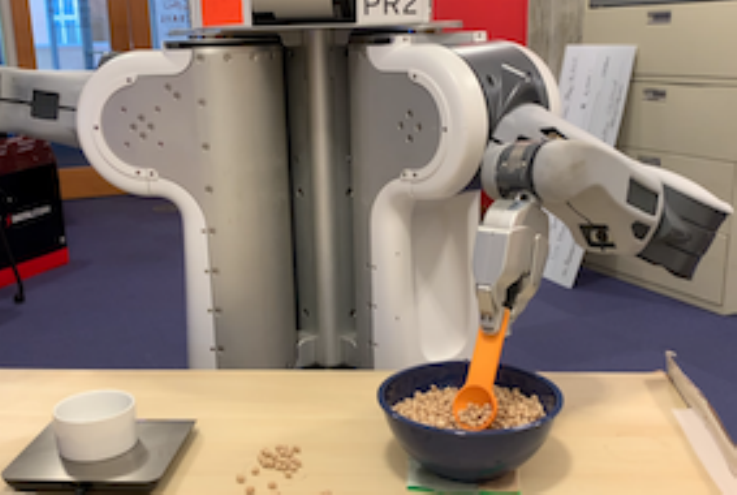}
    \caption{The robot executing actively selected control parameters. The learners intentionally explore control parameters that are near the boundary of the super-level set. {\em Left}: the selected pour successfully produces several particles in the bowl but also spills many particles. {\em Right}: the selected scoop is able to scoop some particles, but the spoon still has the capacity to hold more.}
    \label{fig:active}
\end{figure}

\begin{figure}[ht]
    \centering
    \includegraphics[width=0.31\columnwidth]{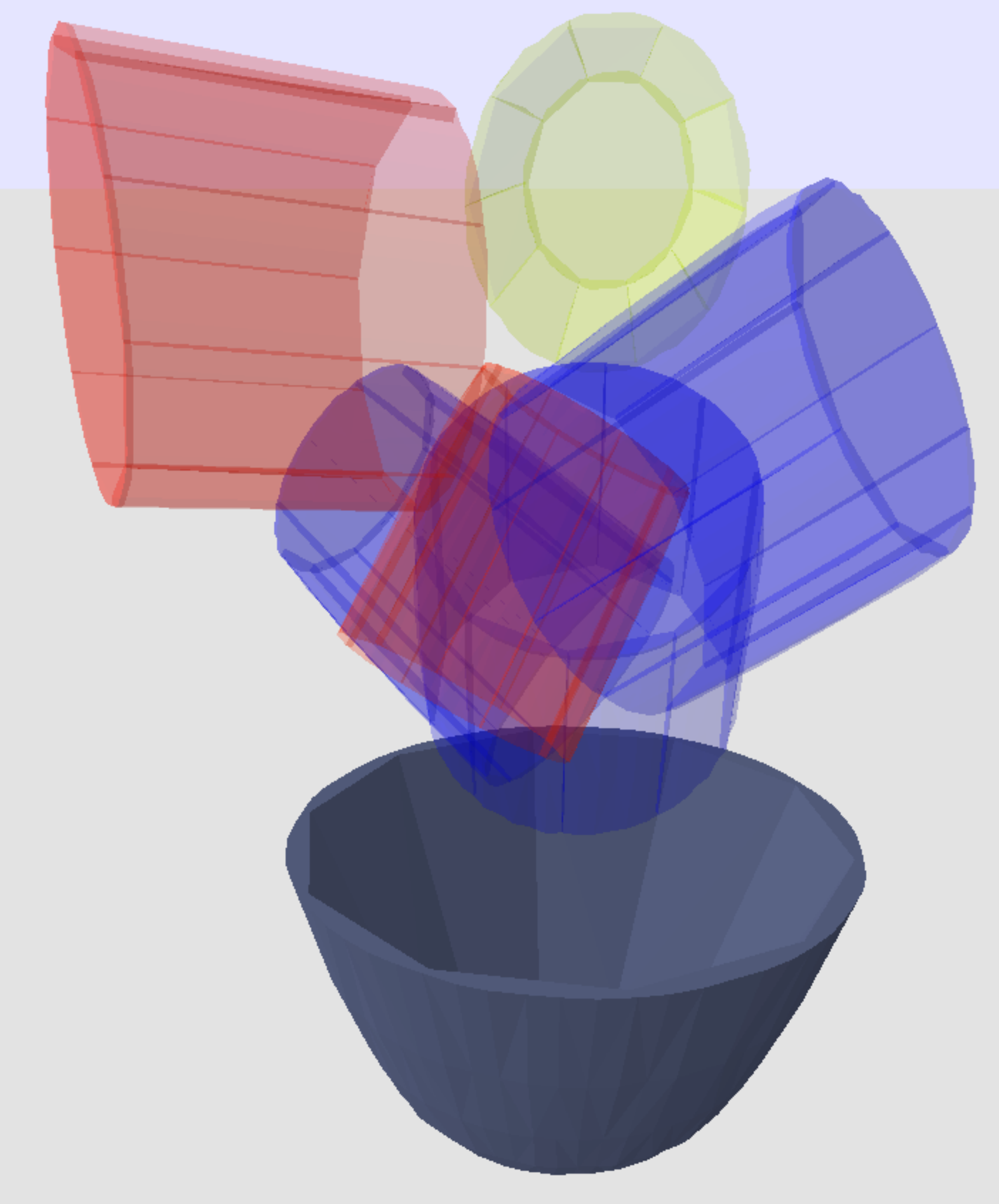}
    \includegraphics[width=0.33\columnwidth]{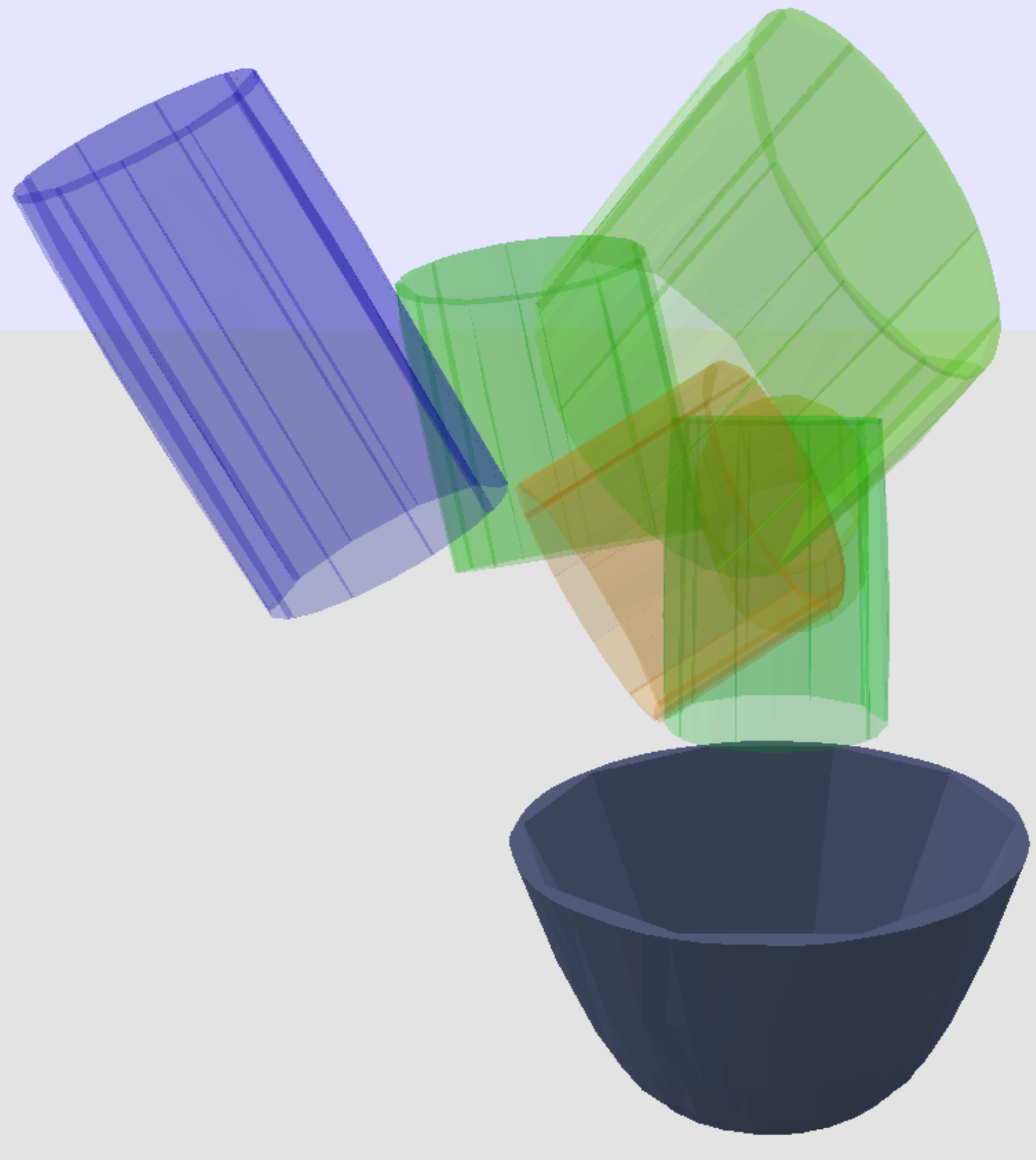}
    \includegraphics[width=0.34\columnwidth]{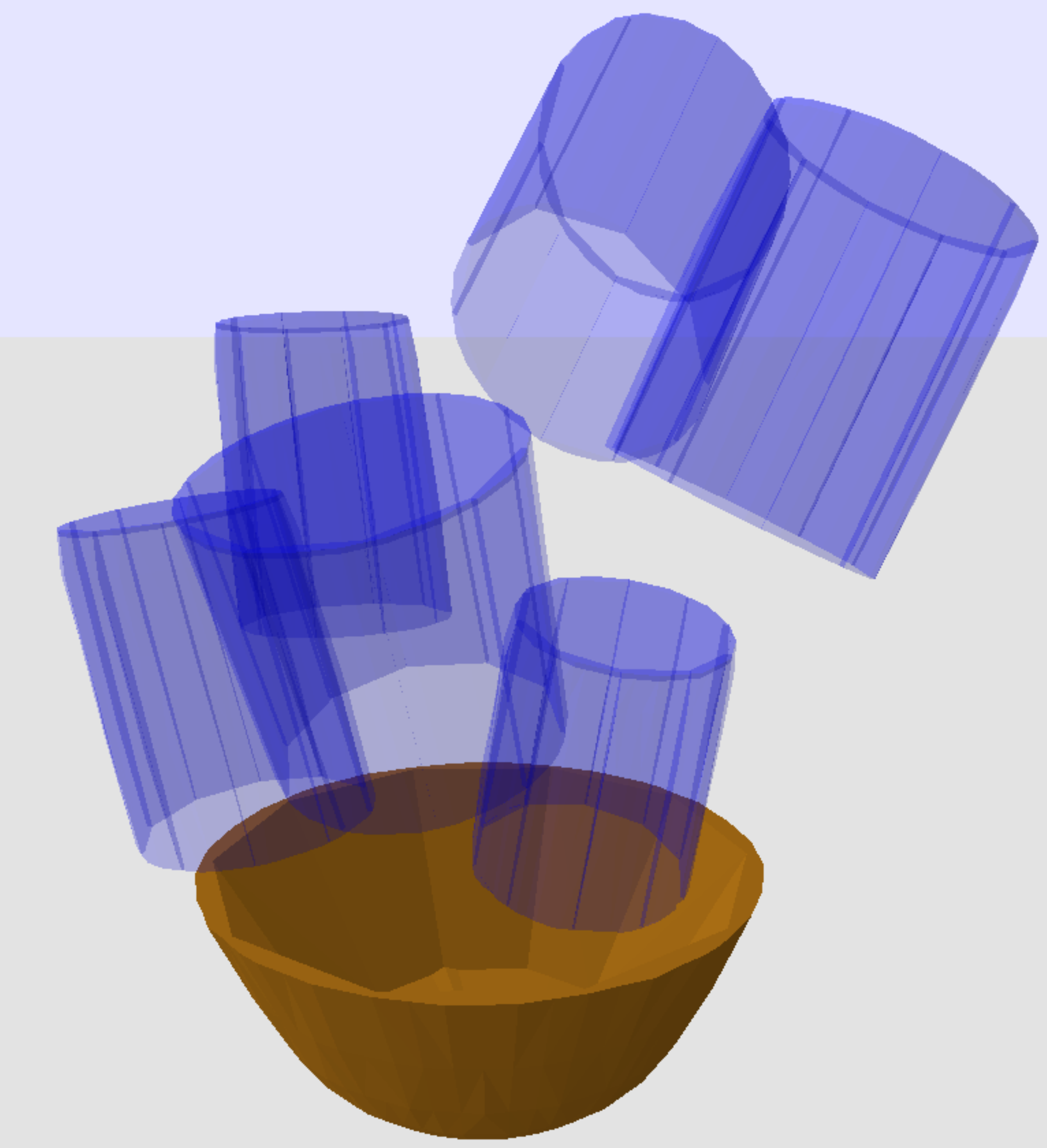}
    \caption{The distribution of selected real-world pours visualized in simulation per selection policy. The measured score of the pour is visualized by the hue of the cup, where red pours are the least successful, green pours are near the zero level-set, and blue pours are most successful. 
    {\em Left}: pours selected {\em uniformly at random} for a single training bowl. {\em Center}: pours selected {\em actively} for the same training bowl. Many selected pours are green, indicating that the learner is exploring the decision boundary. {\em Right}: the most confident pours for a single testing bowl. Each pour is blue, which indicates that all pours were successful.} 
    \label{fig:pour-dist}
\end{figure}

Figure~\ref{fig:active} visualizes the robot executing pour and scoop actions selected using active learning.
Both of these trials demonstrate borderline success, which is consistent with the robot selecting control parameters near the zero level set.
Figure~\ref{fig:pour-dist} visualizes selected pours and their scores overlaid on a particular bowl in our {\em Kitchen3D} simulator.
Red cups indicate pours with negative scores and blue cups indicate pours with positive scores.
The three images compare selections made uniformly at random, by the \gp{} active learner, and by the final trained \gp{} learner on test objects.
Many of the active learner's selections are green, indicating that they are near the zero level set boundary.

\begin{figure*}[ht]
    \centering
    \includegraphics[width=0.49\textwidth]{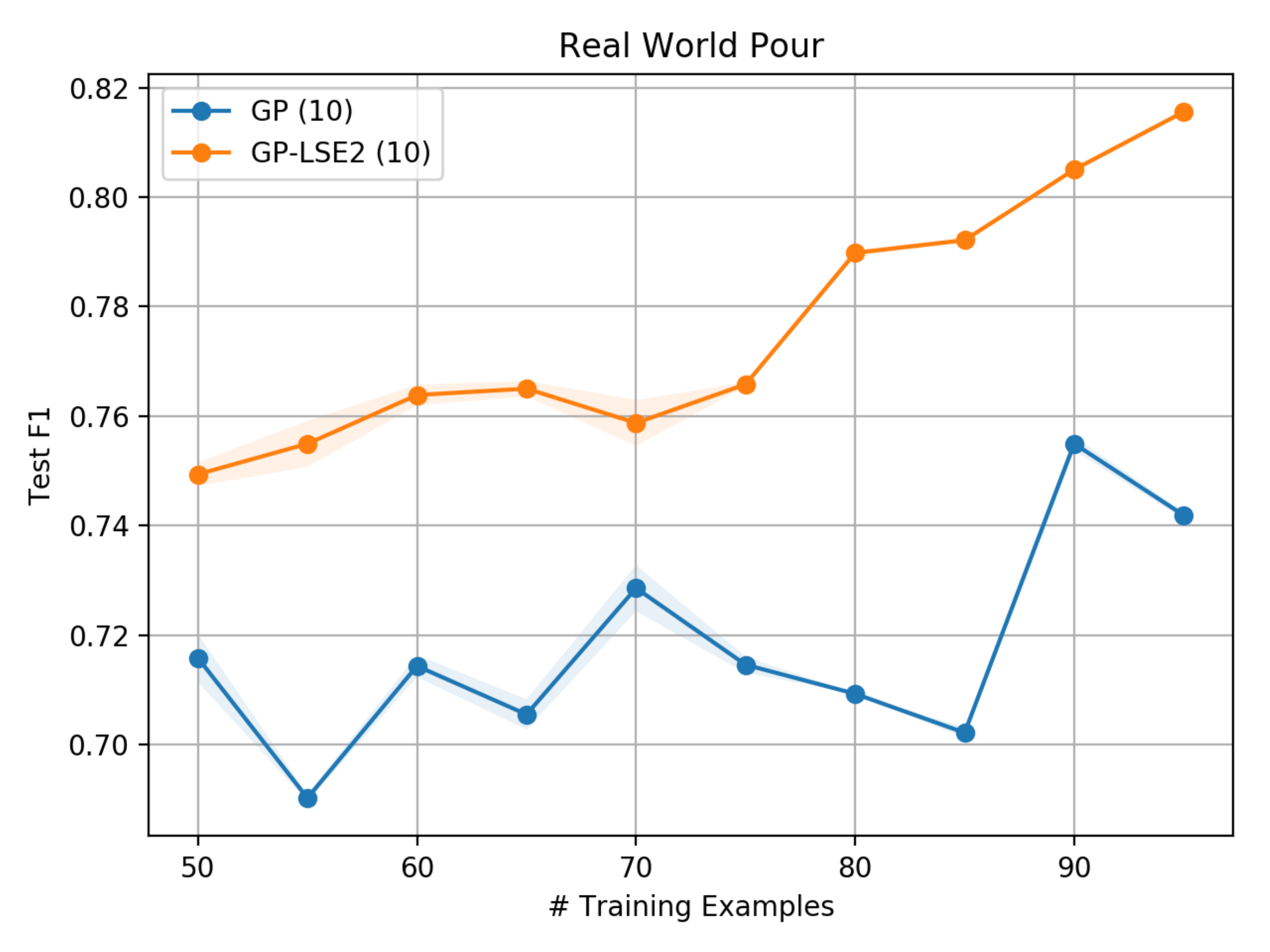}
    \includegraphics[width=0.49\textwidth]{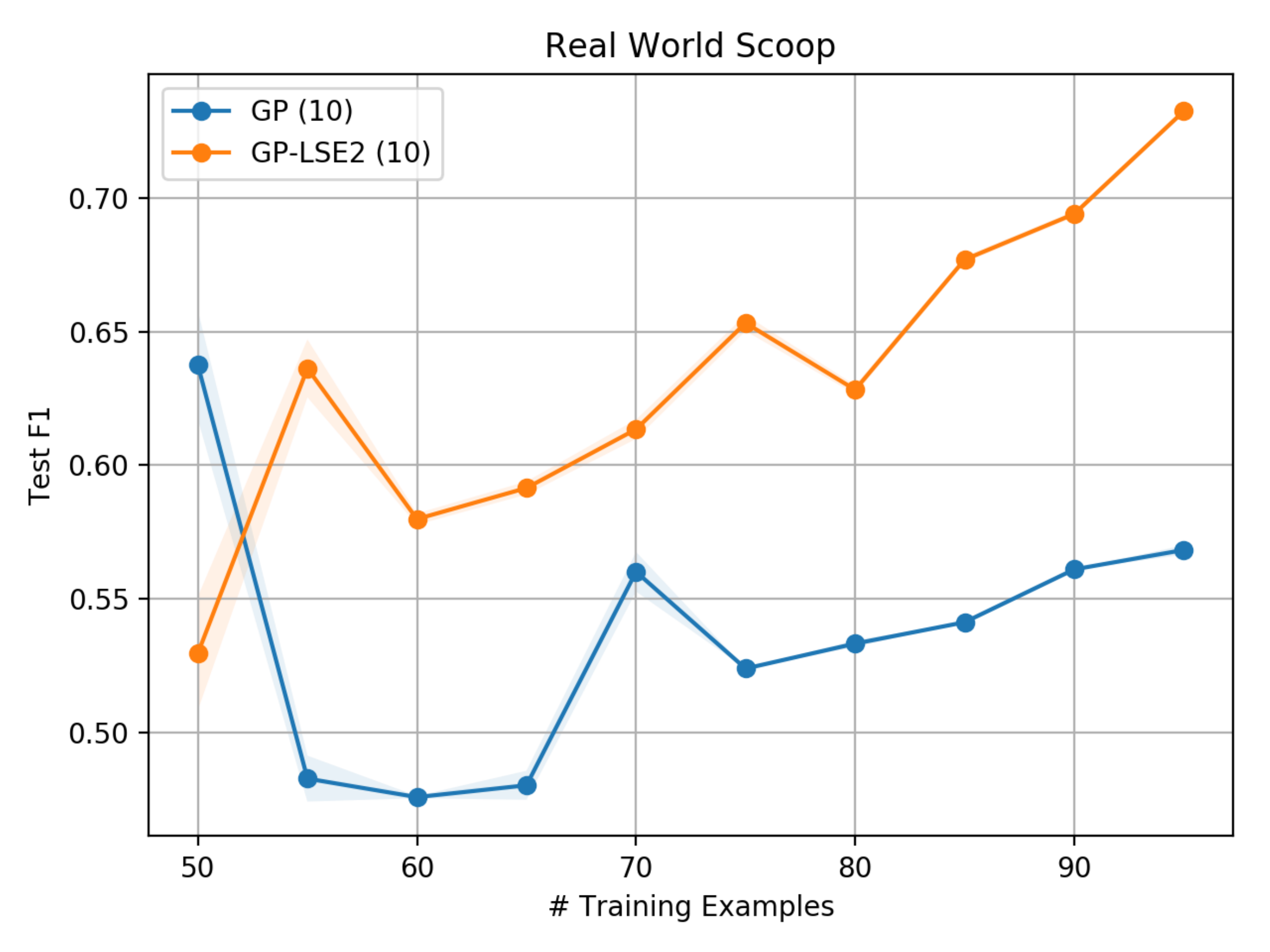}
    \caption{The F1 score for a \gp{} trained both {\em without} (GP) and {\em with} active learning (GP-LSE2) on the testing objects. 
    Each \gp{} was} initially trained with 50 examples collected uniformly at random on training objects. 
    \label{fig:real-pour}
\end{figure*}

During training, we tested how well the learners were able to {\em classify} successful pours and scoops.
This allowed us to obtain an measure of how well the \gp{} was learning without needing to periodically evaluate on testing data during training.
We collected a test data set of 133 pour and 81 scoop examples, sampled uniformly at random on the test objects.
Figure~\ref{fig:real-pour} displays the F1-score learning curves of the \gp{} learners without and with active learning on this data set.
The $1/4$ standard deviation error bounds result from retraining each \gp{} 10 times on the {\em same} data, to account for the stochastic hyper-parameter optimization when retraining.
Active learning enables the \gp{} learner to more quickly classify successful pours and scoops.

\subsection{Most confident prediction}

Recall that our ultimate objective is to sample control parameters that the learner confidently believes to lie in the super-level set. 
We compared the most confident predictions of a \gp{} trained on 50 training examples sampled uniformly at random, 100 (96 for scooping) training examples sampled uniformly at random, and 50 training examples sampled uniformly at random followed by 50 (46 for scooping) selected actively.
We performed one trial per unique bowl-cup and bowl-spoon test pair, resulting in 30 pours per learner and 15 scoops per learner.

Table~\ref{tb:best} lists the performance of each learner when making its most confident prediction on the test objects.
{\em Valid} is the percentage of sampled control parameters for which full motions of the robot could be found.
Recall that the learner may predict control parameters that cannot be safely executed by the robot, such as pours in the interior of a bowl that do not admit any collision-free grasps.
{\em Success} is the percentage of sampled control parameters that were in the super-level set.
{\em Filled} is the percentage of the cup or spoon's capacity was filled.
The active learner outperforms the non-active learners in each metric both for pouring and scooping.

\begin{table}
\begin{center}
\resizebox{\columnwidth}{!}{%
\begin{tabular}{llccc}
\hline
\abovestrut{0.15in}\belowstrut{0.10in}

&   & Batch $N_{50}$ & Batch $N_{100}/N_{96}$ & Active $N_{100}/N_{96}$  \\
\hline
\abovestrut{0.10in}
\parbox[t]{0mm}{\multirow{3}{*}{\rotatebox[origin=c]{90}{Pour}}} 
& Valid (\%) & $0.867$ & $0.833$ & {\color{red} $0.933$} \\
& Success (\%) & $0.923$ & $0.920$ & {\color{red} $1.000$} \\
& Filled (\%) & $0.964$ & $0.958$ & {\color{red} $0.994$} \\
\hline
\abovestrut{0.10in}
\parbox[t]{0mm}{\multirow{3}{*}{\rotatebox[origin=c]{90}{Scoop}}} 
& Valid (\%) & $0.600$ & $0.933$ & {\color{red} $0.933$} \\
& Success (\%) & $0.889$ & $0.786$ & {\color{red} $0.929$} \\
& Filled (\%) & $0.829$ & $0.861$ & {\color{red} $0.952$} \\
\hline
\end{tabular}
}
\end{center}
\caption{When evaluating pours and scoops during testing, the percentage of them that admitted a full robot plan ({\em valid}), the percentage of them that were in the super-level set ({\em successful}), and the average mass inside the scoring bowl relative to the capacity of the involved cup or spoon.
The best value of each metric across the three learners is indicated in red.
} 
\label{tb:best}
\vskip -0.1in
\end{table}



\subsection{Integration}




Finally, we used our learned pouring model within our planner to solve challenging real-world multi-step manipulation problems.
We experimented with two problems where the robot must combine its learned pouring models with motion planners that respect kinematic and collision constraints.
In each problem, the blue cup is initially holding ``liquid'' particles, and the goal is for the brown bowl to instead contain the particles.
The robot must additionally return to its initial configuration with both grippers empty.

\begin{figure*}[ht]
    \vskip 0.1in
    \centering
    \includegraphics[width=0.99\textwidth]{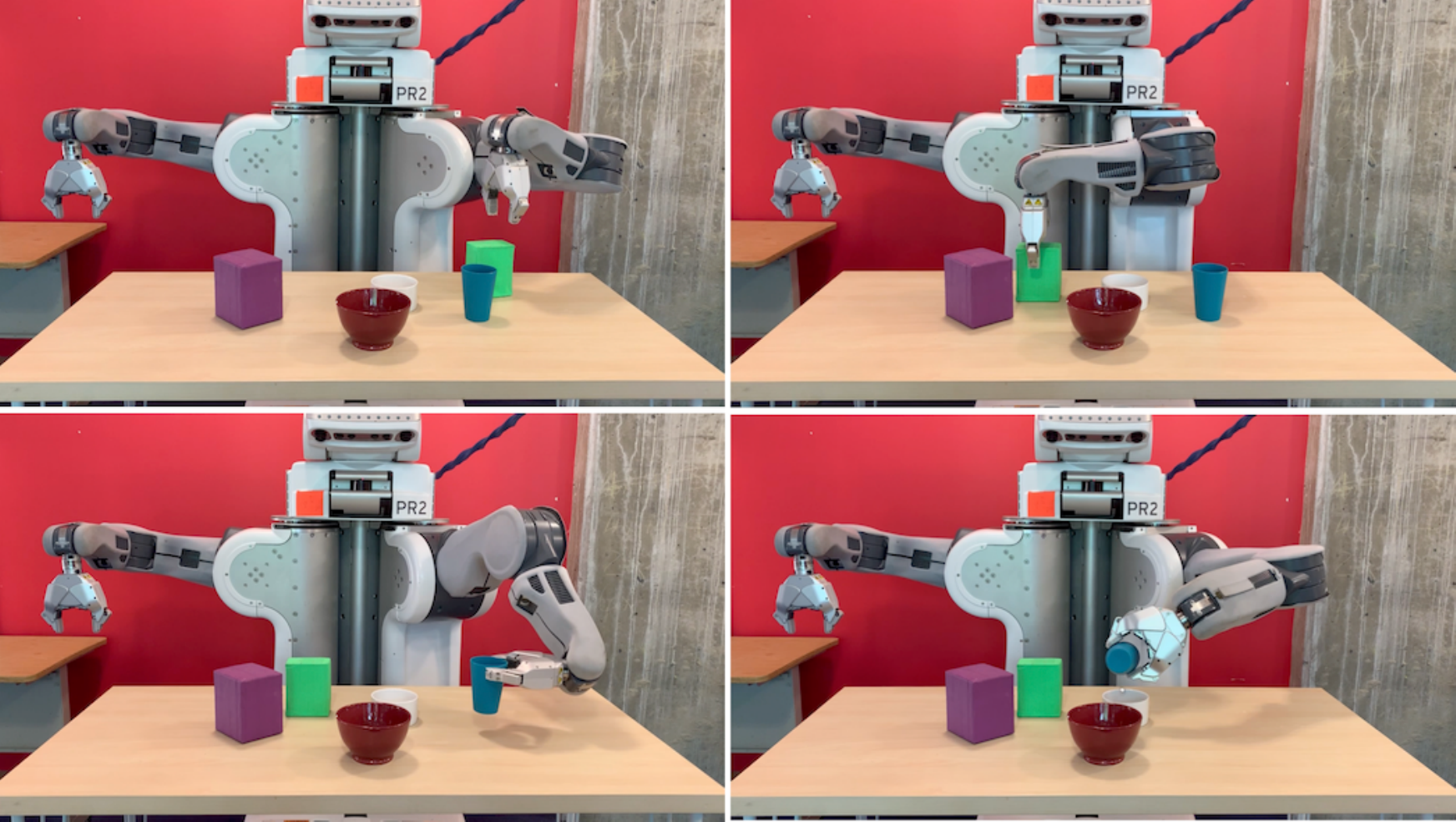}
    \caption{
    The goal is for the particles in the blue cup to be in the white bowl.
    Because the green block obstructs reachable side grasps for the blue cup, the planner automatically plans to relocate the green block before picking the blue cup and pouring its contents into the white bowl. 
    From {\em left-to-right} and {\em top-to-bottom}, the robot picking the green block, the robot placing the green block, the robot picking the blue cup, and the robot pouring the blue cup's contents into the brown bowl.
    }
    \label{fig:pr2} 
    \vskip -0.1in
\end{figure*}

Figure~\ref{fig:pr2} demonstrates the robot solving the first problem.
In this problem, the robot is unable to find a kinematically feasible way of picking the blue cup without colliding with the green block.
Thus, the robot plans to pick the green block and finds a placement for it that allows for the blue cup to be picked.
Afterwards, the robot can now safely pick the blue cup and pour its contents into the brown bowl.
Finally, the robot places the blue cup and moves its left arm back to its initial configuration.
Critically, the robot finds a grasp for the cup that both admits a pour path that is  predicted to be successful  and admits a collision-free pick path when the green block is moved.
See {\small \url{https://youtu.be/a5F1hce4o0o}} for a video of the robot executing this solution.

\begin{figure*}[ht]
    \vskip 0.1in
    \centering
    \includegraphics[width=0.49\textwidth]{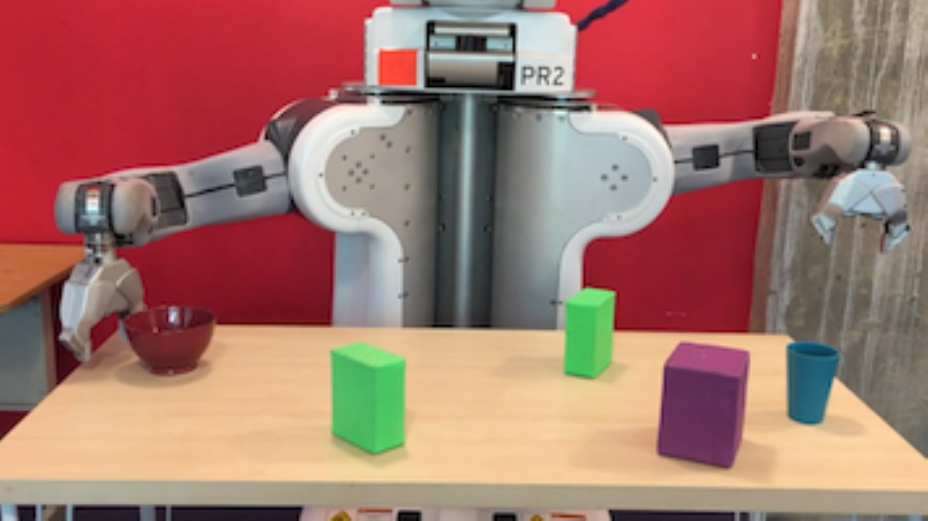}
    \includegraphics[width=0.49\textwidth]{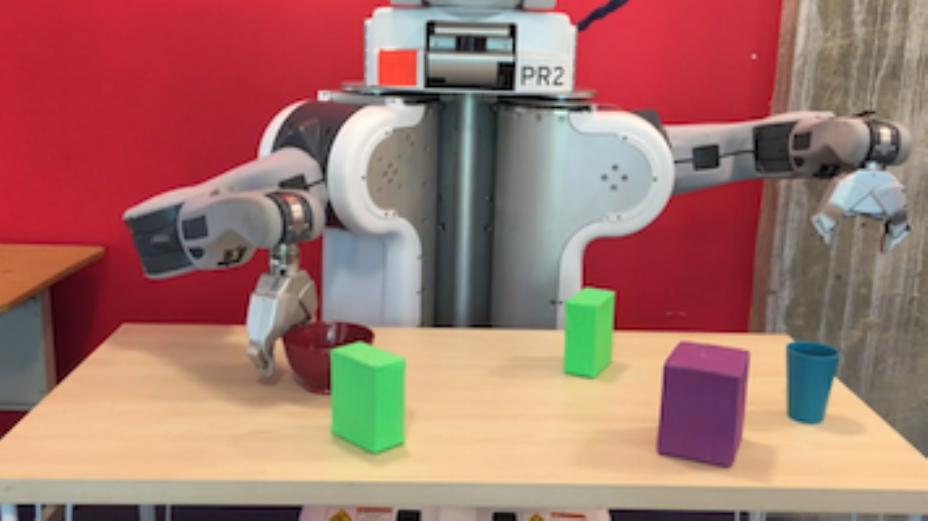}
    \includegraphics[width=0.49\textwidth]{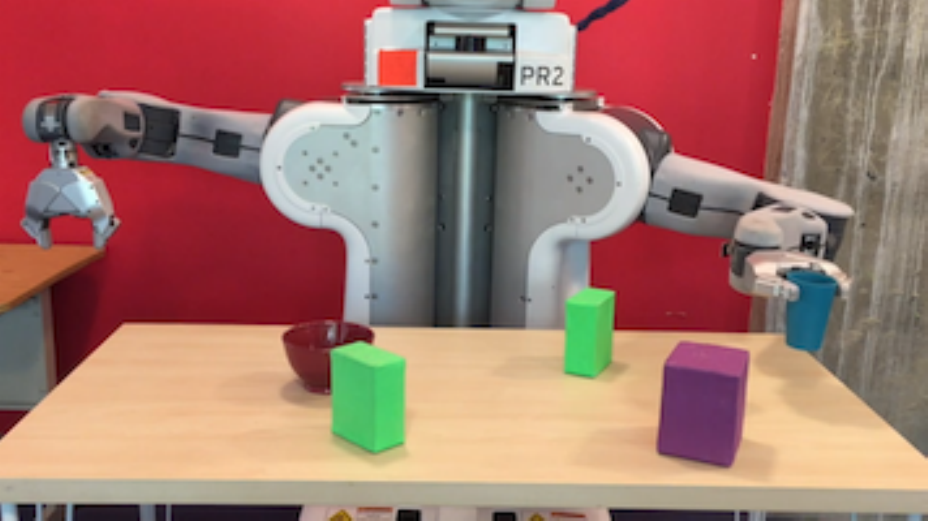}
    \includegraphics[width=0.49\textwidth]{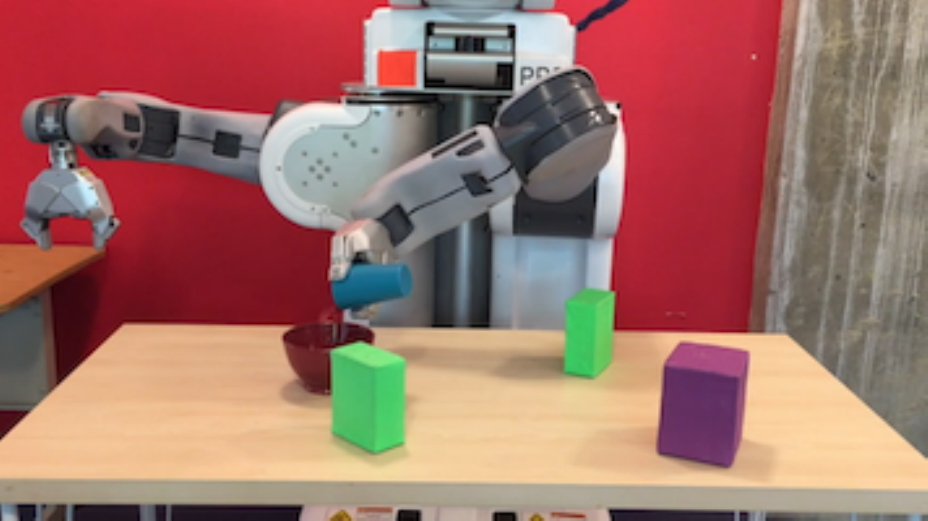}
    \caption{The goal is for the particles in the blue cup to be in the brown bowl.
    Because the robot cannot reach the brown bowl with its left arm, the planner automatically plans to push the bowl towards the center of the table, so it can then pour the blue cup's contents into the brown bowl.
    From {\em left-to-right} and {\em top-to-bottom}, the state before the robot pushes the brown bowl, the resulting state after the push, the robot picking the blue cup, and the robot pouring the blue cup's contents into the brown bowl.
    }
    \label{fig:pr2-push} 
    \vskip -0.1in
\end{figure*}

Figure~\ref{fig:pr2-push} demonstrates the robot solving the second problem.
In this problem, the bowl starts at one side of the table while the blue cup starts on the other side.
Because neither arm can reach both objects, the robot must intentionally manipulate one of the objects with one arm to put it within reach other the other arm.
There are two high-level ways of accomplishing this.
The first requires picking up the blue cup with the robot's left arm and deliberately placing it near the middle of the table, within reach of the right arm.
The second requires pushing the brown bowl with its right arm towards the middle of the table.
Although the robot's planning model can produce both solutions, the planner returned the second solution, likely because it uses fewer actions.
Once the brown bowl is within reach, the robot can successfully pour the contents of the blue cup into the bowl and return to its initial state.
Because the robot was initially kinematically unable to pour using its left arm, it intentionally identifies a pose that it can push the bowl to in order to be within reach.
See {\small \url{https://youtu.be/a5F1hce4o0o?t=43}} for a video of the robot executing this solution.

\section{Conclusion}

This paper addresses learning generative models for dynamic manipulation skills for use during multi-step manipulation planning.
We learn the conditions for which a pour or scoop manipulation skill is sufficiently successful using Gaussian processes.
This allows us to capture the uncertainty in the learner's model, enabling us to make risk-aware predictions and perform active learning to methodically select training examples that best reduce the model's uncertainty.
Through simulated and real-world experiments, we show that active learning reduces the number of robot trials required to learn a skill.
Additionally, we introduce methods for diversely exploring the set of successful pours or scoops.
This enables a planner to quickly find values that admit a full robot plan.
Our integrated planner combines learned models for pouring and scooping with conventional robotics operations, enabling it to generalize across a large set of challenging manipulation problems.

\subsection{Future work}

One important avenue for future work involves incorporating learner predictions into the action {\em cost} of the associated control parameter.
Costs could be derived from the expected score of the parameter or the probability that the parameter is in the super-level set.
There are several approximate methods for performing risk-aware deterministic planning with non-negative additive costs~\citep{garrett2020online}.
This would enable the planner to weigh the expected cost of executing a sequence of control parameters among several candidate plans.

Although we consider both learning in simulation and the real world, we have not addressed sim-to-real transfer; which may be useful in settings where a high-fidelity simulator is available.  In our preliminary investigation, we found that it was challenging to benefit from active learning when training real-world models on simulated data.
Intuitively, if simulation and the real-world mismatch, particularly with respect to which dimensions are most informative, the active learner may explore training examples that do not effectively decrease uncertainty in the model.  Ultimately our goal is to develop methods that can learn effectively from a few real-world samples, without the need to develop a simulation.  However, in settings where a simulator exists, further investigation of the effective integration of active learning and sim-to-real transfer is desirable.


Finally, this paper addresses deterministic planning and open-loop execution; however, the real-world is stochastic and partially observable.  Our current and future work involves learning models for stochastic manipulation actions and observation actions for use during belief-space planning~\citep{garrett2020online,IJRRBel}.

%% file: appendix.tex
\appendix

\section{Gaussian processes}\label{ssec:gp}

Gaussian processes (\gp{}s) represent distributions over functions and serve as a useful representation for Bayesian regression.
In a \gp{}, any finite set of function values has a multivariate
Gaussian distribution. 
We use $GP(\mu,k)$ to denote a \gp{} with mean function $\mu(\vx)$ and kernel function $k(\vx,\vx')$. 
Two frequently used stationary covariance kernel functions are the {\em squared exponential} and {\em Mat\'ern kernels}.
Let $r=(\vx-\vx')^\top(\vx-\vx')$. Then the squared exponential kernel is 
$$k_f(\vx,\vx') = \sigma_f^2 \mathrm e^{-\frac{1}{2\ell_f^2} r},$$ 
with a variance $\sigma_f^2$ and length scale hyper-parameter $\ell_f$. 
The Mat\'ern kernel is given by 
$$k_m(\vx,\vx') = \sigma_m^2\frac{2^{1-\xi}}{\Gamma(\xi)} (\frac{\sqrt{2\xi r}}{h})^{\xi}B_\xi(\frac{\sqrt{2\xi r}}{h}),$$ 
where $\Gamma$ is the gamma function, $B_\xi$ is a modified Bessel function.
Its hyper-parameters are $\sigma_m^2$, $l_m$ and a
roughness parameter $\xi$. 
Additionally, we consider the non-stationary {\em multi-layer perceptron kernel} (also called the neural network kernel)~\citep{neal2012bayesian,rasmussen2006gaussian}, which often better models discontinuous functions such as the score functions in Section~\ref{sec:kitchen3D}~\citep{5152677,doi:10.1177/0278364911421039},
\begin{equation*} \label{eqn:nn-kernel}
k_n(\vx,\vx')=\frac{2\sigma_n^2}{\pi} \text{sin}^{-1} {\frac{\tilde{\vx}^\top \vct{\Sigma}^2 \tilde{\vx}' }{\sqrt{ \tilde{\vx}^\top \vct{\Sigma}^2 \tilde{\vx} + 1} \sqrt{\tilde{\vx}'^\top \vct{\Sigma}^2 \tilde{\vx}' + 1}}}, 
\end{equation*}
where $\tilde{\vx} = [1, \vx]$. 
Its hyper-parameters are a diagonal covariance matrix $\vct{\Sigma}^2$ and variance $\sigma_n^2$.

Let $f$ be a true underlying function sampled from $\GP(0,k)$. 
Given a set of observations $\cd=\{(\vx_t,y_t)\}_{t=1}^{|\cd|}$, 
where $y_t$ is an evaluation of $f$ at $\vx_t$ 
corrupted by i.i.d additive Gaussian noise $\mathcal N(0,\zeta^2)$,
we obtain a posterior \gp{}, with 
mean  
\begin{equation*}
    \mu(\vx) = \vk^{\cd}(\vx)\T(\mK^\cd+\zeta^2\mI)^{-1}\vy^\cd
\end{equation*}
and covariance  
\begin{equation*}
    \sigma^2(\vx, \vx') = k(\vx,\vx') - \vk^\cd(\vx)\T(\mK^\cd+\zeta^2\mI)^{-1} \vk^\cd(\vx')
\end{equation*}
where the kernel matrix $\mK^\cd
=\left[k(\vx_i,\vx_j)\right]_{\vx_i,\vx_j\in \cd}$ and $\vk^\cd(\vx) =
[k(\vx_i,\vx)]_{\vx_i\in \cd}$~\citep{rasmussen2006gaussian}. 
With slight abuse of notation, we denote the posterior variance by
$\sigma^2(\vx) = \sigma^2(\vx,\vx)$, and the posterior \gp{} by $\GP(\mu,
\sigma)$.
We use \cite{gpy2014} for \gp{} training with Auto Relevance Determination (ARD)~\citep{wipf2008new} to optimize for kernel hyper-parameters that maximize the likelihood of the data.


\section{PDDLStream for TAMP} 
\label{sec:pddlstream}

\stripstream{}~\citep{garrett2020PDDLStream} is a framework for planning in the presence of sampling procedures. 
The use of sampling procedures enables \stripstream{} to address hybrid discrete-continuous planning domains, such as {\sc tamp} domains.
\stripstream{} extends Planning Domain Definition Language (\PDDL{})~\citep{mcdermott1998pddl} by adding {\em streams}, declarative specifications of conditional samplers. 
Streams have previously been implemented by a human engineer through leveraging collision checkers, inverse kinematic solvers, and off-the-shelf motion planners.
In this work, we learn new conditional samplers for dynamic skills, such as pouring and scooping skills, which are difficult for a human to correctly specify.
An open-source implementation of \stripstream{} is available at: {\small \url{https://github.com/caelan/pddlstream}}.


\subsection{PDDL}

In \PDDL{}, states consist of a set of true {\em facts}, which are equivalent to parameterized Boolean variables.
Actions (\pddlkw{:action}) are defined by a set of free {\em parameters} (\pddlkw{:param}), a {\em precondition} logical formula of facts (\pddlkw{:pre}) that must hold in a state in order to apply the action, and an {\em effect} logical conjunction of facts (\pddlkw{:eff}) that specifies which facts are set to true or false after applying the action.


Figure~\ref{fig:app-pour} gives the \PDDL{} description of two of the actions that we consider: \pddl{pour} and \pddl{scoop}.
These actions use the parameters \pddl{?arm}, \pddl{?bowl}, \pddl{?cup}, \pddl{?spoon}, and \pddl{?particles} to refer to a robot arm, bowl, cup, spoon, and set of ``liquid'' particles.
Additionally, departing from typical \PDDL{} models, several of our parameters are multi-dimensional {\em continuous} values:
\pddl{?pose} is a stable object placement in $\SE{3}$; \pddl{?grasp} is a rigid gripper grasp of an object in $\SE{3}$; \pddl{?conf} is a robot arm configuration (set of $d$ joint angles) in $R^d$; \pddl{?obj-path} is an object path consisting of a sequence of poses; and \pddl{?arm-path} is a robot arm path consisting of a sequence of configurations.

The \pddl{pour} action can be applied if \pddl{?cup} initially contains \pddl{?particles}.
After execution, \pddl{?bowl} now contains \pddl{?particles} {\em instead} of \pddl{?cup} as successful pours transfer the full contents of \pddl{?cup} into \pddl{?bowl}.
The \pddl{?scoop} action can be applied if \pddl{?bowl} initially contains \pddl{?particles}.
After execution, \pddl{?spoon} now {\em also} contains \pddl{?particles}.
Critically, \pddl{pour} and \pddl{scoop} have \pddl{GoodPour}, \pddl{GoodScoop}, and \pddl{Motion} preconditions defined on their parameter values.
The \pddl{GoodPour} and \pddl{GoodScoop} conditions enforce that the action parameter values correspond to pours and scoops that are likely-to-be successful.
The objective for sampling and thus learning is to produce parameter values that satisfy these constraints.
The \pddl{Motion} fact relates the path of the robot arm to the path of a grasped cup or spoon.
See~\cite{garrett2020PDDLStream} for descriptions of \pddl{move}, \pddl{pick}, and \pddl{place} actions that are representative of the similar actions used in this work. 

\begin{figure*}[ht]
\begin{small}
\begin{lstlisting}
(|\pddlkw{:action}| pour
 |\pddlkw{:param}| (?arm ?bowl ?pose ?cup ?cup-path ?particles ?grasp ?conf1 ?conf2 ?arm-path)
 |\pddlkw{:pre}| (|\textbf{and}| |\underline{(GoodPour ?bowl ?pose ?cup ?cup-path)}|
           |\underline{(Motion ?arm ?cup ?grasp ?cup-path ?conf1 ?conf2 ?arm-path)}|
           (Particles ?particles) (HasParticles ?cup ?particles)
           (AtPose ?bowl ?pose) (AtGrasp ?arm ?cup ?grasp) (AtConf ?arm ?conf1)
           (|\textbf{not}| (UnsafePath ?arm ?arm-path)))
 |\pddlkw{:eff}| (|\textbf{and}| (AtConf ?arm ?conf2) (HasParticles ?bowl ?particles)
           (|\textbf{not}| (AtConf ?arm ?conf1)) (|\textbf{not}| (HasParticles ?cup ?particles))))
           
(|\pddlkw{:action}| scoop
 |\pddlkw{:param}| (?arm ?bowl ?pose ?spoon ?spoon-path ?particles ?grasp ?conf1 ?conf2 ?control)
 |\pddlkw{:pre}| (|\textbf{and}| |\underline{(GoodScoop ?bowl ?pose ?spoon ?spoon-path)}|
           |\underline{(Motion ?arm ?spoon ?grasp ?spoon-path ?conf1 ?conf2 ?arm-path)}|
           (Particles ?particles) (HasParticles ?bowl ?particles)
           (AtPose ?bowl ?pose) (AtGrasp ?arm ?spoon ?grasp) (AtConf ?arm ?conf1)
           (|\textbf{not}| (UnsafeControl ?arm ?control)))
 |\pddlkw{:eff}| (|\textbf{and}| (AtConf ?arm ?conf2) (HasParticles ?spoon ?particles)
           (|\textbf{not}| (AtConf ?arm ?conf1))))
\end{lstlisting}
\end{small}
\caption{The description of the \pddl{pour} and \pddl{scoop} actions. The underlined preconditions highlight facts that are certified by the \gp{} learners.} \label{fig:app-pour} 
\end{figure*}

\subsection{Streams} \label{sec:streams}

Streams are the key extension of \PDDL{} that enable planning for high-dimensional, continuous systems.
Streams have a procedural and a declarative component.
The procedural component is a function from a set of input values to a sampler that generates a sequence of output values.
This procedure is implemented using a programming language, such as Python.
The declarative component specifies conditions on legal input values as well as properties that any generated output values are guaranteed to satisfy.
Its syntax similar to that of actions: the {\em input parameters} (\pddlkw{:inp}) and the {\em output parameters} (\pddlkw{:out}) specify the number and names of streams inputs and outputs.
The {\em domain} keyword (\textbf{:dom}) specifies a logical formula of ``typing'' facts that \pddlkw{:inp} values must satisfy in order to be legal inputs to the sampler.
The {\em certified} keyword (\pddlkw{:cert}) asserts a logical conjunction of facts that pairs of \pddlkw{:inp} and \pddlkw{:out} values {\em always} satisfy. 

Figure~\ref{fig:app-sample-pour} gives the \stripstream{} description of the \pddl{sample-pour-path}, \pddl{sample-scoop-path}, and \pddl{follow-obj-path} streams, which sample values that serve as inputs to the \pddl{pour} and \pddl{scoop} actions.
The \pddl{sample-pour-path} and \pddl{sample-scoop-path} streams take in as inputs a \pddl{?bowl} at a specific \pddl{?pose} as well as a \pddl{?cup} or \pddl{?spoon}.
They output \pddl{?cup} or \pddl{?spoon} paths sampled from the set of paths that are predicted to be within the super-level set of pours or scoops. 
These streams plan for a manipulated object before considering the robot at all.
The \pddl{sample-obj-path} stream takes in as inputs an \pddl{?arm}, object \pddl{?obj} held at \pddl{?grasp}, and a desired path the object should follow.
It outputs robot arm paths such that \pddl{?obj} follows \pddl{?obj-path} when at \pddl{grasp}. 
The specification these three streams modularly separates primitive sampling operations that are solvable using traditional model-based algorithms, such as motion planning, from those that are better addressed using learning.
As a result, our planning approach retains the generalization and theoretical benefits of model-based approaches while also exhibiting the flexibility of learning methods when primitive models are not known.
See~\cite{garrett2020PDDLStream} for descriptions of streams that sample object placements, object grasps, and robot transit motions.

\begin{figure*}[ht]
\begin{small}
\begin{lstlisting}
(|\pddlkw{:stream}| sample-pour-path
 |\pddlkw{:inp}| (?bowl ?pose ?cup)
 |\pddlkw{:dom}| (|\textbf{and}| (Bowl ?bowl) (Pose ?bowl ?pose) (Cup ?cup))
 |\pddlkw{:out}| (?cup-path)
 |\pddlkw{:cert}| (|\textbf{and}| |\underline{(GoodPour ?bowl ?pose ?cup ?cup-path)}| (ObjPath ?cup ?cup-path)))
 
(|\pddlkw{:stream}| sample-scoop-path
 |\pddlkw{:inp}| (?bowl ?pose ?spoon)
 |\pddlkw{:dom}| (|\textbf{and}| (Bowl ?bowl) (Pose ?bowl ?pose) (Spoon ?spoon))
 |\pddlkw{:out}| (?spoon-path)
 |\pddlkw{:cert}| (|\textbf{and}| |\underline{(GoodScoop ?bowl ?pose ?spoon ?spoon-path)}| (ObjPath ?spoon ?spoon-path)))
 
(|\pddlkw{:stream}| follow-obj-path
 |\pddlkw{:inp}| (?arm ?obj ?grasp ?obj-path)
 |\pddlkw{:dom}| (|\textbf{and}| (Arm ?arm) (Grasp ?obj ?grasp) (ObjPath ?obj ?obj-path))
 |\pddlkw{:out}| (?conf1 ?conf2 ?arm-path)
 |\pddlkw{:cert}| (|\textbf{and}| |\underline{(Motion ?arm ?obj ?grasp ?obj-path ?conf1 ?conf2 ?arm-path)}|
            (Conf ?arm ?conf1) (Conf ?arm ?conf2) (ArmPath ?arm ?arm-path))
\end{lstlisting}
\end{small}
\caption{The descriptions of the \pddl{sample-pour-path}, \pddl{sample-scoop-path}, and \pddl{follow-obj-path} streams, which certify the \pddl{GoodPour} predicate, \pddl{GoodScoop} predicate, and \pddl{Motion} predicates respectively.
} \label{fig:app-sample-pour} 
\end{figure*}


\begin{figure*}
    \centering
    \includegraphics[width=\textwidth]{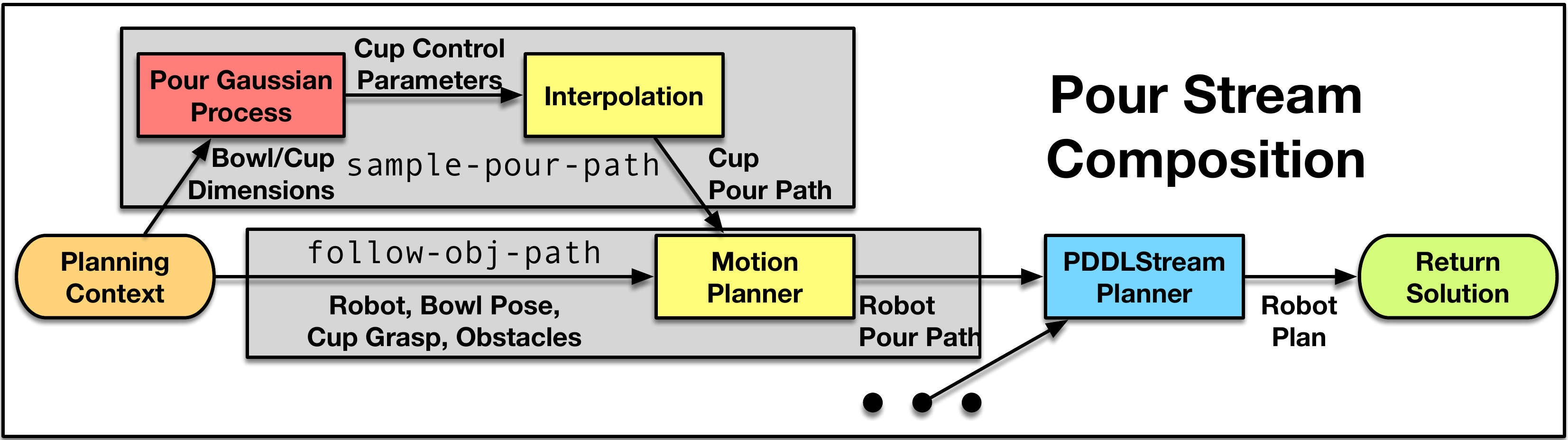}
    \caption{A flowchart that visualizes how the \gp{}s connect to the \pddl{sample-pour-path} and \pddl{follow-obj-path} streams, which certify facts present in \pddl{pour} and \pddl{move} action preconditions. 
    }
    \label{fig:app-test-flowchat}
\end{figure*}

Figure~\ref{fig:app-test-flowchat} demonstrates how the \pddl{sample-pour-path} and \pddl{follow-obj-path} streams compose to ultimately produce robot pouring paths for control parameters values sampled by the pour \gp{}.
The \pddl{sample-pour-path} stream takes in the model of a bowl and cup and featurizes the models using their dimensions, producing a context parameter for the \gp{} learner.
The \gp{} samples a control parameter, which specifies waypoints for the cup.
We linearly interpolate through these waypoints to produce a full path for the cup, which is the output of the \pddl{sample-pour-path} stream.
The \pddl{follow-obj-path} stream takes in the model of the static environment, a model of the robot, and tool paths produced by \pddl{sample-pour-path}, \pddl{sample-scoop-path}, or another stream.
Using Cartesian trajectory tracking, it solves for a robot path that follows the tool path for a particular grasp.
The PDDLStream planner instantiates the \pddl{pour} action using these paths and solves for a plan that uses these and other actions to achieve the goal.

\subsection{Algorithms} \label{sec:incremental}

\stripstream{} problems consist of an initial state, goal state, set of actions, and set of streams.
\stripstream{} algorithms are {\em domain-independent}, meaning that they are able to solve \stripstream{} problems without any additional problem information.
The simplest \stripstream{} algorithm, the \proc{incremental} algorithm~\citep{GarrettRSS17,garrettIJRR2018}, iteratively alternates between a sampling and a searching phase.
During its sampling phase, it passes all legal combinations of input values to each stream and attempts to sample new output values.
During its searching phase, it performs a discrete search, such as a breadth-first search, on the discretized state space resulting from the finite set of currently sampled values. 
If the discrete search finds a solution, \proc{incremental} terminates.
Otherwise, this process repeats.
More advanced algorithms can also be applied using the exact same \stripstream{} problem description.
For example, the \proc{focused} algorithm~\citep{GarrettRSS17,garrettIJRR2018} first searches over {\em plan skeletons}, plans with free parameters, before attempting to sample values for the parameters.
This allows \proc{focused} to more intelligently identify which samplers are relevant for solving the task.


\section{Simulated {\em Kitchen2D} domain}
\label{sec:kitchen2D}

In the earlier version of this work~\citep{Wang2018ActivePlanning}, we built a simulated 2D kitchen environment, {\em Kitchen2D}, based on the physics engine Box2D~\citep{box2d}. For completeness, we include descriptions of {\em Kitchen2D} and its corresponding empirical results for the algorithms in section~\ref{sec:estpre} and section~\ref{sec:planning}. 

In {\em Kitchen2D}, we build in the policies for different skills, {\it e.g.} pouring, scooping, pushing, and demonstrate that it is sample-efficient to learn models of additional skills. Once we obtain those learned models, sampling-based task and motion planners like PDDLStream can make use of them in an effective way to plan efficiently. 

\subsection{Implementation of Kitchen2D} 
\begin{figure*}
    \vskip 0.1in
    \centering
    \includegraphics[width=1.9\columnwidth]{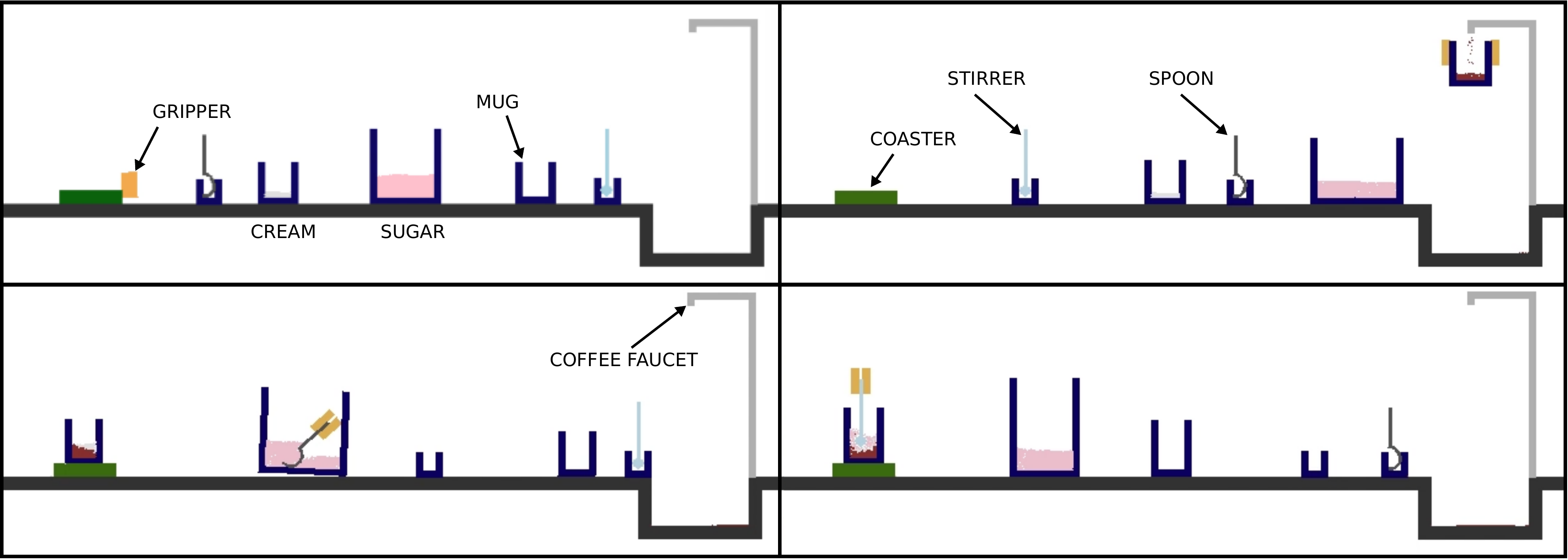}
    \caption{Four arrangements of objects in 2D kitchen, including: green
      coaster, coffee faucet, yellow robot grippers, sugar scoop, stirrer,
    coffee mug, small cup with cream, and larger container with pink sugar.}
    \label{fig:app-kitchen2d}
    \vskip -0.1in
\end{figure*}


\begin{figure}
\centering
\includegraphics[width=\columnwidth]{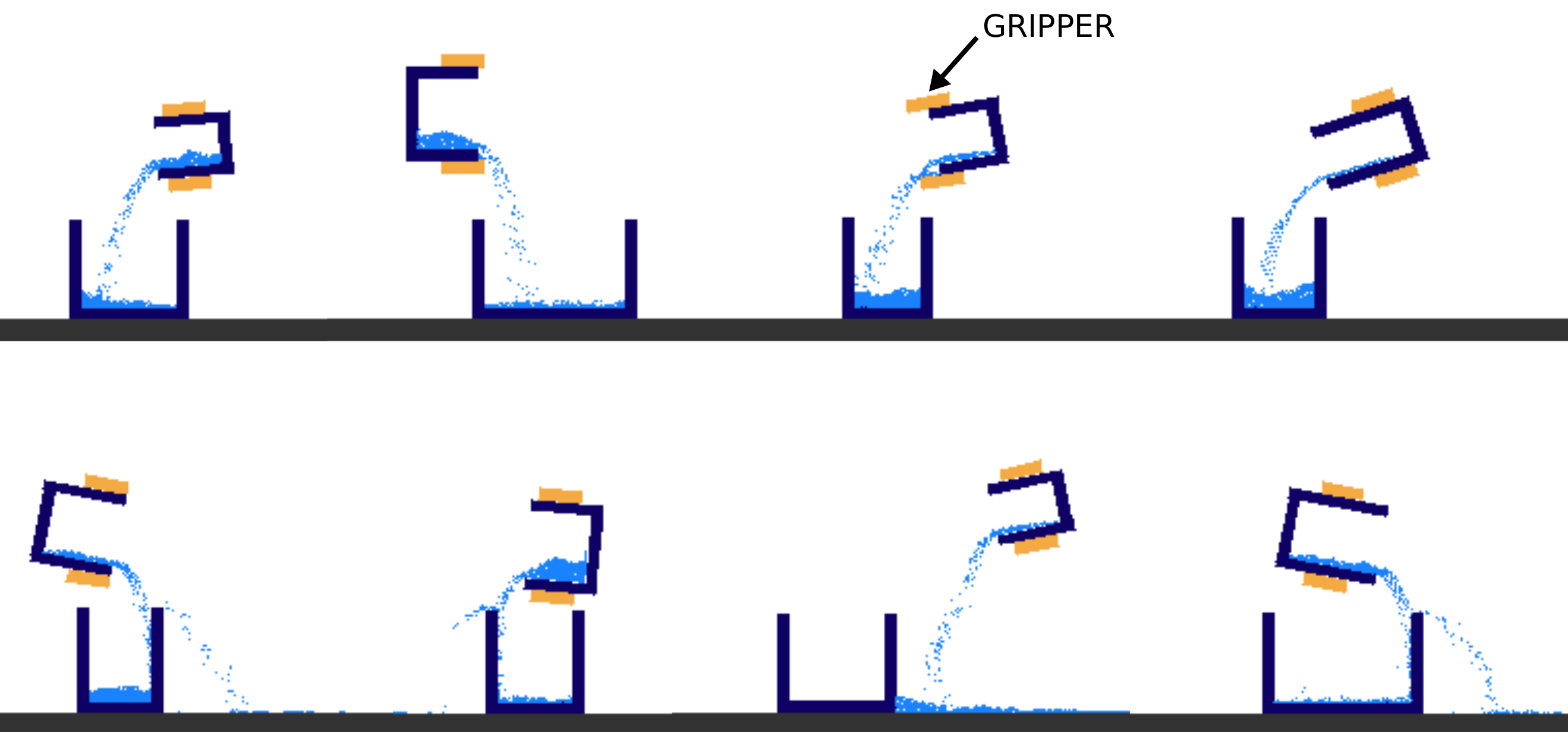}
\caption{Examples of a gripper executing a pouring primitive in {\em Kitchen2D} for several contexts (cup dimensions) and control parameters (relative cup poses).}
\label{fig:app-pouring}
\end{figure}

Figure~\ref{fig:app-kitchen2d}
shows several scenes indicating the variability of arrangements of
objects in the domain.
The parameterized 
actions are: moving the robot (a simple ``free-flying'' hand),
picking up an object, placing an object, pushing an object, filling
a cup from a faucet, pouring a material out of a cup, 
scooping material into a spoon, and dumping material from a spoon. 
The gripper has 3 general movement degrees of freedom (2D position and rotation) and can also open and close its fingers. The
material to be poured or scooped is simulated as small circular
particles. 
We use RRT-Connect~\citep{KuffnerLaValle} to plan motions for the gripper. 

We learn models and samplers for three of these action primitives:
pouring (4 context parameters, 4 predicted parameters, scooping (2
context parameters, 7 predicted parameters), and pushing (2 context
parameters, 6 predicted parameters).  The robot executes trajectories consisting of sequences of waypoints for the gripper, relative to the object it is
interacting with.  

As an example, figure~\ref{fig:app-pouring} illustrates several instances of a
parameterized sensorimotor policy for pouring in {\em Kitchen2D}.
The skill has control parameters $\theta$ that
govern the rate at which the cup is tipped and target velocity of the
poured material.  In addition, several properties of the situation in
which the pouring occurs are very relevant for its success: robot
configuration $c_R$, pouring cup pose and size $p_A, s_A$, and target
cup pose and size $p_B, s_B$.  To model the effects of the action we
need to specify $c_R'$ and $p_A'$, the resulting robot configuration
and pose of the pouring cup $A$.  Only for some settings of the parameters is the action
feasible ({\it i.e.} $\chi(c_R, p_A, s_A, p_B, s_B, \theta, c'_R, p'_A) = 1$): the objective of our work is to efficiently learn a
representation of the feasible region $\chi$ so as to enable a {\sc tamp} planner
to use the skill, in conjunction with other skills, to solve novel problems. 

For pouring, we use the scoring function
$g_{\it pour}(x) = \exp(2(10x - 9.5)) - 1$, where $x$ is the
proportion of the liquid particles that are poured into the target
cup. The constraint $g_{\it pour}(x)>0$ means at least $95\%$ of the
particles are poured correctly to the target cup. 
The context of pouring includes the sizes of the cups, with widths ranging from $3$
to $8$ (units in Box2D), and heights ranging from $3$ to $5$.  For
scooping, we use the proportion of the capacity of the scoop that is
filled with liquid particles, and the scoring function is
$g_{\it scoop}(x) = x-0.5$, where $x$ is the proportion of the spoon
filled with particles.  We fix the size of the spoon and learn the
action parameters for different cup sizes, with width ranging from $5$
to $10$ and height ranging from $4$ to $8$.  For pushing, the scoring
function is $g_{\it push}(x) = 2-\|x-x_{\it goal}\|$ where $x$ is the
position of the pushed object after the pushing action and
$x_{\it goal}$ is the goal position; here the goal position is the
context. The pushing action learned in section~\ref{ssec:app-exp_active} has
the same setting as~\cite{kaelbling2017learning}, viewing the
gripper and object with a bird-eye view.  
The code for the simulation and learning methods is public at 
{\small \url{https://ziw.mit.edu/projects/kitchen2d/}}.


\subsection{Experiments in Kitchen2D}
\label{ssec:app-exp_active}

Similar to our experiments in {\em Kitchen3D}, we show the effectiveness and efficiency of each component of our method independently, and then demonstrate their
collective performance in the context of planning for long-horizon tasks in a high-dimensional continuous domain.

\subsubsection{Active learning:} 
We first demonstrate the performance of using a \gp{} with the straddle
algorithm (\lse) to estimate the level set 
of the constraints on parameters for pushing, pouring and scooping in {\em Kitchen2D}. 
For comparison, we also implemented a simple
method~~\citep{kaelbling2017learning}, which uses a neural network to
map $(\theta, \alpha)$ pairs to predict the probability of success
using a logistic output.   Given a partially trained network and 
a context $\alpha$, the $\theta^* = \argmax_\theta {\rm NN}(\alpha,
\theta)$ which has the highest probability of success with $\alpha$ is
chosen for execution.  Its success or failure is observed, and then
the network is retrained with this added data point.  This method is
called ${\rm NN}_c$ in the results.  In addition, we implemented a
regression-based variation that predicts $g(\theta, \alpha)$ with a
linear output layer, but given an $\alpha$ value still chooses the
maximizing $\theta$.  This method is called ${\rm NN}_r$.  We also
compare to random sampling of $\theta$ values, without any training.

%
%

\begin{figure*}[t]
\vskip 0.1in
\centering
\includegraphics[width=0.99\textwidth]{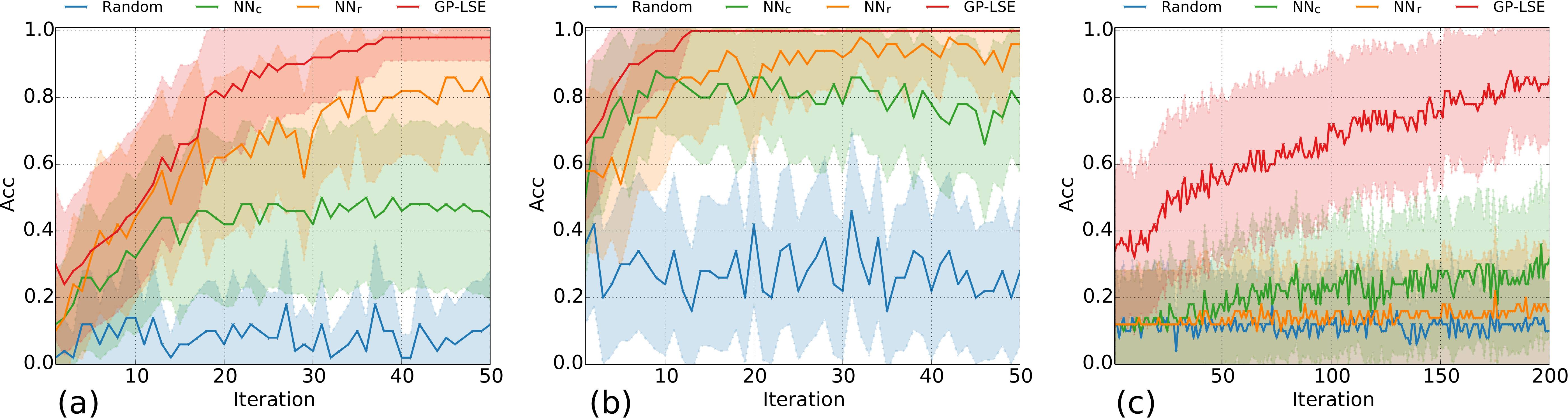}
\caption{Mean success rate (with 1/2 standard deviation on mean shaded)  
  of the first action recommended by random selection
  (Random), regression-based neural network (${\rm NN}_r$),
   classification-based neural network (${\rm NN}_c$) and Gaussian
   process using level-set estimation (\lse) on (a) a pouring task with
   8 parameters (4 are context parameters); (b) a scooping task with 9
   parameters (2 are context parameters) , and (c) a pushing task with
   6 parameters (2 are context parameters).
}
\label{fig:app-model}
\vskip -0.1in
\end{figure*}

\lse is able to learn much more efficiently than the other methods. 
Figure~\ref{fig:app-model} shows the success rate of the first action parameter
vector $\theta$ (value 1 if the action with parameters $\theta$ is
actually successful and 0 otherwise) recommended by each of these
methods as a function of the number of actively gathered training
examples. The results are evaluated through simulation in {\em Kitchen2D}. 
{\sc gp-lse} recommends its first $\theta$ by maximizing
the probability that $g(\theta, \alpha) > 0$. The neural-network
methods recommend their first $\theta$ by maximizing the output value,
while {\sc random} always selects uniformly randomly from the domain
of $\theta$.  

In every case, the \gp{}-based method achieves 
high accuracy well before the others, demonstrating the
effectiveness of uncertainty-driven active sampling methods.



\subsubsection{Adaptive and diverse sampling:}
Given a probabilistic estimate of good $\theta$ values,
obtained by \lse, the next step is to sample values
from that set for planning.  We compare simple rejection
sampling using a uniform proposal distribution (\rej), the basic
adaptive sampler from section~\ref{ssec:adaptive}, and the
diversity-aware sampler from section~\ref{ssec:diverse} with a fixed
kernel:  the results are shown in Table.~\ref{tb:app-sampling}.
The setting for these experiments is exactly as described in section!\ref{ssec:exp-adaptive}

\begin{figure}
\centering
\includegraphics[width=1.0\columnwidth]{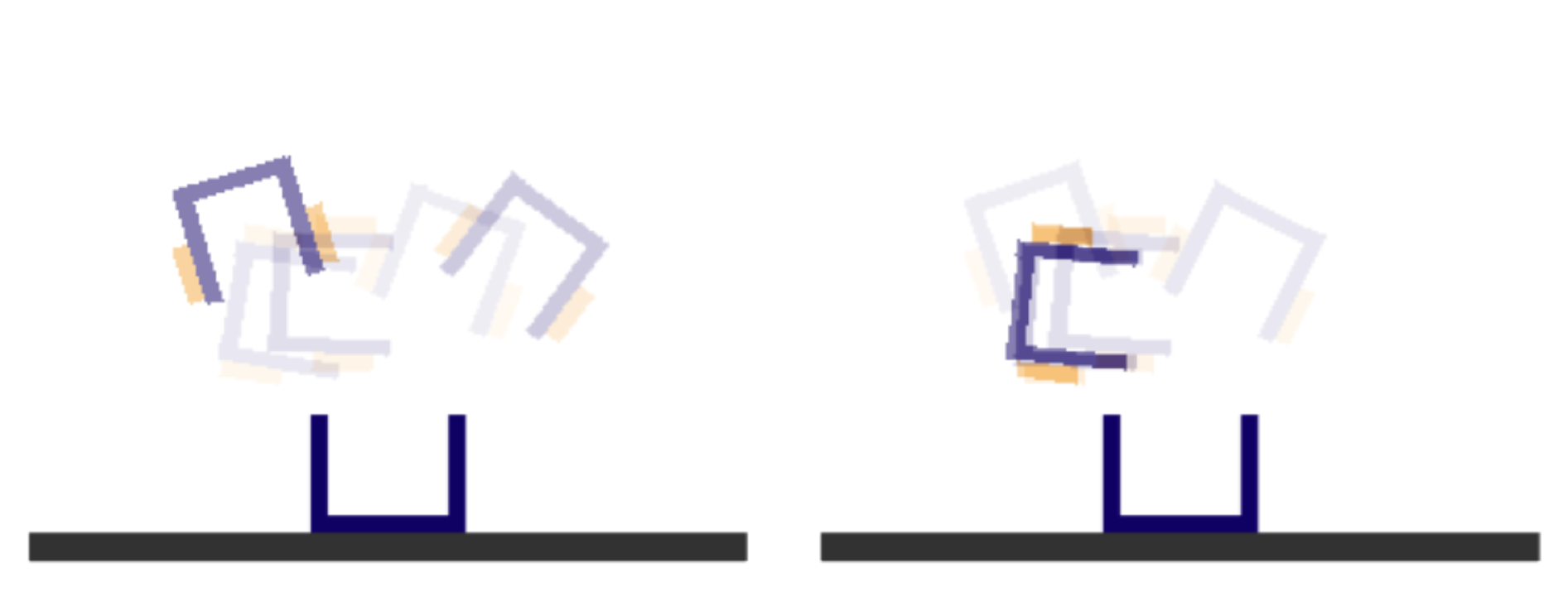}
\caption{Comparing the first 5 samples generated by \diverse
  (left) and \adapt (right) on one of the
  experiments for pouring. The more transparent the pose, the later it
  gets sampled.} 
\label{fig:app-gp-lse-d}
\end{figure}

\begin{table}
\caption{
Effectiveness of adaptive and diverse sampling.
} 
\label{tb:app-sampling}
\begin{center}
\begin{footnotesize}
\begin{tabular}{llccc}
\hline
\abovestrut{0.15in}\belowstrut{0.10in}

&   & \rej & \adapt & \diverse  \\
\hline
\abovestrut{0.10in}
\parbox[t]{0mm}{\multirow{4}{*}{\rotatebox[origin=c]{90}{Pour}}}& FP (\%) & $6.45\pm 8.06$* & {\color{red}$4.04\pm 6.57$} & $5.12\pm 6.94$ \\
& $T_{50}$ (s) & $3.10\pm 1.70$* &  $0.49\pm 0.10$ & $0.53\pm 0.09$ \\
& $N_5$ & $5.51\pm 1.18$* &$5.30\pm 0.92$ & $5.44\pm 0.67$ \\
& Diversity & $17.01\pm 2.90$* & $16.24\pm 3.49$ &  {\color{red}$18.80\pm 3.38$} \\
\hline
\abovestrut{0.10in}
\parbox[t]{0mm}{\multirow{4}{*}{\rotatebox[origin=c]{90}{Scoop}}}& FP (\%) & $0.00^{\dagger}$ & {\color{red}$2.64\pm 6.24$} & $3.52\pm 6.53$ \\
& $T_{50}$ (s) & $9.89\pm 0.88^{\dagger}$ &  $0.74\pm 0.10$ & $0.81\pm 0.11$ \\
& $N_5$ & $5.00^{\dagger}$ & $5.00\pm 0.00$ & $5.10\pm 0.41$ \\
& Diversity & $21.1^{\dagger}$ & $20.89\pm 1.19$ & {\color{red}$21.90\pm 1.04$} \\
\hline
\abovestrut{0.10in}
\parbox[t]{0mm}{\multirow{4}{*}{\rotatebox[origin=c]{90}{Push}}}& FP (\%) & $68.63\pm 46.27^\ddagger$ & {\color{red}$21.36\pm 34.18$} & $38.56\pm 37.60$ \\
& $T_{50}$ (s) & $7.50\pm 3.98^\ddagger$ &  $3.58\pm 0.99$ & $3.49\pm 0.81$ \\
& $N_5$ & $5.00\pm 0.00^\ddagger$ & $5.56\pm 1.51^\triangle$ & $6.44\pm 2.11^\clubsuit$  \\
& Diversity &  $23.06\pm 0.02^\ddagger$ & $10.74\pm 4.92^\triangle$ & $13.89\pm 5.39^\clubsuit$ \\
\hline
\end{tabular}
\end{footnotesize}
\end{center}
*1 out of 50 experiments failed (to generate 50 samples within $10$ seconds);
${}^\dagger$49 out of 50 failed;
${}^\ddagger$34 out of 50 failed;
5 out of 16 experiments failed (to generate 5 positive samples within
$100$ samples); 
${}^\triangle$7 out of 50 failed;
${}^\clubsuit$11 out of 50 failed.
\vskip -0.1in
\end{table}

In these experiments, as in the ones in section~\ref{ssec:exp-adaptive}, \diverse uses more samples than
\adapt to achieve 5 positive ones, and its false positive rate
is slightly higher than \adapt, but the diversity of the samples is notably higher.  The FP rate of \diverse can be
decreased by increasing the confidence bound on the level set.
We illustrate the ending poses of the 5 pouring actions generated by
adaptive sampling with \diverse and \adapt in
Figure~\ref{fig:app-gp-lse-d} illustrating that \diverse is able to generate more diverse action parameters, which may facilitate planning.

\subsubsection{Learning kernels for diverse sampling:} 

\begin{table}
\caption{Effect of distance metric learning on sampling.}
\label{tb:app-timing}
\vskip -0.3in
\begin{center}
\begin{footnotesize}
\begin{tabular}{lccc}
\hline
\abovestrut{0.15in}\belowstrut{0.10in}
 Task I& Runtime (ms) & 0.2s SR  (\%) & 0.02s SR (\%) \\
\hline
\abovestrut{0.10in}
\adapt    & $8.16\pm 12.16$  & $100.0\pm 0.0$& $87.1\pm 0.8$ \\
\gk & $9.63\pm 9.69$        & $100.0\pm 0.0$ & $82.2\pm 1.2 $\\
\lk  & {\color{red}   $5.87\pm4.63$} & $100.0\pm 0.0$ &  {\color{red}$99.9\pm 0.1$}       \\
\hline
\abovestrut{0.15in}\belowstrut{0.10in}
Task II & Runtime (s) & 60s SR  (\%) & 6s SR (\%) \\
\hline
\abovestrut{0.10in}
\adapt  & $3.22\pm 6.51$  & $91.0\pm 2.7$& $82.4\pm 5.6$ \\
\gk & $2.06\pm 1.76$        &  {\color{red} $95.0\pm 1.8 $} & $93.6\pm 2.2 $\\
\lk  & {\color{red}$1.71\pm 1.23$} & {\color{red} $95.0\pm 1.8$} &  {\color{red}$94.0\pm 1.5$}       \\
\hline
\abovestrut{0.15in}\belowstrut{0.10in}
Task III & Runtime (s) & 60s SR  (\%) & 6s SR (\%) \\
\hline
\abovestrut{0.10in}
\adapt   & $5.79\pm 11.04$  & $51.4\pm 3.3$& $40.9\pm 4.1$ \\
\gk & $3.90\pm 5.02$        &  $56.3\pm 2.0 $ & $46.3\pm 2.0 $\\
\lk  & $4.30\pm 6.89$ & {\color{red} $59.1\pm 2.6$} &  {\color{red}$49.1\pm 2.6$}       \\
\hline
\end{tabular}
\end{footnotesize}
\end{center}
\vskip -0.2in
\end{table}

Finally we explore the effectiveness of the
diverse sampling algorithm with task-level kernel learning; the setting is analogous to the one in section~\ref{ssec:exp_plan}. 
We compare \adapt, \gk with a fixed kernel, and diverse sampling with
learned kernel (\lk), in every case using a high-probability
super-level-set estimated by a~\gp. In \lk, we use $\epsilon=0.3$.
\begin{figure*}
\vskip 0.1in
\centering
\includegraphics[width=1.0\textwidth]{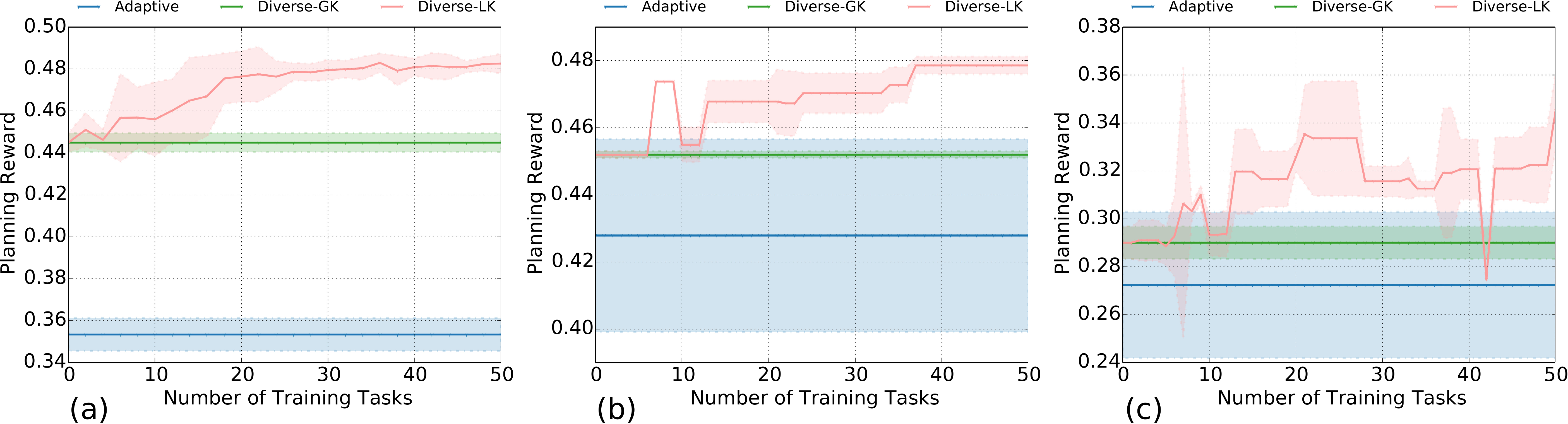}
\caption{The mean learning curve of reward $J(\phi)$ (with 1.96
  standard deviation) as a function of the number of training tasks in
  three domains: 
  (a) pushing an object off the
  table (b) pouring into a cup next to a wall (c) picking up a cup in
  a holder and pour into a cup next to a wall.} 
\label{fig:app-alltask}
\vskip -0.1in
\end{figure*}
We define the planning reward of a sampler to be
$J_k(\phi) =\sum_{n=1}^\infty s(\phi, n)\gamma^n$, where $s(\phi, n)$
is the indicator variable that the $n$-th sample from $\phi$ helped
the planner to generate the final plan for a particular task instance
$k$. The reward is discounted by $\gamma^n$ with $0< \gamma < 1$, so
that earlier samples get higher rewards (we use $\gamma=0.6$). We
average the rewards on tasks drawn from a predefined distribution,
and effectively report a lower bound on $J(\phi)$, by setting a time
limit on the planner. 
 
The first set of tasks (Task I) we consider is a simple controlled
example where the goal is to push an object off a 2D table with the
presence of an obstacle on either one side of the table or the other
(both possible situations are equally likely).  The presence of these
obstacles is not represented in the context of the sampler, but the
planner will reject sample action instances that generate a collision
with an object in the world and request a new sample. 
We use a fixed range of
feasible actions 
sampled from two rectangles in 2D of unequal sizes. 
The optimal strategy is to first randomly sample from one side of the
table and if no plan is found, sample from the other side.

We show the learning curve of \lk with respect to the planning reward 
metric $J(\phi)$ in figure~\ref{fig:app-alltask} (a). 1000 initial
arrangements of obstacles were drawn randomly for testing. 
 We also repeat the experiments 5 times to obtain the $95\%$ confidence 
 interval. 
For \gk, the kernel inverse is initialized
as $[1,1]$ and if, for example, it sampled on the left side of the
object (pushing to the right) and the obstacle is on the right, it may
not choose to sample on the right side because the kernel indicates
that the other feature is has more diversity. However, after a few
planning instances, \lk is able to figure out the right
configuration of the kernel and its sampling strategy becomes the
optimal one. 

We also tested these three sampling algorithms on two more complicated
tasks. We select a fixed test set with 50 task specifications and
repeat the evaluation 5 times.  The first one (Task II) involves
picking up cup A, getting water from a faucet, move to a pouring
position, pour water into cup B, and finally placing cup A back in its
initial position. Cup B is placed randomly either next to the wall on
the left or right. The second task is a harder version of Task II,
with the additional constraint that cup A has a holder and the sampler
also has to figure out that the grasp location must be close to
the top of the cup (Task~III).

We show the learning results in figure~\ref{fig:app-alltask} (b) and (c)  
and timing results in table~\ref{tb:app-timing} (after training). 
We conjecture that the sharp turning points in the learning curves of 
Tasks II and III are a result of high penalty on the kernel length scales and 
the limited size (50) of the test tasks, and we plan to investigate more 
in the future work.  
Nevertheless, \lk is still able to find a better
solution than the alternatives in Tasks II and III. 
Moreover, the two diverse sampling
methods achieve lower variance on the success rate and perform more
stably  after training. 



\subsubsection{Integration} 

Finally, we integrate the learned action sampling models for pour and 
scoop  with 7 pre-existing robot operations (move, push, pick, place,
fill, dump, stir) in a domain specification for \stripstream.  The robot's
goal is to ``serve'' a cup of coffee with cream and sugar by placing
it on the green coaster near the edge of the table.  Accomplishing
this requires general-purpose planning, including picking where to
grasp the objects, where to place them back down on the table, and what
the pre-operation poses of the cups and spoon should be before
initiating the sensorimotor primitives for pouring and scooping should
be.  Significant perturbations of the object arrangements are handled
without difficulty. For example, We use the focused algorithm within \stripstream, 
and it solves the task in 20-40 seconds for a range of different arrangements of objects.
Some resulting plans and execution sequences can be found at {\small \url{https://ziw.mit.edu/projects/kitchen2d/}}. 

In summary, our experiments in {\em Kitchen2D} illustrate a critical ability:  to augment the existing
competences of a robotic system (such as picking and placing objects)
with new sensorimotor primitives by learning probabilistic models of
their preconditions and effects and using a state-of-the-art
domain-independent continuous-space planning algorithm to combine them
fluidly and effectively to achieve complex goals.